\documentclass[manuscript,screen]{acmart}

\AtBeginDocument{%
  \providecommand\BibTeX{{%
    \normalfont B\kern-0.5em{\scshape i\kern-0.25em b}\kern-0.8em\TeX}}}

\acmConference{}{}
\usepackage{balance}

\usepackage{amsmath,amsthm}
\usepackage{subcaption}

\usepackage{enumitem}
\usepackage{algorithm2e}
\usepackage{algpseudocode}

\def\sys{CaiTI}

\newif\ifshowedit
\showeditfalse
\usepackage[normalem]{ulem}

\ifshowedit
\newcommand{\remove}[1]{}

\else
\newcommand{\remove}[1]{}

\fi

\usepackage{tabularx}
\usepackage{pifont}
\usepackage{xcolor}
\usepackage{multirow}
\usepackage{caption}
\usepackage{subcaption}
\usepackage{wrapfig}

\RestyleAlgo{ruled}
\begin{document}

\title{LLM-based Conversational AI Therapist for Daily Functioning Screening and Psychotherapeutic Intervention via Everyday Smart Devices}

% \author{Jingping }
% \affiliation{%

\author{Jingping Nie}
\affiliation{%
  \institution{Columbia University}
  \streetaddress{116th and Broadway}
  \city{New York}
  \country{United States}}
\email{jn2551@columbia.edu}

\author{Hanya (Vera) Shao}
\affiliation{%
  \institution{Kensington Wellness}
  \streetaddress{135 Ocean Pkwy}
  \city{New York}
  \country{United States}}
\email{vera}

\author{Yuang Fan}
\affiliation{%
  \institution{Columbia University}
  \streetaddress{116th and Broadway}
  \city{New York}
  \country{United States}}
\email{yf2676@columbia.edu}

\author{Qijia Shao}
\affiliation{%
  \institution{Columbia University}
  \streetaddress{116th and Broadway}
  \city{New York}
  \country{United States}}
\email{qijia@cs.columbia.edu}

\author{Haoxuan You}
\affiliation{%
  \institution{Columbia University}
  \streetaddress{116th and Broadway}
  \city{New York}
  \country{United States}}
\email{hy2612@columbia.edu}

\author{Matthias Preindl}
\affiliation{%
  \institution{Columbia University}
  \streetaddress{116th and Broadway}
  \city{New York}
  \country{United States}}
\email{matthias.preindl@columbia.edu }

\author{Xiaofan Jiang}
\affiliation{%
  \institution{Columbia University}
  \streetaddress{116th and Broadway}
  \city{New York}
  \country{United States}}
\email{jiang@ee.columbia.edu}

\renewcommand{\shortauthors}{Nie et al.}

\begin{teaserfigure}
\centering
  \includegraphics[width=0.9\textwidth]{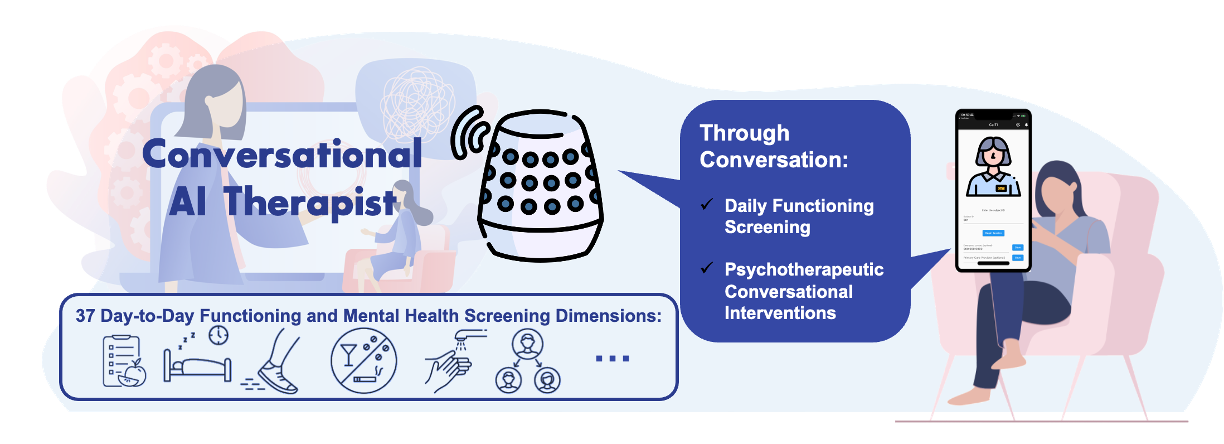}
  \vspace{-0.15in}
  \caption{\sys{} is a conversational ``AI therapist'' that screens and assesses the day-to-day functioning of the occupant across 37 dimensions and provides appropriate psychotherapeutic interventions and empathic validations depending on the physical and mental status of the occupant.}
  \label{fig:teaser}
\end{teaserfigure}
\begin{abstract}
  Despite the global mental health crisis, access to screenings, professionals, and treatments remains high. In collaboration with licensed psychotherapists, we propose a \underline{C}onversational \underline{AI} \underline{T}herapist with psychotherapeutic \underline{I}nterventions (CaiTI), a platform that leverages large language models (LLM)s and smart devices to enable better mental health self-care. \sys{} can screen the day-to-day functioning using natural and psychotherapeutic conversations. \sys{} leverages reinforcement learning to provide personalized conversation flow. \sys{} can accurately understand and interpret user responses. When the user needs further attention during the conversation, CaiTI can provide conversational psychotherapeutic interventions, including cognitive behavioral therapy (CBT) and motivational interviewing (MI). Leveraging the datasets prepared by the licensed psychotherapists, we experiment and microbenchmark various LLMs' performance in tasks along \sys{}'s conversation flow and discuss their strengths and weaknesses. With the psychotherapists, we implement \sys{} and conduct 14-day and 24-week studies. The study results, validated by therapists, demonstrate that CaiTI can converse with users naturally, accurately understand and interpret user responses, and provide psychotherapeutic interventions appropriately and effectively. We showcase the potential of \sys{} LLMs to assist the mental therapy diagnosis and treatment and improve day-to-day functioning screening and precautionary psychotherapeutic intervention systems.

  %We invite submissions that present novel developments and assessments of informatics methods, including those that showcase the strengths and weaknesses of utilizing LLMs in healthcare. We are especially interested in contributions that tackle the complexities and possibilities at the crossroads of these disciplines, providing fresh perspectives on how they can collaborate to push forward healthcare improvements.
\end{abstract}


% \ccsdesc[100]{Human-centered computing~Systems and tools for interaction design}



\maketitle

\section{Introduction}\label{sec:introduction}
Maintaining physical and mental health is crucial for quality of life, particularly for those living alone, experiencing early signs of mental illness, or requiring daily assistance. COVID-19 significantly impacted global mental health with high barriers to accessing mental health screenings and treatments, including home care~\cite{talevi2020mental, WhiteHouse2022}. There are a variety of smart wearables and smart devices to monitor physical and mental health~\cite{nie2021spiders+, morshed2019prediction, zhu2021eit}. Human-Computer Interaction (HCI) researchers are actively working to improve wellness care for the general public and vulnerable population groups~\cite{mishra2019sensing, pendse2021can, tlachac2022studentsadd}. Furthermore, with the growth of the Internet of Things (IoT) devices, the smart speaker market volume reached 200 million and there are more than 6.84 billion smartphones worldwide in 2023 ~\cite{SpeakerMarket, SmartphoneMarket}. Recent advances in artificial intelligence (AI) and large language models (LLMs) further expanded the possibilities for intelligent health-oriented applications~\cite{nie2021spiders+, yunxiang2023chatdoctor, nori2023capabilities, van2023global, dai2023detecting}. 

%% Limitations 
While existing research primarily focuses on understanding emotional states or affective states as indicators of mental well-being \cite{zhou2018emotional}, therapists generally require more knowledge about patients' daily activities and behaviors to accurately assess mental health \cite{helfrich2008mental, bible2017assessment}. Therapists often rely on assessments such as the Daily Living Activities–20 (DLA-20) and the  Global Assessment of Functioning (GAF) to screen day-to-day functions and mental health status~\cite{clausen2016health, guze1995diagnostic, fu2022distributed, morshed2019prediction}. Most existing research efforts focus on screening for physical and mental well-being, with few addressing psychotherapeutic interventions. Psychotherapy refers to a range of interventions based on psychological theories and principles to address emotional and behavioral issues that impact mental health \cite{corey2013theory}. \cite{nie2022conversational} and \cite{zhou2018emotional} propose conversational systems that provide preliminary consolation. While conversational systems and evidence-based treatments like Motivational Interviewing (MI) \cite{naar2017motivational}, Cognitive Behavioral Therapy (CBT) \cite{beck2011cognitive}, and Dialectical Behavior Therapy (DBT) \cite{robins2011dialectical} have been proposed, many lack personalization or user understanding \cite{schroeder2018pocket, sabour2022chatbots}.

Although AI chatbots like ChatGPT show promise in addressing mental health concerns \cite{yusuf2023how, chatgpt_news1}, they often suffer from performance decline over time and limitations in psychotherapeutic considerations \cite{reddit2023using}. Additionally, mental health applications see low usage rates in clinical settings \cite{chandrashekar2018mental, torous2018mental}. Smartphone-based tools may not be user-friendly for individuals with memory or vision impairments, especially the elderly \cite{mohadisdudis2014study}. As such, there is growing interest in exploring objective activity detection through ambient sensing and voice-based chatbots as more inclusive and effective approaches to mental health support.

%% Design Requirements:
Considering these limitations and opportunities, in collaboration with 4 licensed psychotherapists from a major mental health counseling institution with thousands of clients, we propose \emph{\sys{}, a conversational AI therapist that takes advantage of widely-owned smart devices for continuous screening of physical and mental health in a privacy-aware manner, while employing psychotherapeutic interventions} (Figure~\ref{fig:teaser}). Our collaborating psychotherapists have identified several design requirements for \sys{} that can facilitate mental health self-care and assist in psychotherapeutic treatment for individuals: (\emph{i}) provide comprehensive day-to-day functioning screenings and employ evidence-based psychotherapeutic interventions; (\emph{ii}) facilitate natural conversation flow; (\emph{iii}) ensure the quality of care by enabling the system to intelligently interpret user responses and, if necessary, guide the dialogue back toward the psychotherapeutic objectives when the user's responses deviate; and (\emph{iv}) the conversation format (using smartphones/smart speakers) should take into consideration individuals with visual impairments.

%% challenges:
Realizing such a system poses several challenges. Primarily, the system must \emph{fit} within the users' lifestyles and habits, utilizing devices that users already own and prefer. It should facilitate communication through the user's preferred modes—be it verbal or textual—while ensuring \emph{comprehensive screening} and delivering \emph{effective psychotherapeutic interventions} in a \emph{privacy-aware} manner. Additionally, it is imperative that the system is \emph{easy to use} for all individuals, regardless of their technical proficiency. Furthermore, the LLMs in \sys{} should effectively deliver conversational psychotherapy and must be carefully designed to both be \emph{user-friendly/accessible} and \emph{capable} of understanding, reasoning, and responding to an infinitely diverse number of user responses (including both YES/NO answers and open-ended responses). The design of \sys{} must effectively manage varied responses, translate therapists' empirical techniques into a quality-controlled logical flow, and incorporate a recommendation system that dynamically personalizes itself to each user. 

Building upon near-ubiquitous smart devices, \sys{} combines AI techniques, including LLMs, reinforcement learning (RL), and human-computer interaction (HCI) approaches with professional experiences from licensed psychotherapists. 
\sys{} \emph{screens the user along the 37 dimensions of day-to-day functioning} proposed in~\cite{nie2022conversational} by conversing naturally with users with open-ended questions. \sys{} understands verbal and textual responses and activities of the user and employs \emph{conversation-based psychotherapeutic interventions}. To summarize, the main contributions of this paper include:
\begin{itemize}[leftmargin=*, topsep=1mm]
    \item
    {\sys}, an LLM-based conversational ``AI therapist'' that screens and analyzes the day-to-day functioning of users across 37 dimensions. Using the screening results, {\sys} provides appropriate empathic validations and psychotherapies depending on the physical and mental status of the user. {\sys} is accessible through widely available smart devices, including smartphones, computers, and smart speakers, and offers a versatile solution catering to the diverse requirements of the users whether they are indoors or outdoors.
    \item
    To realize more intelligent and \emph{friendly} human-device interaction, we leverage RL to personalize each user's conversation experience during screening in an adaptive manner. {\sys} prioritizes the dimensions that concern psychotherapists more about each user based on his/her historical responses and brings up the dimensions in the order of priority during the conversation.
    \item
    We design the conversation architecture of {\sys} with the therapists, which effectively incorporates Motivational Interviewing (MI) and Cognitive Behavioral Therapy (CBT) -- two commonly used psychotherapeutic interventions administered by psychotherapists -- to provide \emph{Psychotherapeutic Conversational Intervention} in a natural way that closely mirrors the therapists' actual practices.   
    \item 
    To ensure the quality of care and effectiveness of the psychotherapy process and avoid the propagation of biases in AI algorithms and LLMs, {\sys} incorporates multiple task-specific LLM-based \textbf{\texttt{Reasonser}}s, \textbf{\texttt{Guide}}s, and \textbf{\texttt{Validator}} during the psychotherapy process. Leveraging the task-specific conversation datasets prepared and labeled by the licensed psychotherapists, we experiment and microbenchmark the performance of different GPT- and Llama 2-based LLMs with few-shot prompts or fine-tuning in performing tasks along {\sys}'s. \textbf{\emph{We will open-source: (\emph{i}) the datasets prepared by the therapists to facilitate research in this area and (\emph{ii}) the few-shot prompts we designed with the therapists.}}
    \item
    In collaboration with licensed psychotherapists, we design, implement, and deploy a proof-of-concept prototype of {\sys}. Through real-world deployments with 20 subjects for up to 24 weeks, we demonstrate that {\sys} can accurately assess the user's physical and mental status and provide appropriate and effective psychotherapeutic interventions. {\sys} has received positive feedback, endorsements, and validation from both licensed psychotherapists and subjects.
\end{itemize}

\emph{To the best of our knowledge, {\sys} is the first conversational ``AI therapist'' system that leverages smart home devices and LLMs to mimic the psychotherapists' actual practices in clinical sessions and provides continuous monitoring and interaction with the integration of psychotherapies (MI and CBT).}

\section{Psychological Background} \label{sec:psychological}

\subsection{Psychological Assessment}
People who experience mental health adjustment issues and disorders tend to face diminished capacity in professional or academic performance, maintaining social relationships, and self-care~\cite{helfrich2008mental,bible2017assessment}. Traditional screening tools, such as the Mental Status Examination (MSE), require clinicians to observe and assess people's daily functioning, such as physical appearance and presentation, social interaction behaviors, and emotional expression~\cite{trzepacz1993psychiatric}. Other widely used diagnostic assessments, such as the Adult ADHD Self-Report Scale (ASRS-v1.1), the Patient Health Questionnaire-9 (PHQ-9) for depression, and the General Anxiety Disorder-7 (GAD-7), which provide more specific screening options for specific mental health diagnoses, often include questions or items assessing daily functioning~\cite{kroenke2001phq, el2009adult}. For example, the PHQ-9 includes assessments of mood, sleep hygeine, and eating habits~\cite{kroenke2001phq}. 

There are several psychological measurements designed to examine the day-to-day functioning of individuals to evaluate their mental health well-being, such as DLA-20 and GAF~\cite{nie2022conversational,dsm2000diagnostic,scott2001reliability}. DLA, which was developed to evaluate aspects of daily functioning affected by mental illnesses, includes 20 major categories for daily functioning. These categories include interpersonal communication, family relationships, personal hygiene, time management, and productivity at work~\cite{scott2001reliability}. On the other hand, GAF, 
which was introduced in DSM-IV, employs an ordinal scale to evaluate an individual's overall level of functioning~\cite{dsm2000diagnostic}. A lower GAF score indicates the presence of more significant symptoms and difficulties in social, occupational, and psychological functioning.

%%%%%
%%%%%
\subsection{Psychotherapeutic Interventions}
\label{sec:mi_cbt}

Clinicians using evidence-based practices (EBP) in psychology to guide interventions and treatment plans, taking into account relevant research on their clinical practices, have found that the use of EBP helps improve the quality and accountability of clinical practices~\cite{apa2006evidence, spring2007evidence}. Some commonly used EBP include CBT, acceptance and commitment therapy (ACT), DBT, and MI.
CBT is one of the most popular and commonly used psychological interventions. It focuses on challenging one's cognitive distortions and subsequent behaviors to reduce existing mental health symptoms and improve overall mental well-being~\cite{alford1997integrative,beck2011cognitive}. CBT is found to be effective in a variety of diagnoses, such as mood disorders,  Attention-deficit/hyperactivity disorder (ADHD), eating disorders, Obsessive-compulsive disorder, and Post-traumatic stress disorder~\cite{clark2003cognitive,roy2005randomized, emilsson2011cognitive, halmi2005predictors,walsh2004treatment, foa2005randomized, dickstein2013comparing}. CBT also shows promising results in preventative care that may not be tied to a specific diagnosis. It has been effective in various settings, including medical, work, and school environments~\cite{moss2013randomized, tan2014preventing, miller2011effectiveness}. However, despite the abundant evidence of its effectiveness, CBT is associated with a high nonresponse rate, attributed to participants' low motivation~\cite{antony2005improving}. During CBT, therapists assess the validity and utility of participants' responses to understand their thought patterns and beliefs accurately ~\cite{sokol2019comprehensive}. Such an assessment involves identifying, challenging, and reframing cognitive distortions, such as overgeneralization, emotional reasoning, all-or-nothing thinking, catastrophizing, etc 
~\cite{burns1999feeling}.
CBT usually consists of the following steps:
\begin{enumerate}
    \item \textbf{Identify the Situation/Issue:} Start by clearly identifying the situation or issue you want to work on. 
    \item \textbf{Recognize Negative Thoughts:} Think about the thoughts that go through your mind when you experience this issue. These are often automatic or subconscious thoughts that may be irrational or unhelpful. They can be self-critical, overly pessimistic, or unrealistic.
    \item \textbf{Challenge Negative Thoughts:} Challenge means questioning the validity of these thoughts. Are there alternative, more balanced, or rational thoughts that might be more helpful in the situation?
    \item \textbf{Reframe Thoughts and Situations:} Try to reframe your unhelpful thoughts and situations into more balanced, realistic, and constructive ones. This process is about changing the way you think about the situation, which can lead to changes in your emotions and behaviors.
\end{enumerate}

To address issues related to low motivation, researchers have suggested using MI as a complementary approach alongside CBT ~\cite{marker2018efficacy,naar2017motivational,arkowitz2004integrating}. There are four techniques to effectively implement MI ~\cite{miller2012motivational}:
\begin{enumerate}
    \item \textbf{Open-ended questions:} Encouraging elaboration on responses, asking for examples, or exploring the implications of what's been shared;
    \item \textbf{Affirmations:} State strengths and help feel that changes are possible;
    \item \textbf{Reflective listening:} (\emph{i}) Simple reflection: repeating what the client has said, using slightly different words or phrases; (\emph{ii}) Reframe reflection: listening to the client's statements and then reflecting them back in a way that presents a new perspective or interpretation; and (\emph{iii}) Affective reflection: recognizing, understanding, and reflecting back the emotional content of what the client expresses;
    \item \textbf{Summaries:} Use summaries not only to encapsulate discussions but also to highlight progress.
\end{enumerate}
MI is an evidence-based practice for substance use disorders and other addiction issues~\cite{anton2006combined,aarons2017testing}. It is also found to be effective in helping people adapt to various situations, such as managing  diabetes~\cite{kertes2011impact, channon2007multicenter,chen2012effects}. Growing research has shown that the combination of CBT and MI shows effectiveness in a variety of populations and for mental health adjustments~\cite{merlo2010cognitive,marker2018efficacy,kertes2011impact,arkowitz2004integrating}.

\section{Related Work}
\label{sec:related_work}

\subsection{Mental Wellness Self-Screening and Self-Care}
\label{secsec:Wellness Self-Screening and Self-Care}

There are various methods for mental health self-screening~\cite{kruzan2022wanted, brown2016gamification}. While online help-seeking is preferred by many individuals~\cite{gould2002seeking}, these tools provide a limited assessment based on closed-ended questions, potentially leading to omitting important details typically obtained from open-ended questions or in-person interactions~\cite{OnlineTest, OnlineTest2}. Besides the online tools, \cite{liu2022aimse} proposed an AI-based self-administer online web-browser-based mental status examination (MSE). Recently, Experience Sampling Method (ESM) has been widely adopted by HCI researchers for various physical and mental health screening. ESM can be done automatically by sensors or by repeatedly prompting users to answer questions in their normal environments. For example, ESM is used for self-reporting Parkinson's Disease symptoms, chronic pain, designing health technologies for Bipolar Disorder, etc~\cite{vega2018back, adams2018keppi, matthews2015situ}. Most wellness self-screening methods in the literature use close-ended questions and expect close-end results from the user, while \sys{} uses open-ended questions and allows the user to chat freely with any topic. And they usually only focus on particular dimensions of the day-to-day functioning or mental disorders instead of performing a comprehensive screening.

\subsection{LLM-based Healthcare and Mental Healthcare}
\label{secsec:LLM healthcare}
Large Language Models (LLMs) are pre-trained on vast datasets, which equip them with significant prior knowledge and enhanced reasoning skills. The recent state-of-the-art models, including GPT-4~\cite{openai2023gpt4}, GPT-3~\cite{openai2023gpt35}, Claude-3~\cite{anthropic2024claude}, and Gemini 1.5~\cite{google2024gemini15}, exhibt strong capability in reasoning over text. Consequently, recent research increasingly employs LLMs alongside various language, vision, or multimodal models to enable advanced applications in various domains without the need for additional training~\cite{yin2023survey, sharan2023llm, deb2023fill}. For example, IdealGPT combines two LLMs (GPT) with a vision-and-language model to enable a framework that iteratively decomposes vision-and-language reasoning, where the two LLMs are treated as Questioner and Reasoner~\cite{you2023idealgpt}. Additionally, research has shown that LLMs possess the ability to reason with and interpret IoT sensor data~\cite{xu2023penetrative}.

%%%% General Healthcare
Recently, transformer-based Large language foundation models, such as GPT-4~\cite{bubeck2023sparks}, PaLM 2~\cite{anil2023palm}, and LLaMA2~\cite{touvron2023llama}, have demonstrated superior performance across various medical-related NLP tasks. LLMs are used to enable various general healthcare applications. \cite{waisberg2023gpt} showed that GPT-4 has the potential to help drive medical innovation, from aiding with patient discharge notes, summarizing recent clinical trials, and providing information on ethical guidelines. Moreover, Google introduced Med-PaLM and Med-PaLM 2\cite{singhal2023large, singhal2023towards}, LLMs explicitly tailored for the medical domain, providing high-quality responses to medical inquiries.

%%%% Mental Healthcare
Various works also exploit and evaluate the performance of LLMs for mental status classification and assessment. Researchers leveraged LLMs for mental health prediction via online text data and evaluated the capabilities of multiple LLMs on various mental health prediction tasks via online text data~\cite{xu2023leveraging, radwan2024predictive}. In addition, \cite{jiang2023multimodal} leverages RoBERTa~\cite{liu2019roberta} and Llama-65b ~\cite{touvron2023llama} in the system for classifying psychiatric disorder, major depressive disorder, self-rated depression, and self-rated anxiety based on time-series multimodal features. 

%%%%Pyschological counseling}
In addition to assessing and classifying the mental status of the user, researchers have investigated providing psychological consultations. For example, \cite{nie2022conversational} leveraged GPT-3 to construct a home-based AI therapist that detects abnormalities in mental status and daily functioning and generates responses to console users. \cite{lai2023psy} proposed an AI-based assistive tool leveraging the WenZhong model, a pre-trained model trained on a Chinese corpus for question-answering in psychological consultation settings~\cite{lai2023psy}. Researchers also investigate the potential of ChatGPT in powering chatbots to simulate the conversations between psychiatrists and mentally disordered patients~\cite{chen2023llm}. 

Only a few works in the literature focus on using LLM for MI or CBT, developing these psychotherapy systems, and evaluating them in real-world scenarios. \cite{kian2024can} developed a GPT3.5-powered prompt-engineered socially assistive robot (SAR) that guides participants through interactive CBT at-home exercises. Their findings suggest that SAR-guided LLM-powered CBT may yield comparable effectiveness to traditional worksheet methods. However, this study solely focused on employing an LLM-based approach to simulate traditional worksheet-based CBT, without thoroughly examining the validity of user responses to the CBT exercises or ensuring users effectively engaged with the CBT.

% Although there are a large number of 

%  to evaluate the component functionalities and further discuss the potential of enabling these psychotherapies empowered by LLMs. Only a few focus on developing and deploying these psychotherapy systems and evaluating them in real-world scenarios.

%In particular, they fine-tuned PaLM and PaLM-2 with MultiMedQA corpus, which is a data set combining six existing medical question-answering datasets spanning professional medicine, research and consumer queries and a new dataset of medical questions searched online, and achieved 86.5\% on MedQA dataset~\cite{singhal2023large}.

% has examined GPT-4’s ability to inform us about the latest literature in a given medical area, to write a discharge summary for a patient following an uncomplicated surgery, and to identify objects in photos, and .

\section{System Architecture}\label{sec:System_Architecture}

\begin{figure*}[t!]
    \centering
    \includegraphics[width=0.75\textwidth]{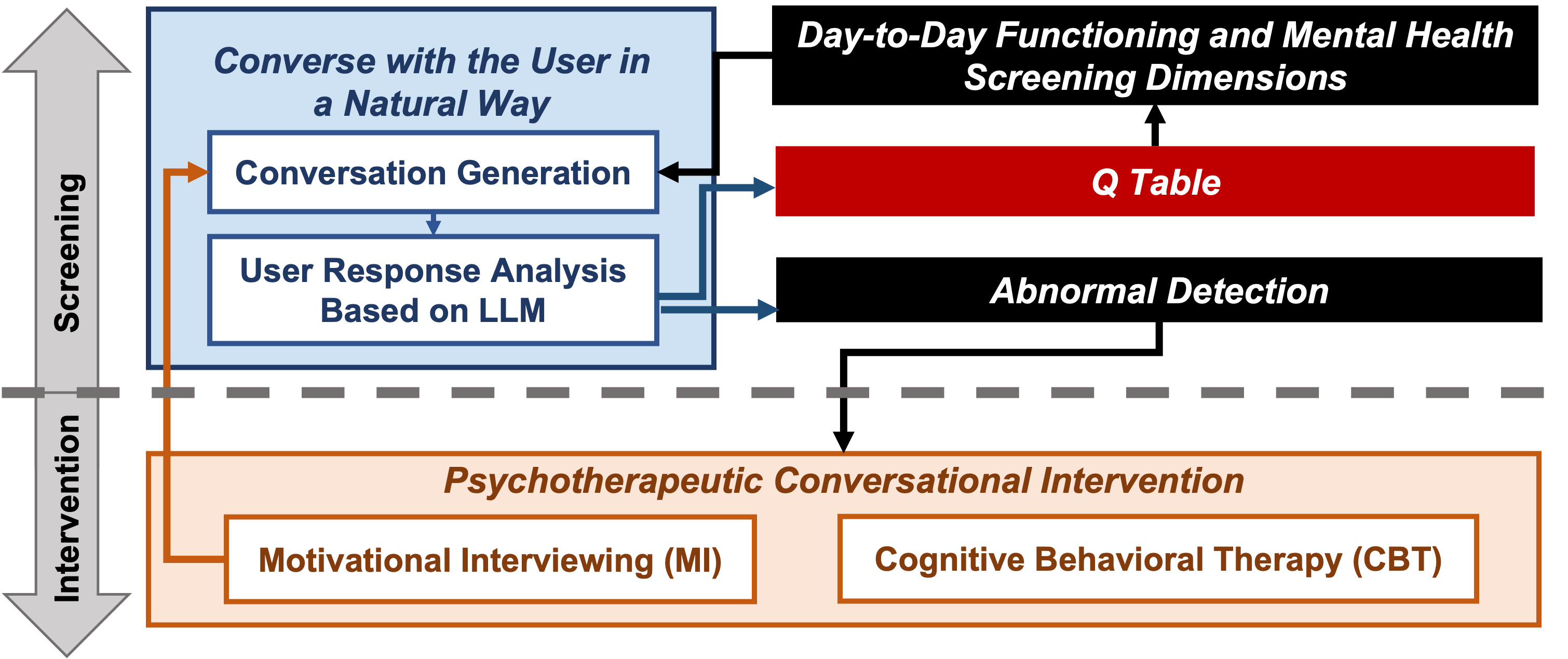}
    \caption{The system architecture of {\sys} consists of two main functionalities: 1) day-to-day functioning screening through natural conversation and 2) precautionary psychotherapeutic conversational interventions.}
    \label{fig:system_architecture}
    \vspace{-2\baselineskip}
\end{figure*}

Considering the design requirements presented in Section~\ref{sec:introduction}, {\sys} includes two main functionalities: \emph{day-to-day functioning screening} and \emph{precautionary psychotherapeutic conversational interventions} as shown in Figure~\ref{fig:system_architecture}. We adopt the 37 dimensions for day-to-day functioning screening proposed in~\cite{nie2022conversational}. For screening, \emph{Converse with the User in a Natural Way} consists of open-ended question generations and semantic analysis of user responses based on LLM. To facilitate \emph{precautionary interventions}, following psychotherapists' guidance, {\sys} effectively integrates the motivational interviewing (MI) and cognitive behavior therapy (CBT) processes into \emph{Psychotherapeutic Conversational Intervention}. Considering the characteristics of MI and CBT, and the actual ways in which the therapists perform during clinical sessions, various MI techniques introduced in Section~\ref{sec:mi_cbt} are applied in different scenarios during the conversation, while the four-step CBT is performed at the end of each conversation session.

\begin{wrapfigure}{r}{0.45\linewidth}
    \vspace{-0.2cm}
    \includegraphics[width=0.45\textwidth]{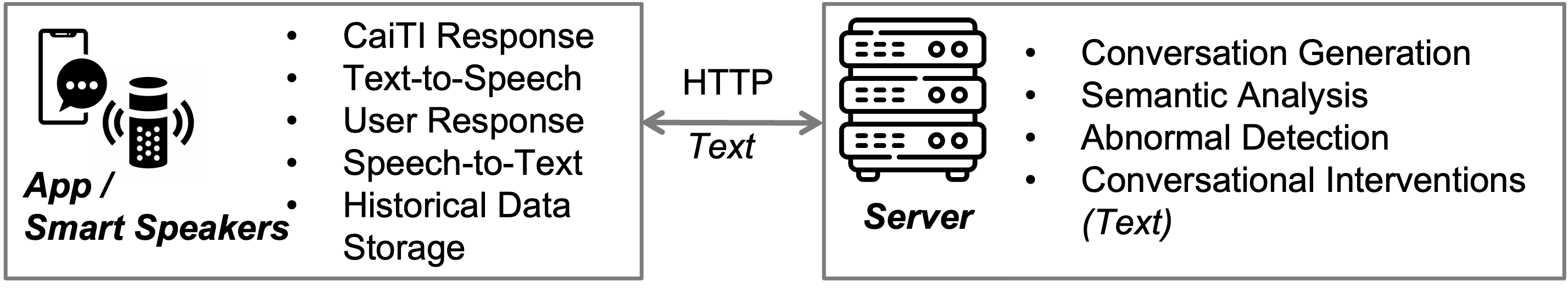}
    \vspace{-0.2cm}
    \caption{The workflow of {\sys}, including the connections between the main components and the locations where different functionalities take place.}
    \label{fig:System_Architecture_Communication}
\end{wrapfigure} 

Each activity screened through conversation sensing results in a \textbf{(Dimension, Score)} pair. Therapists set 3 classes for \textbf{Score} based on their clinical practices (\textbf{Score} $\in \{0, 1, 2\}$), where
(\emph{i}) a score of 0 indicates that the user performs well in this dimension,
(\emph{ii}) a score of 1 indicates that the user has some problems in this dimension, but no immediate action is needed, and
(\emph{iii}) a score of 2 indicates a need for heightened attention from healthcare providers. Figure~\ref{fig:System_Architecture_Communication} shows the flow diagram of {\sys}'s components, and {\sys} stores the historical user data on the front-end devices owned by the user. Due to privacy concerns, {\sys} only conducts semantic analysis on the text of user input, although speech audio is informative~\cite{salekin2017distant}.

\subsection{Conversation Principles and Underlying Rationale}
\label{sec:architecture_chat}

\emph{Converse with the User in a Natural Way} and the \emph{Psychotherapeutic Conversational Intervention} modules of {\sys} are closely related to each other when {\sys} converses with the user. Based on the psychotherapists' experience in dealing with thousands of clients, several factors are considered to shape the conversation process of {\sys}. First of all, when the therapist asks a question, some clients express a lot, while others do not respond to the question, but talk about other things (related to other dimensions). In addition, not all clients are patient enough to go through all dimensions that the therapist wants to check. Psychotherapists usually start to check on the dimensions that the clients didn't do well in previous sessions and are more important for assessment. If clients have a problem in a dimension, the therapists usually follow up to hear more about this dimension and provide quick counseling and therapy addressing the specific issue. This mirrors the psychotherapist's tendency to focus on one problematic dimension extensively rather than treating multiple dimensions at once.

Taking the professional experiences and common practices of the psychotherapists into consideration, to converse with the user in an efficient, intelligent, and natural way to screen physical and mental health status, {\sys}'s conversation process follows four guidelines:
\begin{enumerate}[leftmargin=*]
    \item
    \emph{Prioritize questions intelligently}: {\sys} starts with the dimensions that concern therapists more, while personalizing the priority to each user and formulating questions based on his/her historical responses.
    \item
    \emph{Understand the user input better}: {\sys} checks if the user answers the question asked (\textit{Dimension\_N}), understands how well the user performs in this dimension and decides if follow-up questions and conversational interventions are needed.
    \item
    \emph{Obtain more information through minimal questioning}: {\sys} maps each user response to all possible dimensions to avoid redundant questions. 
    \item
    \emph{Guarantee the quality of psychotherapies}: {\sys} intelligently interprets and reasons the user responses and, when needed, it steers and guides the conversation back to the psychotherapeutic goals if the user's answers stray.
    
\end{enumerate}

\begin{figure*}
    \centering
    \includegraphics[width=0.9\textwidth]{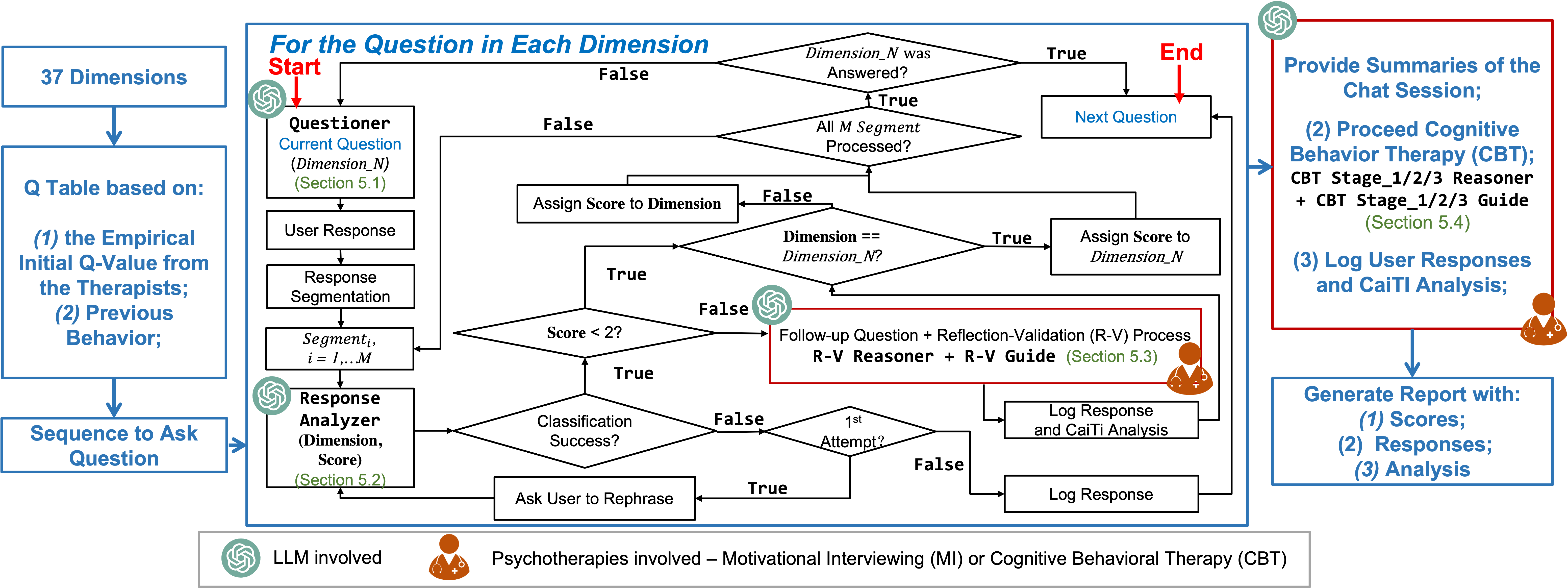}
    \caption{The process to converse with the user naturally to screen day-to-day functioning and provide psychotherapies through the conversation. In particular, \emph{MI} therapies are conducted throughout the conversation, while \emph{CBT} proceeds at the end of the conversation session.}
    \label{fig:system_architecture_chat}
    \vspace{-\baselineskip}
\end{figure*}

\subsection{Conversation Generation, Analysis, and Psychotherapeutic Intervention}
\label{secsec:architecture_communication}

Figure~\ref{fig:system_architecture_chat} shows {\sys}'s process to converse with the user. Generally, \emph{MI} therapies are conducted throughout the conversation, while \emph{CBT} proceeds at the end of the conversation session. There are four modules in which the LLMs are involved: \textbf{\texttt{{\sys} Questioner}}, \textbf{\texttt{Response Analyzer}}, \texttt{\textbf{reflection-validation (R-V)}} process, and \textbf{\texttt{CBT}} process.

In particular, {\sys} asks one question for each dimension if {\sys} does not obtain any information in the dimension from the user's previous responses. A model-free reinforcement learning algorithm, Q-learning, is used to decide the action (i.e., the next question) in the current state (i.e., the current question). For each dimension (\textit{Dimension\_N}),  \textbf{\texttt{{\sys} Questioner}} \emph{formulates} the question and uses the text-to-speech method to converse with the user through the front-end device. The front-end device generates the text of the user response (speech-to-text conversion is used if the user has voice input). The detailed implementation of the front-end device is described in Section~\ref{secsec:implementation}. 

\begin{figure*}
    \centering
    \includegraphics[width=\textwidth]{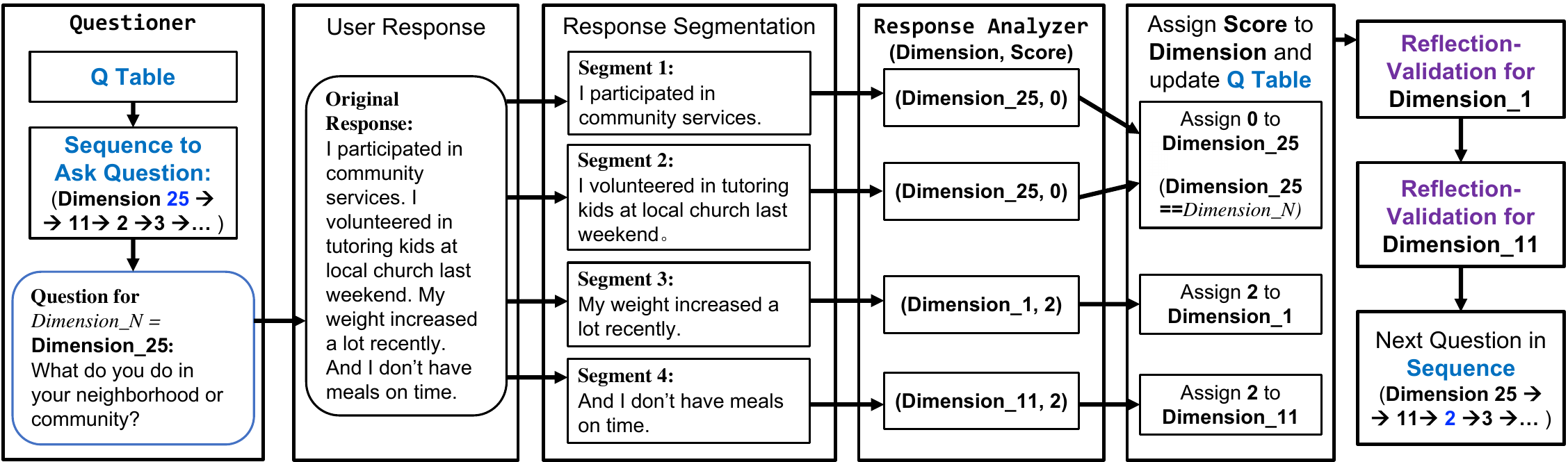}
    \caption{The example process of how {\sys} follows the sequence based on the Q-table, segments the user response, classifies the \emph{Segment} into the format of \textbf{(Dimension, Score)}, and proceeds with reflection-validation (R-V) process.}
    \label{fig:example_process}
    \vspace{-\baselineskip}
\end{figure*}

{\sys} expects the user to chat freely with it and can deal with open-ended responses. {\sys} performs segmentation on the user response in to $M$ \emph{Segment}(s). For each \emph{Segment}, a LLM-based \textbf{\texttt{Response Analyzer}}, described in Section~\ref{secsec:conversation_method_response_analyzer}, is used to classify the \emph{Segment} into \textbf{(Dimension, Score)}. If {\sys} fails to classify the \emph{Segment} into the format of \textbf{(Dimension, Score)}, it asks the user to rephrase the answer. {\sys} logs the user response if {\sys} still fails to classify the rephrased \emph{Segment} into the format of \textbf{(Dimension, Score)}. Otherwise, {\sys} checks the \textbf{Score} no matter if this \emph{Segment} is answering the question asked by {\sys} or not.  If the user needs more attention in this dimension (\textbf{Score} $=2$), {\sys} proceeds with a reflection-validation (R-V) process starting with asking for more information starting with a \emph{simple reflection} in MI. An example of this process is presented in Figure~\ref{fig:example_process}.

The R-V process is described in detail in Section~\ref{secsec:conversation_method_RV_reasoner} and demonstrated in Figure~\ref{fig:MI_reasoner}. To ensure the user provides follow-up information in the right direction, an \textbf{\texttt{R-V Reasoner}} and an \textbf{\texttt{R-V Guide}} are deployed. Based on the user response to the original and follow-up question, {\sys} provides validation, which includes \emph{affective reflection} and \emph{affirmations} in MI. Then, {\sys} assigns the \textbf{Score} to \textbf{Dimension}. After handling all \emph{Segment}(s), {\sys} verifies whether the user does respond to the question asked by {\sys} in \textit{Dimension\_N}. In cases where a user does not answer the question asked by {\sys} (\textit{Dimension\_N}) but talks about something else, {\sys} asks the question in \textit{Dimension\_N} again. 

After {\sys} enumerates all dimensions or the user wants to stop the session, {\sys} provides a summary of the chat session and asks the user to choose a dimension to work on for the CBT process. This CBT process includes the four steps outlined in Section~\ref{sec:mi_cbt}. In particular, {\sys} identifies the situation and issue in the dimension the user chose based on the conversation history. Then, {\sys} leads the user to recognize (\textbf{\texttt{CBT Stage\_1}}), challenge (\textbf{\texttt{CBT Stage\_2}}), and reframe (\textbf{\texttt{CBT Stage\_3}}) the negative thoughts in this situation. To ensure the effectiveness and quality of the CBT process, each CBT stage contains a \textbf{\texttt{Reasoner}} and a \textbf{\texttt{Guide}} (see Section~\ref{secsec: conversation_method_CBT_reasoner}).

\begin{figure*}
    \centering
    \includegraphics[width=\textwidth]{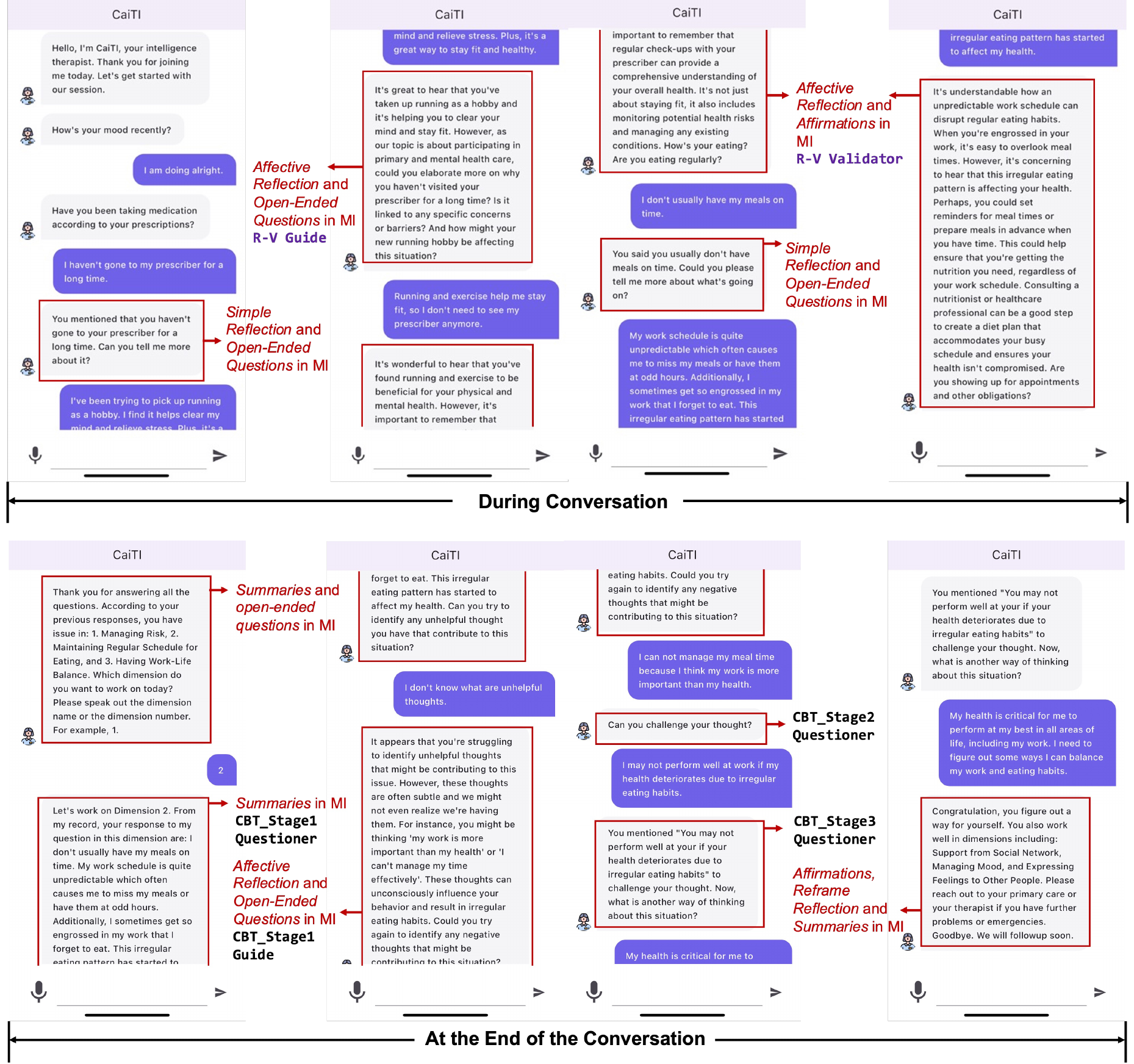}
    \caption{An example conversation using {\sys}'s customized smartphone platform. The MI and CBT embedded in the conversation are annotated.}
    \label{fig:phoneApp_MI_CBT}
    \vspace{-\baselineskip}
\end{figure*}

At the end, {\sys} generates a report that follows the same format as the therapists' notes during their treatment sessions. Appendix~\ref{appendix:37Dimension} reports the details for the 37 dimensions used for day-to-day functioning screening, example questions from {\sys}, and sample responses from the users. Figure~\ref{fig:phoneApp_MI_CBT} shows the smartphone interface for {\sys}'s conversational chatbot, where the various psychotherapeutic interventions applied during different stages of the conversation are annotated. 

\begin{figure}%{r}{0.45\linewidth}
    \vspace{-0.2cm}
    \includegraphics[width=0.25\textwidth]{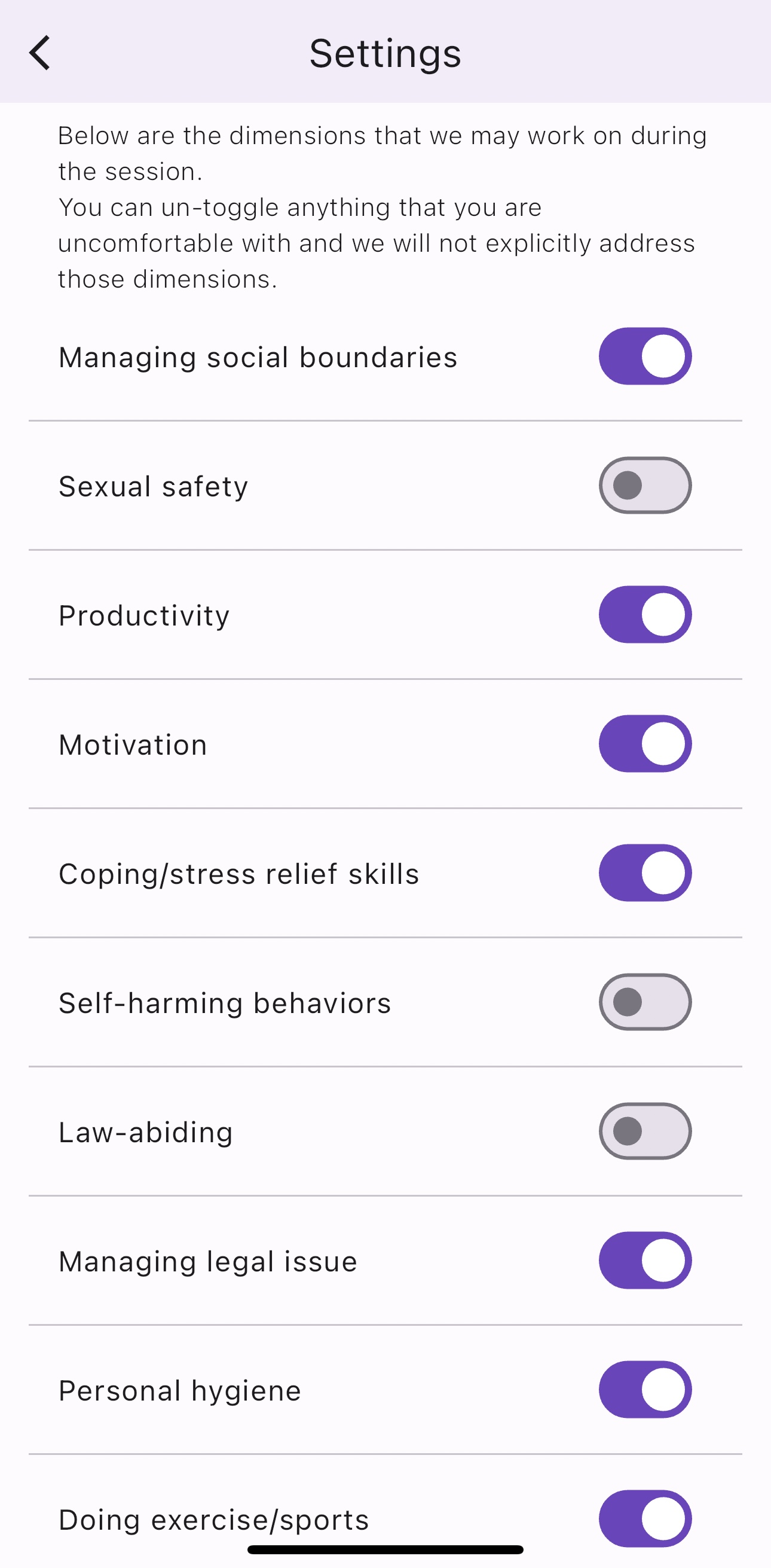}
    \vspace{-0.2cm}
    \caption{The user interface where the user can manually select the dimensions to work on.}
    \label{fig:selection}
\end{figure}

As {\sys} provides comprehensive daily functioning screening, as presented in Appendix~\ref{appendix:37Dimension}, some of the dimensions, such as law-abiding, might be sensitive or uncomfortable for users. Therefore, {\sys} offers the option for users to manually select the dimensions to work on with the smartphone interface shown in Figure~\ref{fig:selection}.

\section{Method and LLM Microbenchmarks}
\label{sec:implementation_and_method}

The methods and LLMs leveraged in: (\emph{i}) {\sys}'s \textbf{\texttt{Questioner}}, (\emph{ii}) \textbf{\texttt{Response Analyzer}}, and (\emph{iii}) task-specific \textbf{\texttt{Reasoner}}s, \textbf{\texttt{Guide}}s, and \textbf{\texttt{Validator}} during the psychotherapies (MI and CBT processes) are introduced in this section. To prevent the propagation of flaws or biases in LLMs, which may lead to ineffective or potentially harmful psychotherapy intervention, instead of leveraging models to handle all tasks during the psychotherapy process, {\sys} divides the tasks and employs different models to specifically handle each subtask. 

Additionally, we present and discuss microbenchmarks comparing different GPT-based methods GPT-4(\texttt{gpt-4})~\cite{openai2023gpt4} and GPT-3.5 Turbo (\texttt{gpt-3.5-turbo-16k})~\cite{openai2023gpt35}) with Llama 2-based methods~\cite{touvron2023llama}. In particular, GPT-4, GPT-3.5 Turbo, Llama-2 13b, and Llama-2 7b possess parameter counts of over a trillion, more than one hundred billion, thirteen billion, and seven billion~\cite{xu2023leveraging}. To carry out the microbenchmarks for LLMs, \emph{datasets were prepared and labeled by four licensed psychotherapists}. These datasets were constructed based on their clinical experiences with clients, ensuring relevance and applicability to real-world therapeutic scenarios. 

Considering the training dataset size for each task provided by the therapists, under the guidance from the therapists, we predominantly use \emph{few-shot} prompting the \texttt{system content} in the \emph{chat completion} in these LLMs to achieve the desired functions. Each prompt outlines: (\emph{i}) the objectives; (\emph{ii}) the information to be included in \texttt{user content}; and (\emph{iii}) the desired goal and response format. The response format for \textbf{\texttt{Reasoner}}s will be \emph{``Decision: 0/1''}, while it is \emph{``Analysis: XXX''} for \textbf{\texttt{Guide}}s and \textbf{\texttt{Validator}}. For \textbf{\texttt{Reasoner}}s, \textbf{\texttt{Guide}}s, and \textbf{\texttt{Validator}}, the prompt includes 3-4 examples encompassing \texttt{user content} alongside corresponding system responses that adhere to the specified format. The examples for \textbf{\texttt{Response Analyzer}} is slightly different and illustrated in Section~\ref{secsec:conversation_method_response_analyzer}. We set the temperature as 0.7 in the LLMs to achieve varied rephrasings of the questions while maintaining certain constraints. The same prompt and hyperparameters are used for different LLMs. We fine-tune a GPT-3.5 Turbo model for \textbf{\texttt{Response Analyzer}}. The therapist also labeled and analyzed the output of \textbf{\texttt{Guide}}s and \textbf{\texttt{Validator}}. \emph{We will open source the prompts we constructed as well as the datasets constructed by the psychotherapists.} We did not conduct microbenchmark tests on basic LLM tasks such as the \texttt{Rephraser} and \texttt{ReflectiveSummarizer}. The former task involves structural rather than semantic rephrasing, while the latter repeats and converts statements from the first person to the third person.

\subsection{{\sys}'s \textbf{\texttt{Questioner}}}\label{secsec:conversation_method_question_generation}

{\sys}'s \textbf{\texttt{Questioner}} drives the conversations based on Epsilon-Greedy Q-learning and a GPT-based ``\texttt{Rephraser}''. To make the conversation more natural, psychotherapists provide a set of questions they typically ask in each dimension (7 to 11 sample questions). We prompt a GPT-4-based \texttt{Rephraser}~\cite{openai2023gpt4} to rephrase these questions (structurally instead of semantically) when asking questions.

Each dimension has one related question. The Q-learning agent has 39 states (37 questions, start, and end). We set the learning rate and discount factor to 0.1 and 0.9, respectively. The probability of selecting the best action is set as $\epsilon=0.9$. The therapists determine the initial Q-values for the Q-table based on their empirical evaluation of the ``importance'' of the dimensions. The Q-value, represents the expected future rewards that can be obtained by taking a given action (next question) in a given state (current question). The \textbf{Score} based on the analysis of user responses is the reward earned in that state.

\subsection{\textbf{\texttt{Response Analyzer}}}
\label{secsec:conversation_method_response_analyzer}

When the user responds to the question asked by {\sys}, {\sys} first segments the response into individual sentences. For each segmented sentence, {\sys} classifies it into \textbf{(Dimension, Score)}, where there are 37 dimensions and 3 scores (\textbf{Score} $\in \{0, 1, 2\}$) -- a total number of 111 classes. In addition, we define 5-class general responses to express \emph{Yes, No, Maybe, Question, and Stop} (e.g., ``Yes'', ``I don't know'', ``Stop'', ``Maybe'', and ``I don't understand your question'') as well as a mapping table between the \textbf{Score}s and general responses for each dimension. For example, the \textbf{Score} of ``Yes'' is 0 to the question \emph{``Are you showing up for work or school?''} in \emph{Managing Work/School}, while it is 2 to the question \emph{``Do you often drink alone?''} in \emph{Alcohol Abuse}.

Since {\sys} asks open-ended questions, user responses are infinitely diverse (YES/NO answers or open-ended responses). With a \textbf{Score} of 2, {\sys} will conduct the psychotherapeutic conversational intervention. Otherwise, {\sys} will ask the next question based on the Q table. When {\sys} meets out-of-context responses, it asks the user to rephrase and follow the process illustrated in Figure ~\ref{fig:system_architecture_chat}. 

\subsubsection{Microbenchmark -- \textbf{\texttt{Response Analyzer}}}
%User responses can vary to an unlimited extent.
To the best of our knowledge, no dataset exists with responses to these questions in the 37 dimensions. Therefore, psychotherapists create a dataset, which includes: (\emph{i}) 6,950 user responses sample with the \textbf{(Dimension, Score)} labeled by the therapists, and (\emph{ii}) 300 5-class general responses to express \emph{Yes, No, Maybe, Question, and Stop}. Note that one user response may have one or more \textbf{(Dimension, Score)}. As such, there are 7,000 \textbf{(Dimension, Score)} for the 6,950 responses. The number of responses per dimension is 103 to 177.

The datasets are split into 90\% and 10\% for training and testing set to fine-tune and evaluate the GPT-3.5 Turbo model (\texttt{gpt-3.5-turbo} is recommended by OpenAI for fine-tuning task~\cite{gpt35finetune}). For a fair comparison, the same testing set is used to evaluate the performance of the LLMs with \emph{few-shot} prompt. Therapists select 2-5 examples for each score in each dimension from the training set to construct the \emph{few-shot} \texttt{system content} prompt. The constructed prompt contains around 8,140 tokens. As shown in Table~\ref{tab:response_analyzer}, the prompt GPT-4 model has comparable performance to the fine-tuned GPT-3.5-Turbo model. Llama-based models perform less ideal in correctly classifying the \textbf{Dimension} and \textbf{Score}. In particular, Llama-based models have issues identifying the responses that don't need further attention (\textbf{Score} = 0 and 1).

\begin{table*}[t!]
\caption{\textbf{Score} and \textbf{Dimension} assignment performance comparisons for \textbf{\texttt{Response Analyzer}} using different LLM models.}
\vspace{-\baselineskip}
\begin{tabularx}{\textwidth}{l|X|X|X}
\hline
    
    & \textbf{Score Accuracy} & \textbf{Dimension Accuracy} & \textbf{General Response Accuracy}\\ \hline\hline
    \textbf{Fine-tune GPT-3.5-Turbo} & 95.58\% & 96\% & 93.75\%\\\hline
    \textbf{Prompt GPT-4} & 94.43\% & 95.14\% & 95.85\%\\\hline
    \textbf{Prompt GPT-3.5 Turbo} & 92.81\% & 94.14\% & 96.14\%\\\hline
    \textbf{Prompt Llama-2-13b} & 54.39\% & 59.42\% & 63.33\%\\\hline
    \textbf{Prompt Llama-2-7b} & 55.11\% & 48.63\% & 53.33\%\\\hline

\end{tabularx}
\label{tab:response_analyzer}
\end{table*}

\subsection{Reflection-Validation \texttt{Reasoner}, \texttt{Guide}, and \texttt{Validator}}\label{secsec:conversation_method_RV_reasoner}

As depicted in Section~\ref{secsec:architecture_communication} and Figure~\ref{fig:system_architecture_chat}, during the conversation screening process, if the user's response includes situations that require further attention (with \textbf{Score} = 2), {\sys} will pose a follow-up question to gather more information about the contributing factors to this situation. The follow-up question would start with the \emph{simple reflection} in MI, a technique where the psychotherapist or counselor mirrors what the client has said. A GPT-4-based \texttt{ReflectiveSummarizer} is prompted to provide the \emph{simple reflection}, which essentially rephrases or repeats the client's own words, altering any self-references from the first person to the third person ~\cite{openai2023gpt4}.  

\begin{figure*}
    \centering
    \includegraphics[width=0.95\textwidth]{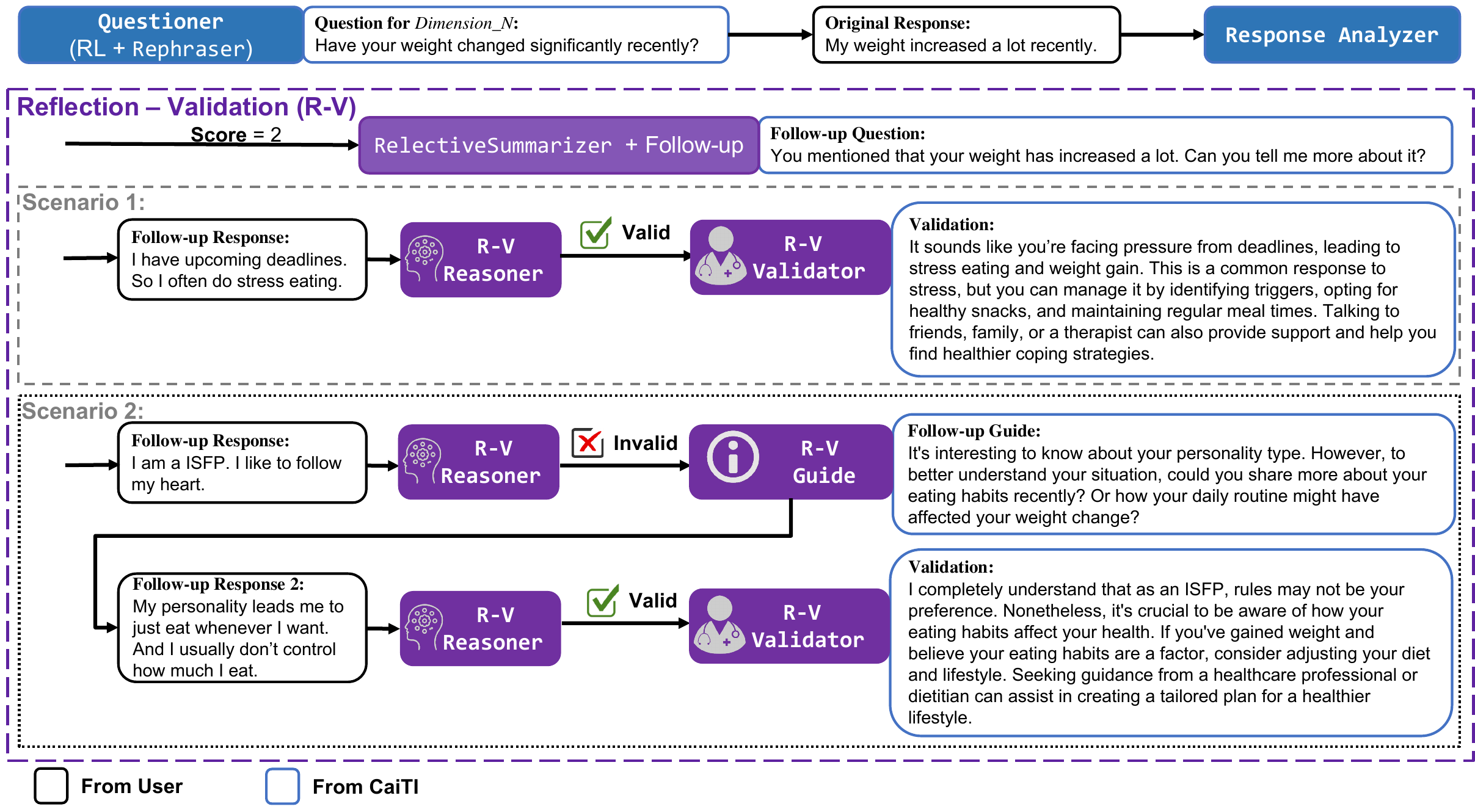}
    \vspace{-\baselineskip}
    \caption{The reflection-validation (R-V) pipeline illustrates how the \emph{R-V Reasoner} is utilized for reasoning when the user provides a valid follow-up response that necessitates further attention, including two example scenarios: Scenario 1, where the user provides a valid follow-up response on the first attempt; and Scenario 2, where the user initially fails to provide a valid follow-up response, and {\sys} guides the user towards providing valid follow-up responses.}
    \label{fig:MI_reasoner}
    \vspace{-\baselineskip}
\end{figure*}

The follow-up question initiates a reflection-validation (R-V) process to provide effective empathic validation and support. Given that follow-up responses from the user are open-ended and have infinite possibilities, {\sys} incorporates a \texttt{\textbf{R-V Reasoner}} to determine whether the follow-up response is related to the original response or the question asked in the current state. As illustrated in Scenario 1 in Figure~\ref{fig:MI_reasoner}, {\sys} will offer empathic validation if the user provides a valid follow-up response. Otherwise, the \texttt{\textbf{R-V Guide}} will assist the user in providing a follow-up response that more accurately describes the situation at hand before proceeding to empathic validation (Scenario 2 in Figure~\ref{fig:MI_reasoner}).

With a valid follow-up response, a \texttt{\textbf{R-V Validator}} is used to provide empathic validation and support to the user, which incorporates the \emph{affective reflection} and \emph{affirmation} techniques in MI. These two techniques aim at demonstrating empathy and understanding of the client's feelings, creating a supportive environment that validates the client's experiences and feelings, and fostering a deeper connection.

\subsubsection{Microbenchmark -- Reflection-Validation \texttt{Reasoner}, \texttt{Guide}, and \texttt{Validator}}
Therapists provide us with 593 example follow-up responses for the original responses that need further attention (\textbf{Score} = 2). Within these 593 responses, 244 follow-up responses are invalid (1: Invalid), which trigger \texttt{\textbf{R-V Guide}}s before prompting the \texttt{\textbf{R-V Validator}}s. In addition, 6 examples for \texttt{\textbf{R-V Guide}}s and \texttt{\textbf{R-V Validator}}s are given by the psychotherapists to construct the prompts. As the output of \textbf{\texttt{R-V Guide}} and \textbf{\texttt{R-V Validator}} are open-ended, therapists label the 244 \textbf{\texttt{R-V Guide}}s and 593 \textbf{\texttt{R-V Validator}}s generated by each of the 4 prompt-based LLM models.

Table~\ref{tab:RV_res} presents the performance metrics of \textbf{\texttt{R-V Reasoner}}, \textbf{\texttt{R-V Guide}}, and \textbf{\texttt{R-V Validator}} across the four LLM models, where we prompted the \texttt{system content} in the \emph{chat completion} functionality. The data presented in Table~\ref{tab:RV_res} illustrate that methods based on Llama demonstrate inferior reasoning capabilities in assessing the validity of a follow-up response in relation to the original question and response within the given context. Typically, Llama-based approaches are prone to inaccurately classifying valid follow-up responses as invalid. Additionally, the GPT-based method demonstrates a remarkable capability in accurately identifying all invalid responses, evidenced by its complete absence of false-negative errors (the misclassification of invalid responses as valid). 

\textbf{\texttt{R-V Guide}}s generated by GPT-based methods also outperform those generated by Llama-based methods. Therapists point out that the major issue for Llama-based \texttt{Guide}s is they sometimes get confused or distracted by the irrelevant follow-up answer and reorient from the original dimension to ask about contents mentioned in the follow-up. This phenomenon is more obvious in the \texttt{Guide}s generated by Llama-2-13b. Therapists also observe that Llama-based \texttt{Guide}s cannot capture all the points in the user response, often focusing only on one aspect. 

Moreover, the empathic validations generated by the Llama-based \textbf{\texttt{R-V Validator}}s are substandard in quality control. In particular, Llama-based \textbf{\texttt{R-V Validator}}s perform poorly in following the objectives and format specified in the few-shot prompt. Some outputs included excessive small talk, some focused on follow-up responses that were irrelevant to the target dimension that was the primary focus of the discussion, and some failed to provide reflection or affirmation. For example, one output generated by Llama-2-13b \textbf{\texttt{R-V Validator}} engaged in questions aiming to problem solve based on the user's responses, rather than offering empathetic validation as intended, which deviates significantly from the expected function of providing empathetic support.  As indicated in Table~\ref{tab:RV_res}, GPT-based \textbf{\texttt{R-V Validator}}s deliver more regulated and superior quality empathic validations, largely conforming to the prompt's requirements. However, a minor number of empathic validations by GPT-based \textbf{\texttt{R-V Validator}}s encounter issues such as the use of inappropriate words or tones, or overinterpreting the feeings and emotions in user responses.

\begin{table*}[t!]
\caption{ The performance for \textbf{\texttt{R-V Reasoner}} (0: Valid, 1: Invalid), \textbf{\texttt{R-V Guide}}, and \textbf{\texttt{R-V Validator}} with different LLM models.}
\vspace{-\baselineskip}
\begin{tabularx}{\textwidth}{l|XXX|X|X}
\hline
    \textbf{Prompted} & \multicolumn{3}{c|}{\textbf{\texttt{R-V Reasoner}}} & \textbf{\texttt{R-V Guide}} & \textbf{\texttt{R-V Validator}}\\
    \textbf{Models} & \small{\textbf{Accuracy}} & \small{\textbf{Precision}} & \small{\textbf{Recall}} & \small{\textbf{Accuracy}}& \small{\textbf{Accuracy}}\\ \hline\hline
\textbf{GPT-4} & 97.8\% & 94.94\% & 100\% & 94.67\% & 96.79\%  \\\hline
\textbf{GPT-3.5 Turbo} & 97.3\% & 93.85\% & 100\% & 95.08\% & 95.95\%  \\\hline
\textbf{Llama-2-13b} & 69.3\% & 68.7\% & 84.5\% & 68.85\% & 83.47\%  \\\hline
\textbf{Llama-2-7b} &74.54\% & 65.76\% & 70.50\% & 75.00 \% & 63.52\%  \\\hline
\end{tabularx}
\label{tab:RV_res}
\vspace{-\baselineskip}
\end{table*}

\subsection{Cognitive Behavioral Therapy \texttt{Reasoner} and \texttt{Guide}}
\label{secsec: conversation_method_CBT_reasoner}

As described in Section~\ref{sec:mi_cbt} and Section~\ref{secsec:architecture_communication}, the CBT process usually includes four steps and {\sys} completes the first step -- identifying the situation and issues -- for the user based on the historical user responses in the current conversation session. As such, there are three stages remaining: recognizing the negative thoughts (\texttt{CBT\_Stage1}), challenging the negative thoughts (\texttt{CBT\_Stage2}), and reframing the thoughts and the situations (\texttt{CBT\_Stage3}). As the three stages target at different objectives, there are a \emph{Questioner}, \emph{Reasoner}, and \emph{Guide} in each stage. 

\begin{figure*}
    \centering
    \includegraphics[width=\textwidth]{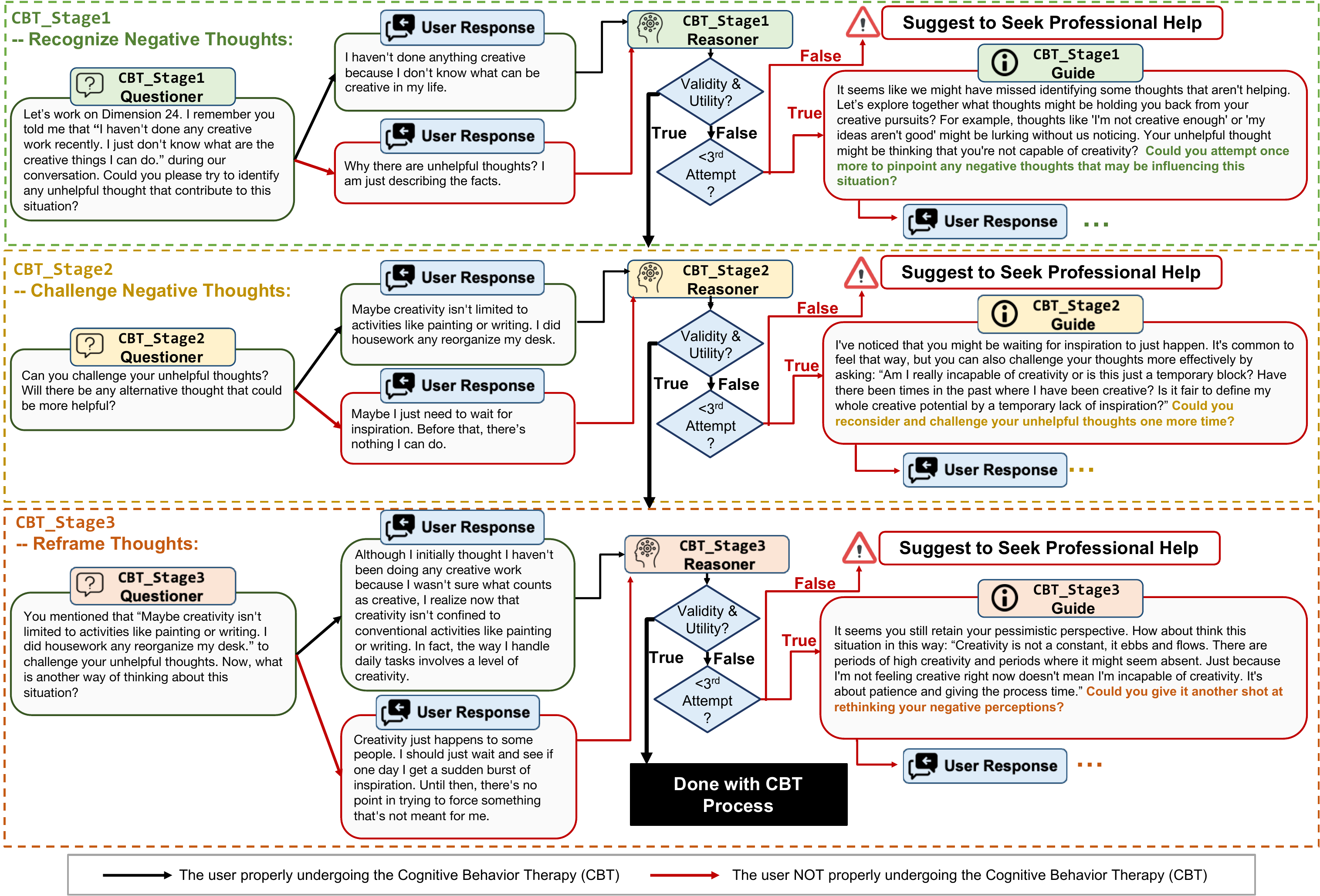}
    \vspace{-\baselineskip}
    \caption{The pipeline of the CBT process, which includes \textbf{\texttt{Questioner}}s, \textbf{\texttt{Reasoner}}s and \textbf{\texttt{Guide}}s. The CBT process in {\sys} contains three stages: \texttt{CBT\_Stage1} -- Recognize Negative Thoughts, \texttt{CBT\_Stage2} -- Challenge Negative Thoughts, and \texttt{CBT\_Stage3} -- Reframe Thoughts. The example scenarios where the user properly undergoes and fails to undergo each CBT stage are presented. }
    \label{fig:CBT_reasoner_guide}
    \vspace{-\baselineskip}
\end{figure*}

Figure~\ref{fig:CBT_reasoner_guide} shows the pipeline for the three-stage CBT process. In each stage, the \textbf{\texttt{Questioner}} poses questions aligned with that stage's specific objectives. The \textbf{\texttt{Reasoner}} then evaluates the user's response for its validity and utility. {\sys} progresses to the next stage if the response is deemed valid. If not, the \textbf{\texttt{Guide}} aids the user in crafting a valid response by identifying issues in the current response and suggesting possible improvements for a valid answer. If the user does not provide a valid response after two attempts, {\sys} will conclude the CBT process and recommend that the user seek professional assistance for a more effective and valid CBT experience.

Therapists also point out that an acceptable response involves identifications of cognitive distortion, such as polarized thinking, overgeneralization, emotional reasoning, catastrophizing, and jumping to conclusions. The \textbf{\texttt{Reasoner}} is tasked with recognizing responses containing cognitive distortions as valid, especially for \textbf{\texttt{CBT\_Stage1 Reasoner}}. Meanwhile, if the response with cognitive distortions is invalid (e.g., not relevant to the situation), the \textbf{\texttt{Guide}} must take these distortions into account when assisting the user in formulating a valid response. 

\subsubsection{Microbenchmark -- Cognitive Behavioral Therapy \texttt{Reasoner} and \texttt{Guide}}\label{secsec: conversation_method_CBT_reasoner_micro}

Psychotherapists provide us with 146 example situations within the 37 dimensions and the three-stage CBT responses for each situation. In addition, they label the validity of each response (0: Valid, 1: Invalid). In particular, there are 15, 17, and 33 responses that are invalid and need ``Guide'' in \texttt{CBT\_Stage1}, \texttt{CBT\_Stage2}, and \texttt{CBT\_Stage3}, respectively. With guidance from therapists, we developed \texttt{system content} prompts to evaluate whether users can identify, challenge, and reframe negative thoughts effectively for \textbf{\texttt{CBT\_Stage1 Reasoner}}, \textbf{\texttt{CBT\_Stage2 Reasoner}}, and \textbf{\texttt{CBT\_Stage3 Reasoner}}. And the performances of these \textbf{\texttt{CBT Reasoner}}s using different LLMs are listed in Table~\ref{tab:CBT_reasoner}.  The \texttt{user content} for each stage encompasses all responses from the user and {\sys} within the current stage as well as from preceding stages. Overall, the performance for \textbf{\texttt{CBT\_Stage3}} is higher. GPT-3.5 Turbo has comparable performance to GPT-4, while both GPT-based models significantly outperform the Llama-2-based models in reasoning. 

\begin{table*}[t!]

\caption{\textbf{\texttt{CBT Reasoner}} performance for using different LLM models (0: Valid, 1: Invalid).}
\vspace{-\baselineskip}
\begin{tabularx}{\textwidth}{l|XXX|XXX|XXX}
\hline
    \textbf{Prompted} & \multicolumn{3}{c|}{\textbf{\texttt{CBT\_Stage1 Reasoner}}} & \multicolumn{3}{c|}{\textbf{\texttt{CBT\_Stage2 Reasoner}}} & \multicolumn{3}{c}{\textbf{\texttt{CBT\_Stage3 Reasoner}}} \\
    \textbf{Models} & \small{\textbf{Accuracy}} & \small{\textbf{Precision}} & \small{\textbf{Recall}} & \small{\textbf{Accuracy}} & \small{\textbf{Precision}} & \small{\textbf{Recall}} & \small{\textbf{Accuracy}} & \small{\textbf{Precision}} & \small{\textbf{Recall}} \\ \hline\hline
\textbf{GPT-4} & 95.89\% & 72.68\% & 93.33\% & 95.21\% & 72\% & 100\% & 99.32\% & 100\% & 96.97\% \\\hline
\textbf{GPT-3.5 Turbo} & 95.21\% & 83.33\% & 66.67\% & 94.52\% &  90.91\% & 58.82\% & 95.89\% & 100\% & 82.35\% \\\hline
\textbf{Llama-2-13b} & 67.81\% & 23.33\% & 93.33\% & 91.1\% & 59.09\% & 76.47\% & 96.58\% & 96.67\% & 87.88\% \\\hline
\textbf{Llama-2-7b} & 69.86\% & 10.81\% & 26.67\% & 88.36\% & 50\% & 29.41\% & 81.51\% & 58.33\% & 63.64\% \\\hline

\end{tabularx}
\label{tab:CBT_reasoner}
\end{table*}

As mentioned before, \textbf{\texttt{CBT Guide}} needs to be aware of cognitive distortions and point them out to guide effectively. According to psychotherapists' experiences, identifying unhelpful thoughts is the most challenging stage within the three stages of the CBT process, with the highest probability of containing cognitive distortions and less information obtained from the user in this stage. Table~\ref{tab:CBT_guide} shows the performance of different LLMs in providing appropriate \textbf{\texttt{CBT Guide}} when responses are invalid in the three CBT stages. In general, using the same prompt, GPT-3.5 Turbo performs the best, even outperforming GPT-4. Therapists comment that \emph{``GPT-4 sometimes sounds like it is reading into the user's feelings''} instead of guiding the user objectively. Additionally, Llama-2-13b and Llama-2-7b show fairly good performance in \textbf{\texttt{CBT\_Stage2}} and \textbf{\texttt{CBT\_Stage3}}, with very few instances of ``overanalyzing'' or ``reading into emotions''. However, as previously mentioned, Llama models have comparably poor reasoning capabilities and struggle with processing more complex information, which might contribute to their inferior performance in \textbf{\texttt{CBT\_Stage1}}.

\begin{table*}[t!]
\caption{\textbf{\texttt{CBT Guide}} accuracies for using different LLM models.}
\vspace{-\baselineskip}
\begin{tabularx}{\textwidth}{l|X|X|X}
\hline
    \textbf{Prompted Models} & \textbf{\texttt{CBT\_Stage1 Guide}} & \textbf{\texttt{CBT\_Stage2 Guide}} & \textbf{\texttt{CBT\_Stage3 Guide}} \\ \hline\hline
\textbf{GPT-4} & 93.33\% & 94.12\% & 90.9\% \\\hline
\textbf{GPT-3.5 Turbo} & 100\% & 100\% & 96.97\% \\\hline
\textbf{Llama-2-13b} & 73.33\% & 94.12\% & 100\% \\\hline
\textbf{Llama-2-7b} & 73.33\% & 100\% & 93.93\% \\\hline

\end{tabularx}
\label{tab:CBT_guide}
\end{table*}

\subsection{Discussion and Observations on different LLMs}
As illustrated in Section~\ref{secsec:conversation_method_response_analyzer} to~\ref{secsec: conversation_method_CBT_reasoner}, overall, GPT-based models have higher performance compared to Llama-based models in achieving the desired functionalities in {\sys} and following the instructions specified in the few-shot prompts, especially in complex tasks, such as: (\emph{i}) \textbf{\texttt{Response Analyzer}}, which needs to classify the user response into \textbf{(Dimension, Score)} (5 general response classes, 37 dimensions, and 3 scores), and (\emph{ii}) different \texttt{\textbf{Reasonser}}s, \texttt{\textbf{R-V Guide}} and \texttt{\textbf{R-V Validator}}, where the user responses may span infinitely diverse spectrum and may not fall within the current conversation topic. Llama-based models had a hard time following the instructions in the few-shot prompts when the expressions from the user lacked logical consistency and with cognitive distortions.

Moreover, GPT-based models sometimes add their own interpretation of users' feelings instead of providing an objective, matter-of-fact output based on the user responses. Llama-based models with few-shot prompts have more stable performance for \textbf{\texttt{CBT\_Stage2 Guide}} and \textbf{\texttt{CBT\_Stage3 Guide}}, where the user responses are more standard and controlled thanks to the filtering of \textbf{\texttt{CBT Reasoner}}s and the tasks, challenging and reframing the negative thoughts, are more straight forward. These observations are expected, as GPT-based models have larger parameters and are pre-trained on a larger corpus of data. Indeed, there are methods that can further improve the quality of the outputs generated by these LLM models, which will be further discussed in Section~\ref{sec:future}.

\section{Implementation and Study Design}
\label{sec:study_design}

In this section, we outline the subject recruitment procedures, describe the implementation, and detail the study design.

\subsection{Subject Recruitment}
20 subjects voluntarily participated (received informed consent from each subject) in our study (approved by the Institutional Review Board), including 10 men and 10 women between 18 and 40 years old from different races. All participants reported having normal hearing and cognition with no history of serious mental or physical illness. All subjects were either students or employed. Each subject was assigned a random subject ID (e.g., S01) for data identification.

\subsection{LLMs Implementation} \label{secsec:LLM-implementation}

Considering the response time and computational resources requirement, the best performing LLM according to the microbenchmarks is implemented for different modules, as illustrated in Figure~\ref{fig:CBT_reasoner_guide}. As discussed in Section~\ref{sec:implementation_and_method}, different LLMs excel at handling different tasks in \sys{}'s conversation flow. For example, GPT-3.5 turbo is more suitable for providing guide and empathic validation, while GPT-4 is better at reasoning the validity and utility of user response.

\begin{figure*}
    \centering
    \includegraphics[width=0.6\textwidth]{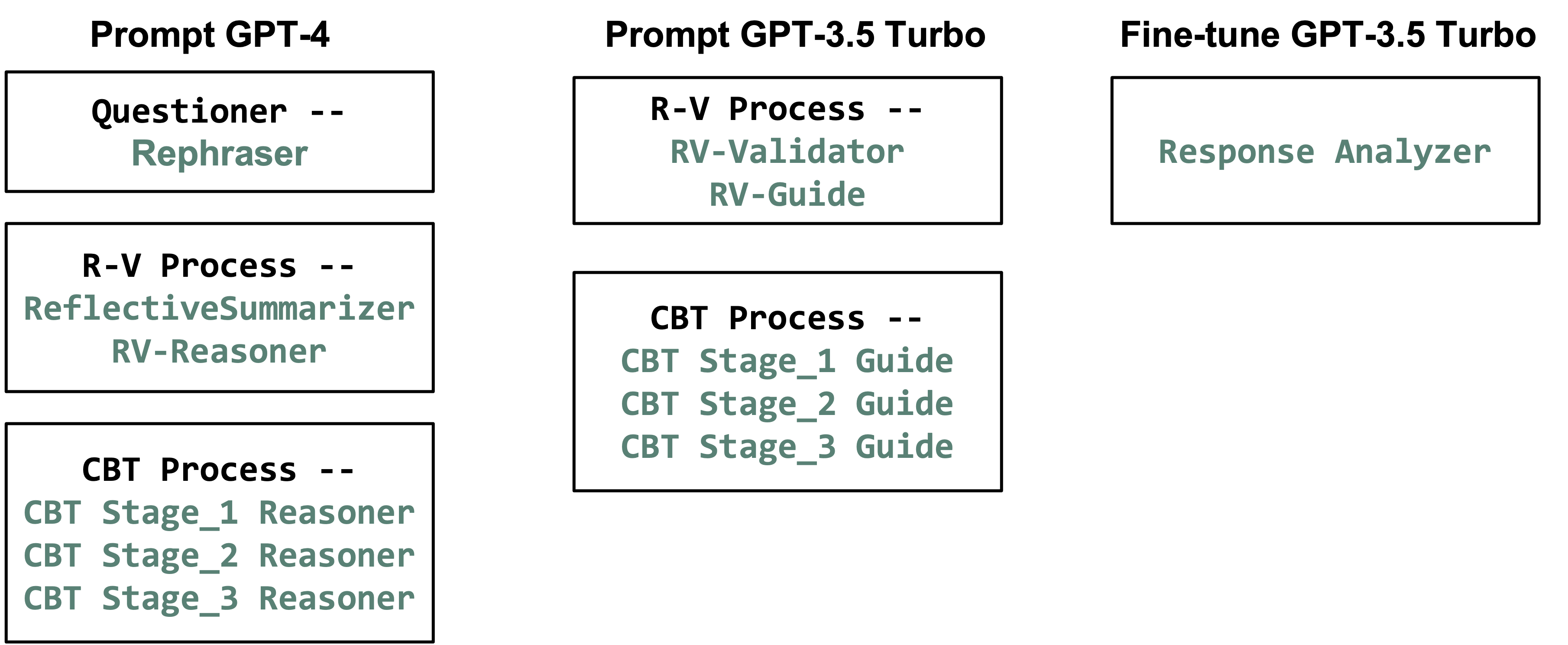}
    \caption{The LLM implementation for different tasks during the study.}
    \label{fig:LLM_implementation}
    \vspace{-\baselineskip}
\end{figure*}

\subsection{Implementation} \label{secsec:implementation}
As mentioned in Section~\ref{sec:introduction}, to accommodate individuals with different needs and with possible memory or vision impairments, particularly the elderly, \sys{} is available in two physical form factors: a customized multi-platform app and a smart speaker (Amazon Alexa). \sys{} integrates custom Alexa Skills with an Amazon Echo device and a Flutter-based application, enabling flexible interaction through voice or text on multiple platforms such as Android, iOS, Windows, and macOS.

Figure~\ref{fig:System_Architecture_Communication} depicts the System's architecture, highlighting the communication pathway where only the conversation text is transmitted between the user interfaces and the server. This design decision is implemented to mitigate the risk of compromising sensitive information that may be inherent in voice data. Voice interactions are facilitated by APIs such as the Alexa Skills Kit~\cite{AlexaSkills} and Google's speech-to-text API~\cite{Speech-to-text}. The server's role is to handle all LLM-based tasks, including interpreting text inputs and generating appropriate follow-up questions or psychotherapeutic interventions that are then delivered to the user interface.

\begin{figure*}[t!]
     \centering
    \begin{subfigure}[b]{0.2\textwidth}
         \centering
         \includegraphics[height=4cm]{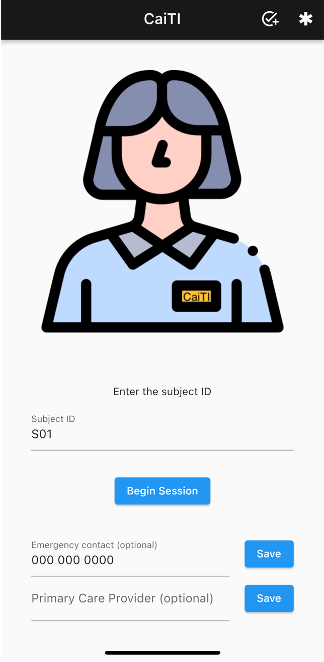}
    \caption{}
    \label{fig:SmartphoneHome}
     \end{subfigure}
     \hspace{2mm}
     \begin{subfigure}[b]{0.23\textwidth}
         \centering
         \includegraphics[height=4cm]{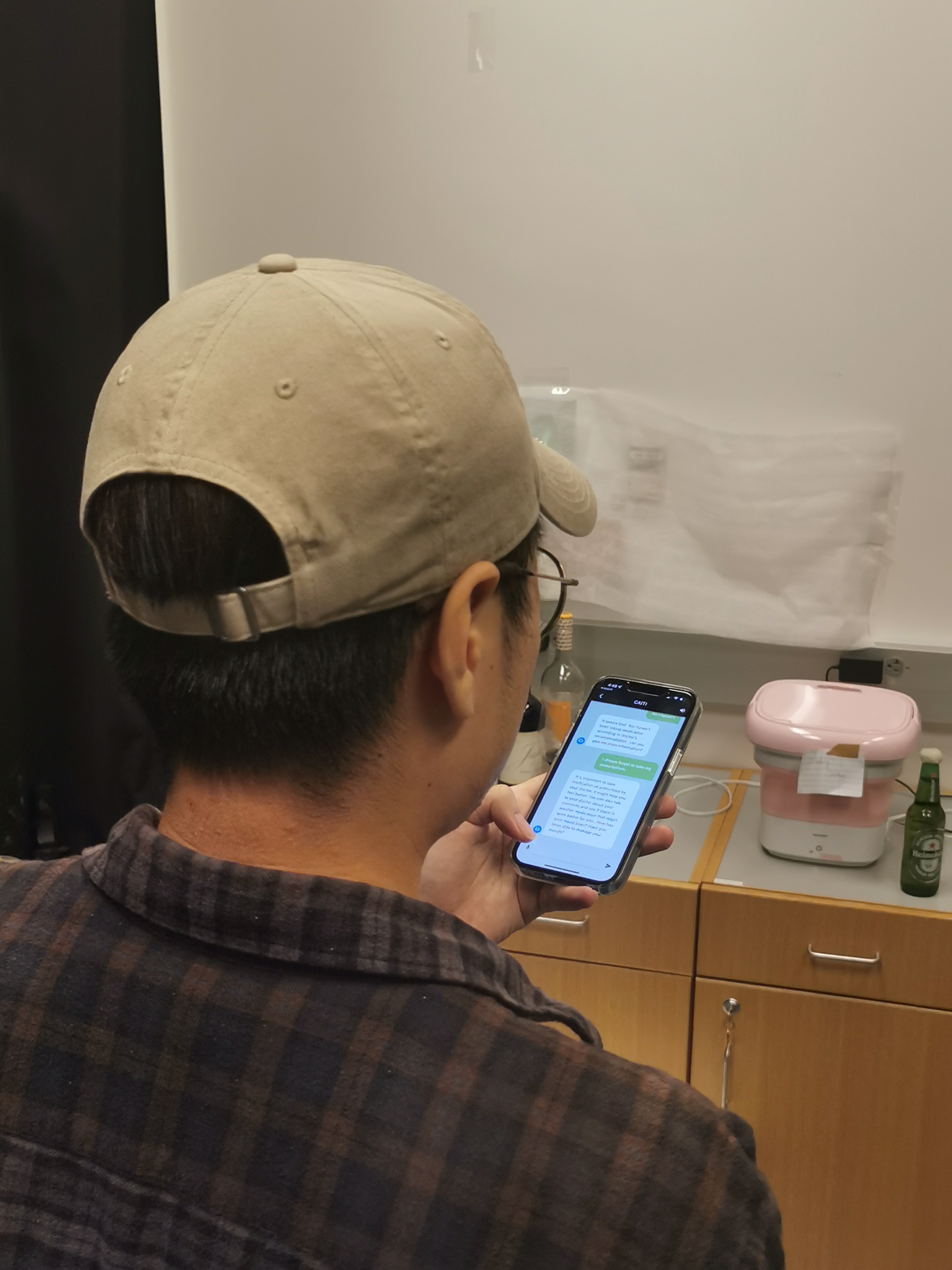}
    \caption{}
    \label{fig:Human-Interaction}
     \end{subfigure}
     \hspace{2mm}
     \begin{subfigure}[b]{0.23\textwidth}
         \centering
         \includegraphics[height=4cm]{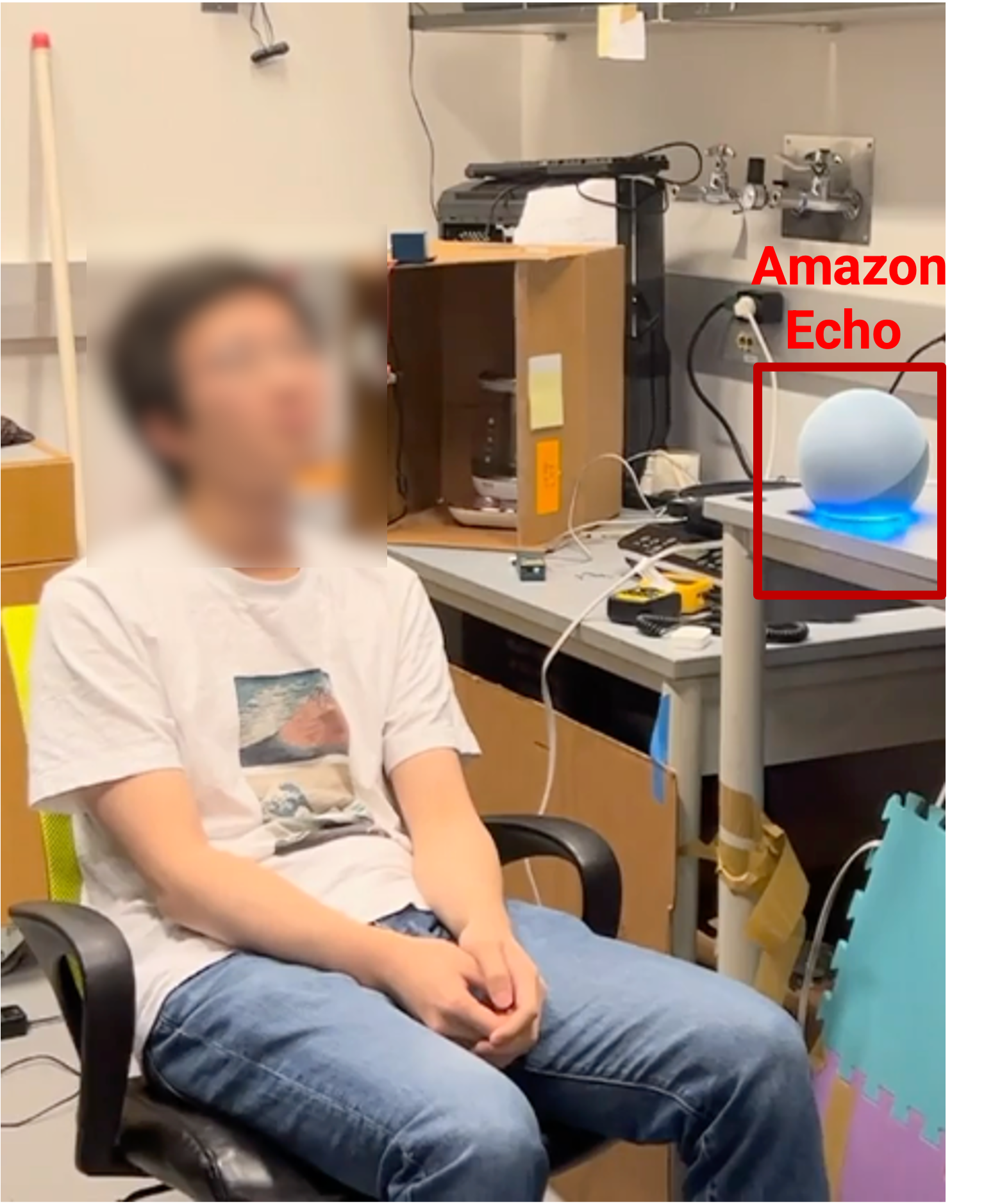}
    \caption{}
    \label{fig:Human-Interaction1}
     \end{subfigure}
     \hspace{2mm}
     \begin{subfigure}[b]{0.2\textwidth}
         \centering
         \includegraphics[height=4cm]{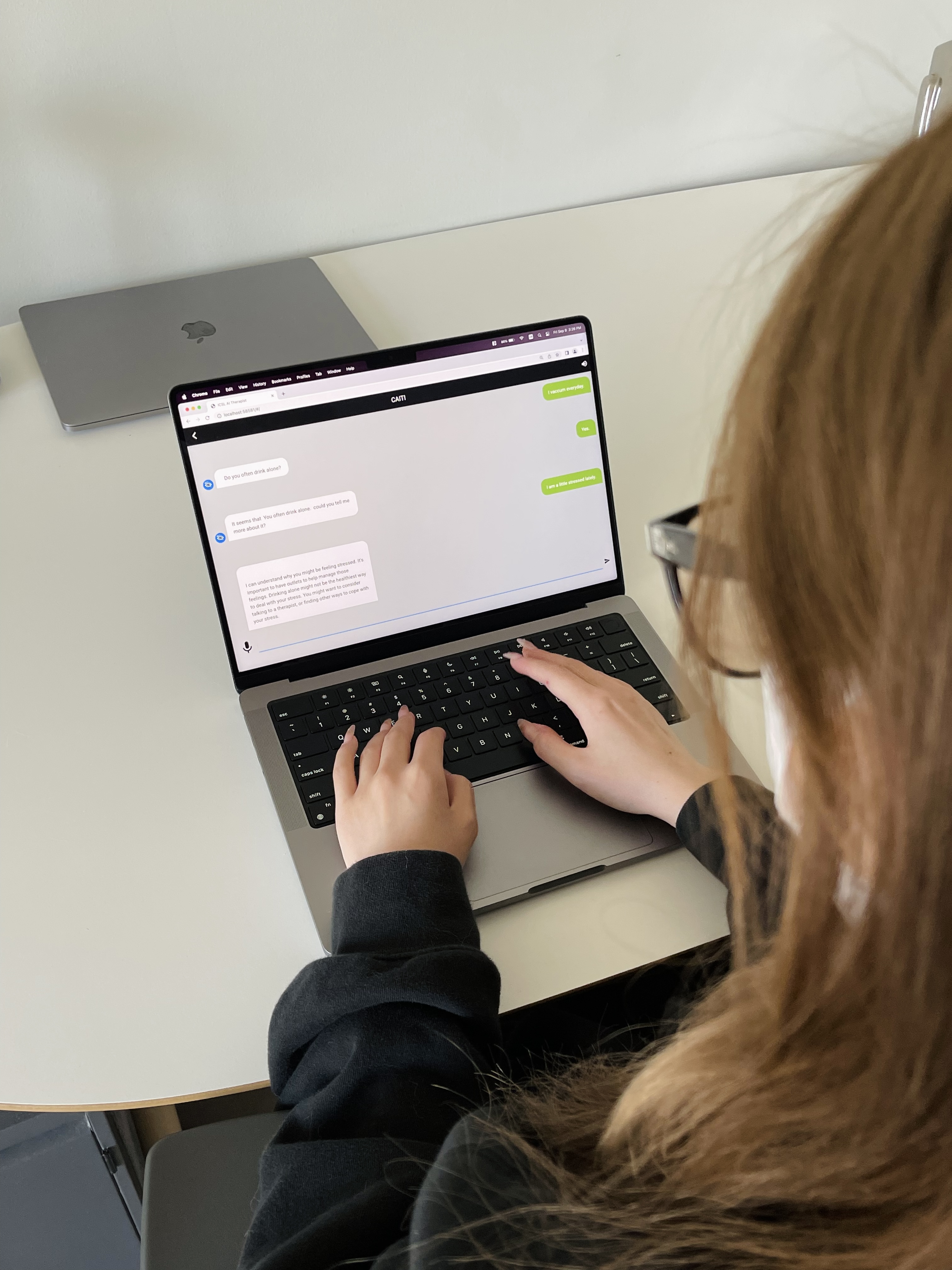}
    \caption{}
    \label{fig:subjectHome}
     \end{subfigure}
     \vspace{-\baselineskip}
        \caption{(a): Home page of \sys{}'s smartphone platform. (b) and (c): Subject chat with \sys{} via the smartphone/Amazon Echo at the end of an in-lab session. (d):A subject chats with \sys{} on her computer at home.}
    \vspace{-\baselineskip}
   \label{fig:implementation}
    
\end{figure*}

Figure~\ref{fig:phoneApp_MI_CBT} shows the smartphone interface for \sys{}'s conversation session, where the various psychotherapeutic interventions applied during different stages of the conversation are annotated. Figure~\ref{fig:implementation} displays the home page of the \sys{} on a smartphone (Figure~\ref{fig:SmartphoneHome}), shows user interactions with \sys{} via voice commands using both the smartphone and an Amazon Echo during in-lab sessions (Figures~\ref{fig:Human-Interaction} and~\ref{fig:Human-Interaction1}), and depicts a user at home interacting with the \sys{} through text input on a computer (Figure~\ref{fig:subjectHome}).

\subsection{Study Design}\label{secsec:study_design}
\begin{wrapfigure}{r}{0.45\linewidth}
    \vspace{-0.2cm}
    \includegraphics[width=0.45\textwidth]{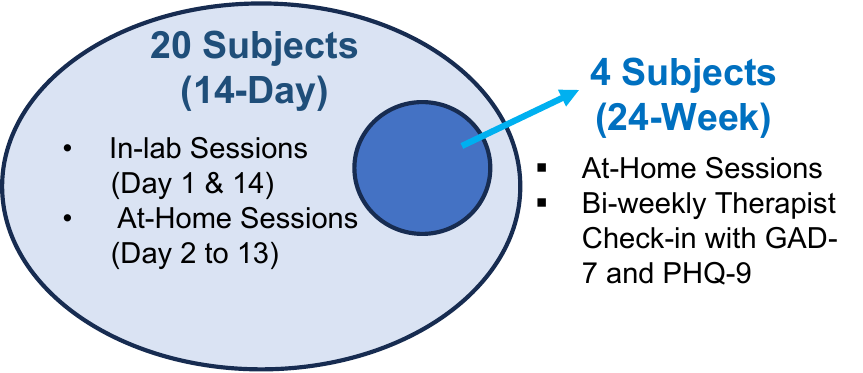}
    \vspace{-0.2cm}
    \caption{The modules and tasks involved in the 14-day study and 24-week longitudinal study.}
    \label{fig:study_design}
\end{wrapfigure}

We first conducted a 14-day study with in-lab and at-home sessions for each subject. The subjects participated in the in-lab session on the first and last days. Afterward, 4 subjects voluntarily participated in a 24-week at-home longitudinal study. A licensed psychotherapist implemented bi-weekly PHQ-9 and GAD-7 assessments with these 4 subjects to measure the effectiveness of \sys{}. Figure~\ref{fig:study_design} shows the modules and tasks implemented in each setup. Subjects were told they were able to unselect some dimensions if they felt uncomfortable through the smartphone interface shown in Figure~\ref{fig:selection}, but they were encouraged to select all. All conversation sessions (user and \sys{} responses, \textbf{\texttt{Response Anayzer}}'s results, \textbf{\texttt{Reasoner}}s' results) are saved for evaluation purposes. 

\subsubsection{In-Lab Session}
The scope of the study and a tutorial about \sys{} system are given at the first in-lab session after informed consent was obtained. During this session, subjects were informed that the dialogue with \sys{} would incorporate elements of psychotherapy. However, to prevent their responses from being influenced and to maintain uniformity in the user experience, the underlying principles and methodologies of the psychotherapies (Motivational Interviewing (MI) and Cognitive Behavioral Therapy (CBT)) were not disclosed. Then, each subject is asked to choose their favorite method (smartphone platform/laptop platform/Amazon Echo) to converse with \sys{}. At the end of every in-lab session, subjects evaluated the system and provided feedback. This evaluative data will be examined in detail in Section~\ref{sec:user_study}. 

\subsubsection{At-Home Session}
Participants were requested to engage in dialogues with the \sys{} at their convenience, with a recommended frequency of once daily or, at a minimum, twice weekly. Among the cohort, four subjects agreed to extend their participation to a 24-week duration. These individuals had regular sessions with a licensed therapist throughout the study period. To evaluate the severity of depression and generalized anxiety disorder, the PHQ-9 and GAD-7 scales were administered bi-weekly, respectively. Psychotherapists closely monitored the participants to ascertain that involvement in the study did not exert any adverse effects on their daily lives.
\section{System Evaluation} ~\label{sec:evaluation_microbenchmarks}
We combine all the logs from conversation sessions during the 14-day and 24-week study and evaluate \sys{}'s system performance in this section. There are 454 conversation sessions with 18,309 and 2,013 segments of subjects' responses to \sys{}'s original and follow-up questions, respectively. Some subjects answered the question with more than one sentence (segment). There are 107 times when the subjects failed to provide valid follow-up responses at the first attempt and triggered \textbf{\texttt{R-V Guide}}. \sys{} provides 2,013 empathic validation and support sessions with \textbf{\texttt{R-V Validator}}. 

As described in Section~\ref{secsec: conversation_method_CBT_reasoner}, \sys{} will conclude the CBT process, if the user does
not provide a valid response after two attempts in each stage of the three-stage CBT process. There are 454 CBT sessions at the end of each conversation session. Among these CBT sessions, as shown in Table~\ref{tab:CBT_evaluation} 3, 6, and 3 of the CBT sessions terminated (fail to provide valid or relevant responses within 3 attempts) in \textbf{\texttt{CBT\_Stage1}}, \textbf{\texttt{CBT\_Stage2}}, and \textbf{\texttt{CBT\_Stage3}}, respectively. The number of user attempts in each CBT stage is shown in this table. As such, there are 33 \textbf{\texttt{CBT\_Stage1 Guide}}s, 44 \textbf{\texttt{CBT\_Stage2 Guide}}s, and 26 \textbf{\texttt{CBT\_Stage3 Guide}}s, respectively. And the \textbf{\texttt{CBT\_Stage1 Reasoner}}, \textbf{\texttt{CBT\_Stage2 Reasoner}}, and \textbf{\texttt{CBT\_Stage3 Reasoner}} are called 487, 495, 474 times, respectively.

\begin{wrapfigure}{r}{0.3\linewidth}
    \vspace{-0.2cm}
    \includegraphics[width=0.3\textwidth]{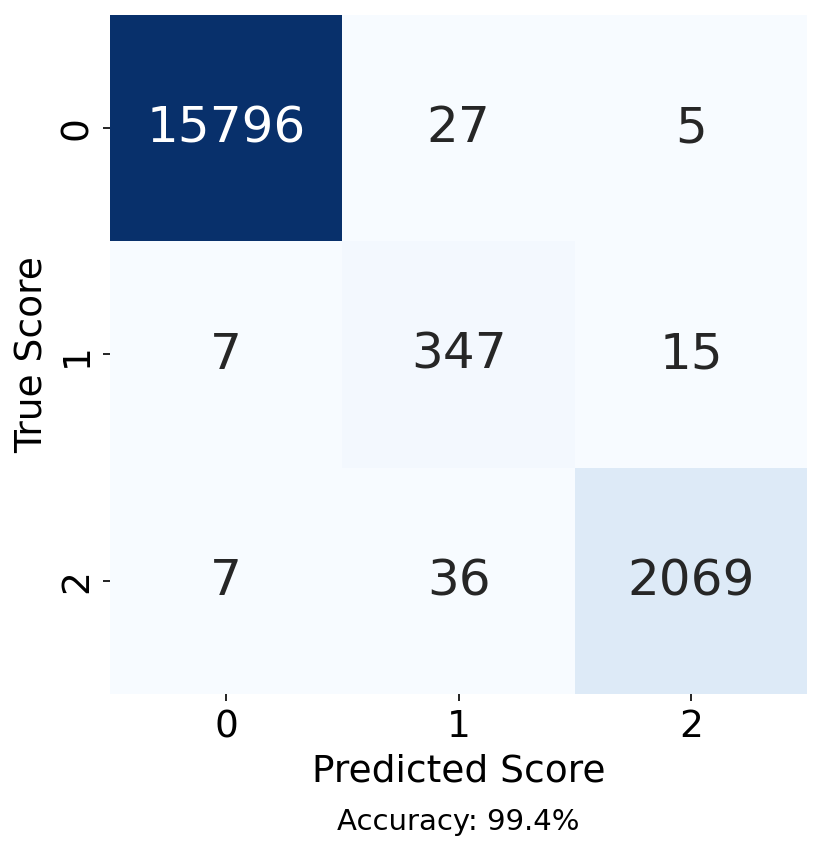}
    \vspace{-0.2cm}
    \caption{\sys{}'s performance to assign \textbf{Score} to the \emph{Segment}s of the user responses.}
    \label{fig:confusion_chat}
    \vspace{-\baselineskip}
\end{wrapfigure}

4 licensed psychotherapists label the ground truth of output generated by these \textbf{\texttt{Reasoner}}s, \textbf{\texttt{Validator}}, and \textbf{\texttt{Guide}}s to evaluate their performance. Specifically, each therapist individually labels the outputs, with the ground truth determined by the majority vote among their evaluations.

In general, during the study with subjects, the performances of various LLM-based functional modules were either better than or comparable to their performances on the datasets provided by psychotherapists during the microbenchmark experiments. This phenomenon is expected, since there is a lower probability for subjects to provide invalid or illogical responses compared to the proportion of ``invalid'' or ``inappropriate'' responses in the datasets provided by the therapists. \emph{Note that \sys{} is designed for precautionary screening, assistance, and conversational psychotherapeutic intervention, and it is not intended to replace the process of diagnosis or clinical treatment.}

\begin{table*}[t!]
\caption{Number of times that \textbf{\texttt{CBT Reasoner}} and \textbf{\texttt{CBT Guides}} triggered in each attempt and the number of CBT sessions terminated in the three CBT stages.}
\vspace{-\baselineskip}
\begin{tabularx}{0.7\textwidth}{l|X|X|X}
\hline
     & \textbf{\texttt{CBT\_Stage1}} & \textbf{\texttt{CBT\_Stage2}} & \textbf{\texttt{CBT\_Stage3}} \\ \hline\hline
\textbf{\texttt{CBT Reasoner} in Attempt 1} & 454 & 451 & 448 \\\hline
\textbf{\texttt{CBT Guide} in Attempt 1} & 23 & 37 & 21 \\\hline
\textbf{\texttt{CBT Reasoner} in Attempt 2} & 23 & 37 & 21 \\\hline
\textbf{\texttt{CBT Guide} in Attempt 2} & 10 & 7 & 5 \\\hline
\textbf{\texttt{CBT Reasoner} in Attempt 3} & 10 & 7 & 5 \\\hline
\textbf{Terminated in Attempt 3} & 3 & 6 & 3 \\\hline

\end{tabularx}
\label{tab:CBT_evaluation}
\end{table*}

\subsection{Response Analyzer}
The user responses to the original questions were divided into segments. Each response segment was classified into \textbf{(Dimension, Score)} pairs, with ground truth labeled by the therapists. Dimension classification accuracy (5 classes of general responses and 37 dimensions) reached 97.6\%. Figure~\ref{fig:confusion_chat} presents the confusion matrix for the \textbf{Score} of the 7,989 segments with a 99.4\% accuracy.

\subsection{\textbf{\texttt{R-V Reasoner}}, \textbf{\texttt{R-V Guide}}, and \textbf{\texttt{R-V Validator}}}

\begin{table*}[t!]
\caption{ The performance for \textbf{\texttt{R-V Reasoner}} (0: Valid, 1: Invalid), \textbf{\texttt{R-V Guide}}, and \textbf{\texttt{R-V Validator}} in handling the responses from subjects during the experiment.}
\vspace{-\baselineskip}
\begin{tabularx}{\textwidth}{XXX|X|X}
\hline
      \multicolumn{3}{c|}{\textbf{\texttt{R-V Reasoner}}} & \textbf{\texttt{R-V Guide}} & \textbf{\texttt{R-V Validator}}\\
      \small{\textbf{Accuracy}} & \small{\textbf{Precision}} & \small{\textbf{Recall}} & \small{\textbf{Accuracy}}& \small{\textbf{Accuracy}}\\ \hline\hline
 97.97\% & 70\% & 98.99\% & 95.32\% & 96.57\%  \\\hline
\end{tabularx}
\label{tab:RV_res_eval}
\vspace{-0.5\baselineskip}
\end{table*}

During the 2,013 reflection-validation (R-V) process, \textbf{\texttt{R-V Reasoner}} and \textbf{\texttt{R-V Guide}} were activated 2,120 and 107 times, respectively. As shown in Table~\ref{tab:RV_res_eval}, \sys{}'s \textbf{\texttt{R-V Reasoner}} almost perfectly identifies all ``invalid'' follow-up responses from the user with only 1 exception. \textbf{\texttt{R-V Reasoner}} misclassified 42 valid follow-up responses as invalid, which is acceptable in the context of precautionary psychotherapeutic intervention, as it would guide the user to provide valid follow-up responses with better quality. In addition, \textbf{\texttt{R-V Guide}} achieved an accuracy of 96.57\%, with only 5 guides being slightly not perfect. The causes for these 5 imperfect guides are overinterpreting the relationship between the follow-up response and the original response and missing some information from the user responses.

The therapists also checked all the 2,013 empathic validations and supports provided by the \textbf{\texttt{R-V Validator}} to the subjects through a majority vote, with 69 being slightly inappropriate. Specifically, when follow-up responses were too brief (under 3 words), \sys{} struggled to comprehend, leading to 24 improper empathic supports. 15 inappropriate instances resulted from inconsistencies between the subjects' original and follow-up responses. Another 30 inappropriate validations are due to the GPT-based \textbf{\texttt{R-V Validator}} adding its own interpretation of user responses into the empathic validation. Overall, \sys{} effectively delivered empathic validation and support in over 96.5\% of instances. Although there are concerns about bias in large language models, the fine-tuned models in this work perform well in user studies and exhibit minimal bias. 

\subsection{\textbf{\texttt{CBT Reasoner}} and \textbf{\texttt{CBT Guide}}}

Table~\ref{tab:CBT_reasoner_guide_eval} illustrates the performance of \textbf{\texttt{CBT Reasoner}}s and \textbf{\texttt{CBT Guide}}s when handling the responses from subjects from all attempts during the three-stage CBT process. It is shown that \textbf{\texttt{CBT Reasoner}}s in each stage have high accuracies in identifying the validity and utility of the user responses to meet the psychotherapeutic goal of the CBT process. The \textbf{\texttt{CBT Reasoner}}s also achieve high recall, which demonstrates that it is extremely rare for \textbf{\texttt{CBT Reasoner}}s to miss any user response that's not valid or not related.

\textbf{\texttt{CBT\_Stage1 Guide}}, \textbf{\texttt{CBT\_Stage2 Guide}}, and \textbf{\texttt{CBT\_Stage3 Guide}} generated only 3, 3, and 2 less ideal context to guide the subjects. The most common issues for these 8 suboptimal \textbf{\texttt{CBT Guide}}s is that they tried to read minds and made excessive assumptions about the relationships of the user responses earlier in the CBT process. None of these suboptimal \textbf{\texttt{CBT Guide}}s completely misguide the user or lead them in the wrong direction.

\begin{table*}[t!]
\caption{\textbf{\texttt{CBT Reasoner}} and \textbf{\texttt{CBT Guide}} performance in handling the responses from subjects during the three-stage CBT process (0: Valid, 1: Invalid).}
\vspace{-\baselineskip}
\begin{tabularx}{\textwidth}{XXX|XXX|XXX}
\hline
     \multicolumn{3}{c|}{\textbf{\texttt{CBT\_Stage1 Reasoner}}} & \multicolumn{3}{c|}{\textbf{\texttt{CBT\_Stage2 Reasoner}}} & \multicolumn{3}{c}{\textbf{\texttt{CBT\_Stage3 Reasoner}}} \\
     \small{\textbf{Accuracy}} & \small{\textbf{Precision}} & \small{\textbf{Recall}} & \small{\textbf{Accuracy}} & \small{\textbf{Precision}} & \small{\textbf{Recall}} & \small{\textbf{Accuracy}} & \small{\textbf{Precision}} & \small{\textbf{Recall}} \\ \hline\hline
     99.38\% & 93.39\% & 96.88\% & 98.78\% & 88.64\% & 97.5\% & 99.57\% & 100\% & 92.85\% \\\hline\hline
     \multicolumn{3}{c|}{\textbf{\texttt{CBT\_Stage1 Guide}}} & \multicolumn{3}{c|}{\textbf{\texttt{CBT\_Stage2 Guide}}} & \multicolumn{3}{c}{\textbf{\texttt{CBT\_Stage3 Guide}}} \\
     \multicolumn{3}{c|}{\small{\textbf{Accuracy}}} & \multicolumn{3}{c|}{\small{\textbf{Accuracy}}} & \multicolumn{3}{c}{\small{\textbf{Accuracy}}} \\\hline\hline
     \multicolumn{3}{c|}{90.9\%} & \multicolumn{3}{c|}{95.45\%} & \multicolumn{3}{c}{92.23\%} \\\hline
     
\end{tabularx}
\label{tab:CBT_reasoner_guide_eval}
\end{table*}
\section{User Study} \label{sec:user_study}
In this section, we present the quantitative and qualitative feedback of the 20 subjects who participated in our study (20 subjects participated in the 14-day study and 4 subjects extended to 24 weeks). We also organize and discuss the qualitative feedback and evaluations from the 4 psychotherapists working with us on this project.

\subsection{User Adaptation Indicated by Device Usage}
\begin{figure*}[t!]
         \centering
         \includegraphics[width=0.7\textwidth]{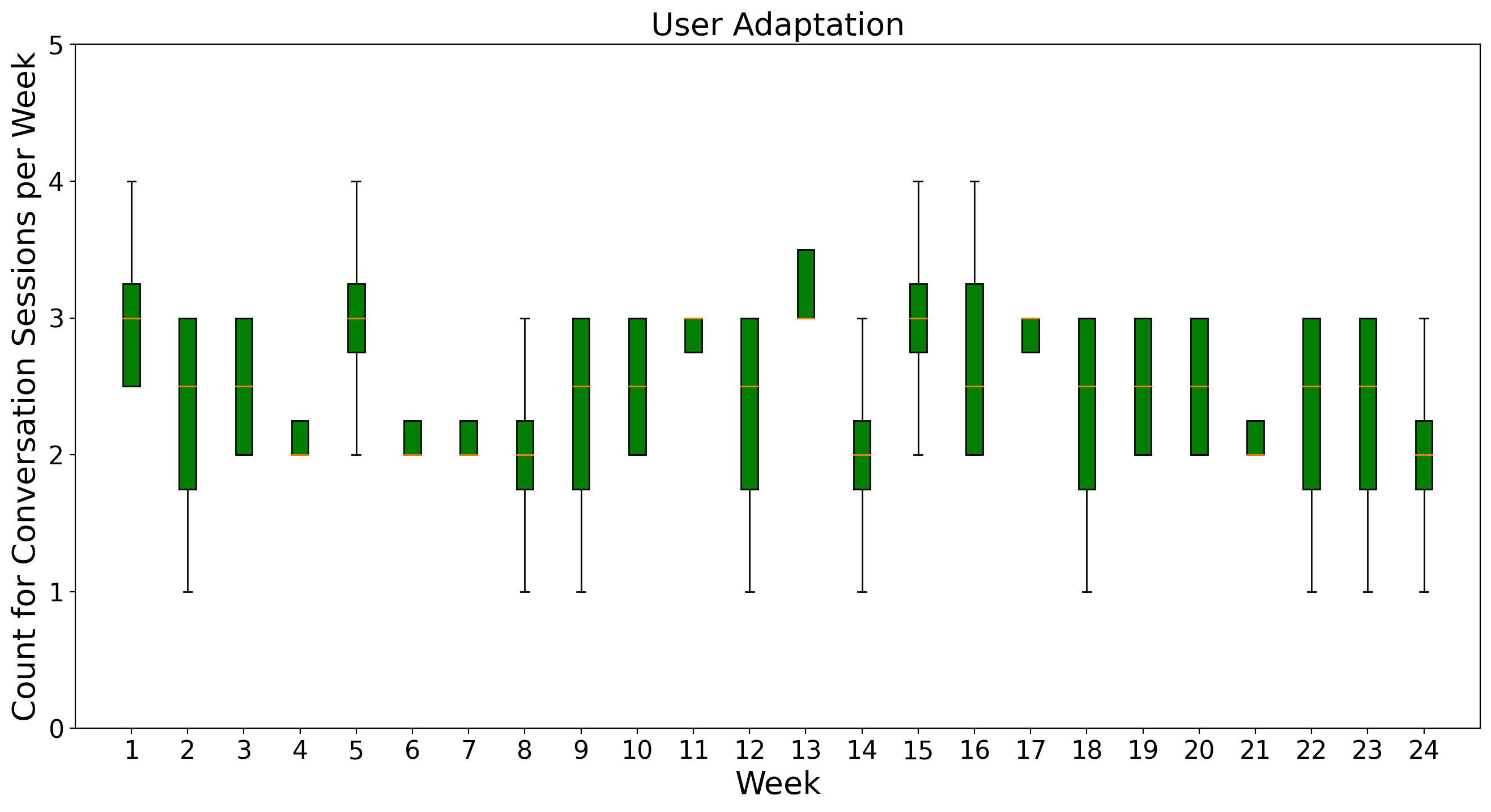}
         \vspace{-\baselineskip}
         \caption{The user adaptation \sys{} during the 24-week study.}
         \label{fig:user_adaptation}
         \vspace{-2\baselineskip}
\end{figure*}

During the 24-week study, we meticulously tracked the number of conversation sessions each subject had per week with \sys{}. As illustrated in Figure~\ref{fig:user_adaptation}, the data indicates that all four subjects engaged in dialogue with \sys{} with a frequency of 2 to 3 times daily. Notably, this consistent engagement over the course of the study suggests a sustained use of the system by the participants. The frequency of their interactions with \sys{} did not exhibit a significant decline over time, implying a stable user adaptation and a persistent incorporation of the system into their daily routines. This enduring engagement underlines the utility and user-friendliness of \sys{}, as well as its potential to maintain user interest and interaction over extended periods.

\subsection{Quantitative Analysis}

ChatBot Usability Scale (BUS-15) is a recently developed tool to assess end-users' satisfaction with chatbots~\cite{borsci2022chatbot}. Taking BUS-15 into account, we devised the 10 aspects listed in Table ~\ref{tab:user_feedback}. Subjects rated the system from 1 (poor) to 5 (excellent) for these 10 aspects on the first and last day of the experiment. In general, subjects gave high ratings to \sys{}. Most subjects thought positively of the conversations with \sys{} and were willing to recommend and continue using \sys{} in the future. Only one subject showed a slightly negative attitude towards \sys{} and gave low scores. This subjective thought that \emph{``the technology is too intelligent and makes me worried that AI might be playing a too important role in my daily life''} (S19).

A Wilcoxon signed-rank test, a nonparametric statistical within-subject test commonly used in behavioral science~\cite{shin2022chatbots}, was performed to compare the subjects' day-to-day functioning on the first and last days of the 14-day study based on their responses \textbf{Score} on all 37 dimensions (\textbf{Score} $\in \{0,1,2\}$, described in Section~\ref{sec:System_Architecture}). The test results show a statistically significant decrease in the dimensions with \textbf{Score} equal to 1 and 2 on the last day, as compared to the first day of the experiment (z = -2.68, p < .01), showing a decrease in dimensions subjects reported with concerns functioning. This suggests an overall reduction in the dimensions that subjects reported as having functional concerns. The results indicate that the subjects' day-to-day functioning may have improved during the experiment using \sys.

In a 24-week longitudinal study, 4 subjects completed additional GAD-7 and PHQ-9 assessments every two weeks. These tools evaluate the severity of anxiety and depression symptoms. With the small sample size (n=4), therapists reviewed the questionnaire results individually. During the study, two subjects improved from moderate-to-severe anxiety and depression to mild levels, initially facing difficulties in functioning but eventually reporting none. One subject progressed from mild to minimal anxiety and depression and reported no functional difficulties. The remaining participant exhibited fluctuating symptoms, ranging from mild to minimal, initially facing some difficulties in functioning, but reporting none by the study's end. The findings of this study indicate that \sys{} was effective in reducing the severity of anxiety and depression, as well as enhancing daily functioning.
            
\begin{table*}[t!]
\caption{The average ratings on the first day (Day 1) and last day (Day 14) of the study.}
\vspace{-\baselineskip}
\begin{tabular}{||l|c|c||}
\hline
      \textbf{Question Asked}      & \multicolumn{2}{c||}{\textbf{Average Rating}} \\ \cline{2-3}
                                    & \textbf{Day 1} & \textbf{Day 14} \\ \hline\hline
     How do you like \sys{} in general? & 4.3 & 4.45 \\\hline
     Are you willing to continue using \sys{} in the future? & 4.3 & 4.35 \\\hline
     Will you recommend \sys{} to others? & 4.2 & 4.55 \\\hline
     Do you think \sys's functions are easily detectable? & 4.4 & 4.5 \\\hline
     Do you think \sys{} can understand you and chat with you naturally? & 4.35 & 4.5 \\\hline
     Do you like the interface design of \sys? & 4.6 & 4.2 \\ \hline
     Do you think the wording used in \sys{} is appropriate? & 4.6 & 4.75 \\\hline
     Do you think the psychotherapies are helpful? & 4.65 & 4.6 \\\hline
     How do you feel about the response time from \sys{}? & 4.8 & 4.85\\\hline
     Do you feel that the conversation flow of \sys{} is gradually tailoring to you? & N.A. & 4.75 \\\hline
     
\end{tabular}
\label{tab:user_feedback}
\vspace{-1.5\baselineskip}
\end{table*}

\subsection{Qualitative Evaluation from the Subjects and Therapists}

\subsubsection{Experiences and Feedback from Subjects}

All 20 subjects who participated in the study found \sys{} to be \textbf{valid and effective}. Subjects see using \sys{} brings awareness to their mental health daily, which makes them feel they are \emph{``relying on the system to do the self-reflection work''} (S10) every day. One of the subjects said: \emph{``I feel in good shape doing these check-ins every day, in between my weekly sessions with individual therapist''} (S11). Subjects were also surprised about how well \emph{``\sys{} can understand me and makes me feel validated''}(S17). One subject stated that:

\begin{center}
\begin{minipage}[!h]{0.8\linewidth}
\emph{``I think \sys{} does a pretty good job validating my feelings and encouraging me to be more active. I had a very positive experience with the comforting part. I also like the sensing part, as \sys{} knows what I am doing at home and directly provides help if needed. I don't need to describe my daily routine and recall when I am with \sys{}. I need to do these annoying things during my therapist visits. I think \sys{} is a good add-on in mental healthcare''} (S15).
\end{minipage}
\end{center}

Several subjects complimented on \sys's \textbf{interactiveness} and stated \emph{``\sys{} is really plug-and-play and easy to use''} (S01). Additionally, a large portion of subjects stated that they liked the way that \sys{} ``talked'' to them as \emph{``the conversation is very genuine''} (S12). For example, one subject said:

\begin{center}
\begin{minipage}[!h]{0.8\linewidth}
\emph{``I feel like that I have a companion. When \sys{} talks to me through my Alexa, it listens and converses naturally and reminds me of parts of my therapy sessions. The system reformulates the questions it asks me every time. Also, I realize that if I use \sys{} more frequently, \sys{} is more attuned to me because it changes the way it asks the question according to my answers''} (S02).
\end{minipage}
\end{center}

Subjects found the \textbf{psychotherapies helpful}, encouraging, and valuable. A subject said \emph{``I like the counseling part of the system. The tone is supportive and encouraging. The system really understands what I said to it and provides reasonable and applicable guides''} (S08). In addition, a few subjects find the guidance (\textbf{\texttt{Guide}} feature) provided by \sys{} during the psychotherapies helpful. 

\begin{center}
\begin{minipage}[!h]{0.8\linewidth}
\emph{``To be honest, I was not that familiar with what they called the CBT procedures at the end of each conversation. Initially, I did not know what are the ``unhelpful thoughts" in my situations. But, you know, after hanging using {\sys} a few times, I started to get the hang of these helpful thinking strategies. I am getting more optimistic and better at boosting my own confidence, when I face challenges.''} (S01).
\end{minipage}
\end{center}

A few subjects feel that \emph{``the consolation is not pointed enough''} (S02), as it provides very general comments that are suitable for \emph{``everyone who has the same problem''} (S02). They would like to have more personalized experiences with more targeted suggestions.

\subsubsection{Comments from Therapists}

4 psychotherapists approved of \sys{}, recognizing its potential for \emph{``combining physical and mental wellness screening and providing psychotherapeutic precautionary interventions in daily life for everyone''} (T02).  They provided positive feedback on the conversational daily functioning screening, the integration of psychotherapeutic techniques within the conversation flow, and the overall style of the dialogue. They \textbf{approved the content and style of the system's language}: \emph{``\sys{} engaged me in conversations with effective content control. However, ChatGPT sometimes produced responses that are not up to clinical standards''} (T01).

The therapists also saw the interventions being \textbf{potentially used in addition to clinical treatment}. They spoke posivitely of the empathic validation and support in the MI process and valued the significant influence of \sys{}'s \textbf{\texttt{Reasoner}} and \textbf{\texttt{Guide}} features in guiding and steering the user towards more adaptive thinking in the CBT process. They felt that the combination of reasoning and guidance not only prevents users from getting stuck in incorrect ways of thinking but also positively directs and facilitates gradual learning. This, in turn, can support overall mental health well-being, especially in the long term. They mentioned these designs could \emph{``enforce the effectiveness of \sys's psychotherapy''} (T03) and \emph{``provide just the right amount of guidance to support users in situations that require additional attention''} (T01). One therapist stated: 

\begin{center}
\begin{minipage}[!h]{0.8\linewidth}
\emph{``From my observations on the interactions between \sys{} and our subjects, I think the validation implementations of CBT and MI are appropriate. Although very few exchanges are not perfect if the responses from the subjects are ambiguous, \sys{} impresses me with emotionally supportive text formulation. I also answered questions in a way a few times to indicate daily functioning concerns to \sys{} during different sessions, and \sys{} came up with different ways to guide and help me to improve the situation.''} (T02).
\end{minipage}
\end{center}

Most therapists found \sys{} to \textbf{aid traditional psychotherapeutic processes}, stating that \textbf{more frequent data collection offers more insights} between therapy sessions. \emph{``I would love to see some of my clients use this''} (T04), one therapist commented. Another therapist mentioned: 

\begin{center}
\begin{minipage}[!h]{0.8\textwidth}
\emph{``Doing everyday check-ins helps people in general to bring awareness to mental health in their everyday activities. Sometimes, I assign my clients homework to log their everyday activities, just for them to keep doing the work outside of therapy sessions. It would be great if more people can use this system daily and become more intentional in their daily routine``} (T02).
\end{minipage}
\end{center}

They also validated \sys{}'s use of reinforcement learning for conversation generation and \emph{``are surprised by how attentive CaiTI is and how good the flow of the conversation we have''} (T04). Additionally, they offered suggestions for future enhancements, including adjustments to \sys{}'s audio tone.

\section{Discussion and Future Work}
\label{sec:future}

As shown in Sections~\ref{sec:evaluation_microbenchmarks} and~\ref{sec:user_study}, \sys{} can effectively perform \emph{Converse with the User in a Natural Way}, and provide \emph{Psychotherapeutic Conversational Intervention}. The qualitative evaluations further attest to the system's usability, the relevance and effectiveness of its psychotherapeutic content, the helpfulness of the guides, and the efficacy of the empathic validation it offers, making it a promising tool for personalized mental health care. Throughout the rest of this section, we summarize some of the current limitations and propose plans for future improvements and visions.
%We also envision \sys{} to be one of the many applications commonly integrated into smart homes and psychotherapy treatment in the future. 

First of all, we will continue our collaboration with the psychotherapists, we plan to include real patients with different kinds and severity of mental disorders in longitudinal studies. We will evaluate how well \sys{} can assist the treatment provided by the therapists and improve the mental well-being of the patients.

Moreover, during the psychotherapy process, although \sys{} breaks the tasks down and leverages LLM-based \textbf{\texttt{Reasoner}}s, \textbf{\texttt{Guide}}s, and \textbf{\texttt{Validator}} to specifically handle each subtask, there is room to improve the accuracies of \textbf{\texttt{Reasoner}}s and quality of \textbf{\texttt{Guide}}s, and \textbf{\texttt{Validator}}. In particular, we plan to add step-by-step ``system reasoners'' (Chain-of-Thought) to evaluate if the \textbf{\texttt{Guide}}s and \textbf{\texttt{Validator}}s are suboptimal because of ``reading user's minds'' and making excessive assumptions~\cite{radhakrishnan2023question}. Additionally, we plan to investigate if further breaking down the tasks would improve the system's performance. For example, we will investigate whether using two LLM modules for classifying \textbf{Dimension} and \textbf{Score} in \textbf{\texttt{Response Analyzer}} instead of one will improve the accuracy, and how this change affects the performance of different LLM models. We are aware that these modifications will increase the computational overhead for the system, yielding longer system response time and affecting the user experience. As such, we also plan to investigate the trade-off between the system complexity and user experiences in multiple aspects.

We see the potential of incorporating common wearable devices, such as smartwatches and smartphones, as potential platforms into \sys{}. In fact, around half of the subjects in this study actively use smart devices for health and fitness monitoring. We plan to leverage the health and fitness data analyzed by smartphones or smartwatches (e.g., Apple HealthKit data) as a source to perform activity detection with existing commercial smart home devices.

In addition, the smart home market is rapidly expanding with new sensors and devices. \sys{} has the potential to be one of the many applications commonly integrated into smart home ecosystems. As such, we plan to investigate how to take advantage of smart home sensors, devices, and robots to provide a more comprehensive and invasive screening of users' daily functioning through smart sensors and various kinds of interactions and interventions through home robots or other devices. With omnipresent modules being included in \sys{}, to make users feel less ``invasive'', we plan to make \sys{} plug-and-play. Users can turn on devices that make them feel comfortable at that moment, \sys{} can automatically discover available resources and generate execution pipelines to screen the wellness of the user and provide interventions if necessary.

Furthermore, equipping with the RL recommender as well as fine-tuned and few-shot prompted GPT-based models for conversation, conversational psychotherapeutic intervention generations, \sys{} can generate speech for conversation and lead the conversation flow in a more human-like way compared to other platforms. However, since \sys{} uses the text-to-speech API, the tone and inflection of the voice generated to ``talk'' to the user are still not entirely identical to what a real person would perform. A therapist commented: \emph{``It's not exactly how a real person would speak in terms of the tone and inflection but I think that definitely will improve in time as you know text-to-speech and other natural language things become better and better. However, the content and style of the sentences and words are very much in line with what a typical person would say''} (T04). We plan to investigate and incorporate deep learning methods using to add ``emotion'' to the audio output accordingly~\cite{adigwe2018emotional}.

We have designed \sys{} to minimize bias by implementing several modules of LLMs tailored to specific tasks instead of relying on a single model. We acknowledge that despite our best efforts, all AI applications, including this one, are subject to potential bias and the ethical concerns of AI for psychotherapies still remain. Therefore, the intended application of \sys{} is primarily for precautionary day-to-day functioning screenings and psychotherapeutic interventions, aiming for better self-care and assisting the professional psychotherapy process.

\section{Conclusion}
\label{sec:conclusion}

In this work, collaborating with licensed psychotherapists, we propose \sys{}, a conversational ``AI therapist'' that leverages the LLMs and RL to screen and analyze the day-to-day functioning of the user across 37 dimensions and provides appropriate and effective psychotherapies, including MI and CBT, depending on the physical and mental status of the user through \emph{natural} and \emph{personalized} conversations. Accessible on common smart devices like smartphones, computers, and smart speakers, \sys{} provides a versatile solution for users, both indoors and outdoors.

To enhance care quality and psychotherapy effectiveness while minimizing AI biases, \sys{} employs specialized LLM-based \textbf{\texttt{Reasonser}}s, \textbf{\texttt{Guide}}s, and \textbf{\texttt{Validator}} to provide psychotherapy during the conversation. By using conversation datasets annotated by licensed psychotherapists, we assess various GPT and Llama-2 LLMs' performances through few-shot prompts or fine-tuning across different LLM-based tasks in {\sys}. Based on the assessment results and observations, we collaborate with the therapists to develop a proof-of-concept prototype of \sys{} and conduct 14-day and 24-week studies in real-world trials. The evidence gathered through our studies showcases \sys{}'s effectiveness in screening daily functioning and providing psychotherapies. Additionally, the empirical evidence underscores \sys{}'s capability in enhancing daily functioning and mental well-being. We see the potential for LLM-based conversational ``AI therapists'' to aid the traditional psychotherapy process and provide precautionary mental health and psychotherapeutic self-care to a larger population.

\bibliographystyle{ACM-Reference-Format}
\bibliography{reference.bib}


\begin{thebibliography}{102}

%%% the Web address.

\ifx \showCODEN    \undefined \def \showCODEN     #1{\unskip}     \fi
\ifx \showDOI      \undefined \def \showDOI       #1{#1}\fi
\ifx \showISBNx    \undefined \def \showISBNx     #1{\unskip}     \fi
\ifx \showISBNxiii \undefined \def \showISBNxiii  #1{\unskip}     \fi
\ifx \showISSN     \undefined \def \showISSN      #1{\unskip}     \fi
\ifx \showLCCN     \undefined \def \showLCCN      #1{\unskip}     \fi
\ifx \shownote     \undefined \def \shownote      #1{#1}          \fi
\ifx \showarticletitle \undefined \def \showarticletitle #1{#1}   \fi
\ifx \showURL      \undefined \def \showURL       {\relax}        \fi
% invisible to TeX
\providecommand\bibfield[2]{#2}
\providecommand\bibinfo[2]{#2}
\providecommand\natexlab[1]{#1}
\providecommand\showeprint[2][]{arXiv:#2}

\bibitem[Onl(2022)]%
        {OnlineTest2}
 \bibinfo{year}{2022}\natexlab{}.
\newblock \bibinfo{title}{{Mental Health Tests, Quizzes, Self-Assessments, \& Screening Tools}}.
\newblock \bibinfo{howpublished}{\url{https://www.psycom.net/quizzes}}.
\newblock
\newblock
\shownote{[Online; accessed 16-August-2022]}.


\bibitem[Spe(2022)]%
        {Speech-to-text}
 \bibinfo{year}{2022}\natexlab{}.
\newblock \bibinfo{title}{{Speech-to-Text}}.
\newblock \bibinfo{howpublished}{\url{https://cloud.google.com/speech-to-text}}.
\newblock
\newblock
\shownote{[Online; accessed 24-April-2022]}.


\bibitem[cha(2023)]%
        {chatgpt_news1}
 \bibinfo{year}{2023}\natexlab{}.
\newblock \bibinfo{title}{3 things to know before talking to ChatGPT about your mental health}.
\newblock \bibinfo{howpublished}{\url{https://mashable.com/article/how-to-chat-with-chatgpt-mental-health-therapy}}.
\newblock


\bibitem[red(2023)]%
        {reddit2023using}
 \bibinfo{year}{2023}\natexlab{}.
\newblock \bibinfo{title}{Using {ChatGPT} as a therapist?}
\newblock
\newblock
\urldef\tempurl%
\url{https://www.reddit.com/r/ChatGPTPro/comments/126rtvb/using_chatgpt_as_a_therapist/}
\showURL{%
\tempurl}


\bibitem[Ale(2023)]%
        {AlexaSkills}
 \bibinfo{year}{2023}\natexlab{}.
\newblock \bibinfo{title}{What is the Alexa Skills Kit?}
\newblock
\newblock
\urldef\tempurl%
\url{https://developer.amazon.com/en-US/docs/alexa/ask-overviews/what-is-the-alexa-skills-kit.html}
\showURL{%
\tempurl}


\bibitem[Aarons et~al\mbox{.}(2017)]%
        {aarons2017testing}
\bibfield{author}{\bibinfo{person}{Gregory~A Aarons}, \bibinfo{person}{Mark~G Ehrhart}, \bibinfo{person}{Joanna~C Moullin}, \bibinfo{person}{Elisa~M Torres}, {and} \bibinfo{person}{Amy~E Green}.} \bibinfo{year}{2017}\natexlab{}.
\newblock \showarticletitle{Testing the leadership and organizational change for implementation (LOCI) intervention in substance abuse treatment: a cluster randomized trial study protocol}.
\newblock \bibinfo{journal}{\emph{Implementation Science}} \bibinfo{volume}{12}, \bibinfo{number}{1} (\bibinfo{year}{2017}), \bibinfo{pages}{1--11}.
\newblock


\bibitem[Adams et~al\mbox{.}(2018)]%
        {adams2018keppi}
\bibfield{author}{\bibinfo{person}{Alexander~T Adams}, \bibinfo{person}{Elizabeth~L Murnane}, \bibinfo{person}{Phil Adams}, \bibinfo{person}{Michael Elfenbein}, \bibinfo{person}{Pamara~F Chang}, \bibinfo{person}{Shruti Sannon}, \bibinfo{person}{Geri Gay}, {and} \bibinfo{person}{Tanzeem Choudhury}.} \bibinfo{year}{2018}\natexlab{}.
\newblock \showarticletitle{Keppi: A tangible user interface for self-reporting pain}. In \bibinfo{booktitle}{\emph{Proceedings of the 2018 CHI Conference on Human Factors in Computing Systems}}. \bibinfo{pages}{1--13}.
\newblock


\bibitem[Adigwe et~al\mbox{.}(2018)]%
        {adigwe2018emotional}
\bibfield{author}{\bibinfo{person}{Adaeze Adigwe}, \bibinfo{person}{No{\'e} Tits}, \bibinfo{person}{Kevin~El Haddad}, \bibinfo{person}{Sarah Ostadabbas}, {and} \bibinfo{person}{Thierry Dutoit}.} \bibinfo{year}{2018}\natexlab{}.
\newblock \showarticletitle{The emotional voices database: Towards controlling the emotion dimension in voice generation systems}.
\newblock \bibinfo{journal}{\emph{arXiv preprint arXiv:1806.09514}} (\bibinfo{year}{2018}).
\newblock


\bibitem[Alford et~al\mbox{.}(1997)]%
        {alford1997integrative}
\bibfield{author}{\bibinfo{person}{Brad~A Alford}, \bibinfo{person}{Aaron~T Beck}, {and} \bibinfo{person}{John~V Jones~Jr}.} \bibinfo{year}{1997}\natexlab{}.
\newblock \bibinfo{title}{The integrative power of cognitive therapy}.
\newblock
\newblock


\bibitem[Anil et~al\mbox{.}(2023)]%
        {anil2023palm}
\bibfield{author}{\bibinfo{person}{Rohan Anil}, \bibinfo{person}{Andrew~M Dai}, \bibinfo{person}{Orhan Firat}, \bibinfo{person}{Melvin Johnson}, \bibinfo{person}{Dmitry Lepikhin}, \bibinfo{person}{Alexandre Passos}, \bibinfo{person}{Siamak Shakeri}, \bibinfo{person}{Emanuel Taropa}, \bibinfo{person}{Paige Bailey}, \bibinfo{person}{Zhifeng Chen}, {et~al\mbox{.}}} \bibinfo{year}{2023}\natexlab{}.
\newblock \showarticletitle{Palm 2 technical report}.
\newblock \bibinfo{journal}{\emph{arXiv preprint arXiv:2305.10403}} (\bibinfo{year}{2023}).
\newblock


\bibitem[Anthropic(2024)]%
        {anthropic2024claude}
\bibfield{author}{\bibinfo{person}{Anthropic}.} \bibinfo{year}{2024}\natexlab{}.
\newblock \bibinfo{title}{Introducing the next generation of Claude}.
\newblock \bibinfo{howpublished}{\url{https://www.anthropic.com/news/claude-3-family}}.
\newblock
\newblock
\shownote{Accessed: 2024-03-06}.


\bibitem[Anton et~al\mbox{.}(2006)]%
        {anton2006combined}
\bibfield{author}{\bibinfo{person}{Raymond~F Anton}, \bibinfo{person}{Stephanie~S O’Malley}, \bibinfo{person}{Domenic~A Ciraulo}, \bibinfo{person}{Ron~A Cisler}, \bibinfo{person}{David Couper}, \bibinfo{person}{Dennis~M Donovan}, \bibinfo{person}{David~R Gastfriend}, \bibinfo{person}{James~D Hosking}, \bibinfo{person}{Bankole~A Johnson}, \bibinfo{person}{Joseph~S LoCastro}, {et~al\mbox{.}}} \bibinfo{year}{2006}\natexlab{}.
\newblock \showarticletitle{Combined pharmacotherapies and behavioral interventions for alcohol dependence: the COMBINE study: a randomized controlled trial}.
\newblock \bibinfo{journal}{\emph{Jama}} \bibinfo{volume}{295}, \bibinfo{number}{17} (\bibinfo{year}{2006}), \bibinfo{pages}{2003--2017}.
\newblock


\bibitem[Antony et~al\mbox{.}(2005)]%
        {antony2005improving}
\bibfield{author}{\bibinfo{person}{Martin~M Antony}, \bibinfo{person}{Deborah~Roth Ledley}, {and} \bibinfo{person}{Richard~G Heimberg}.} \bibinfo{year}{2005}\natexlab{}.
\newblock \bibinfo{booktitle}{\emph{Improving outcomes and preventing relapse in cognitive-behavioral therapy}}.
\newblock \bibinfo{publisher}{Guilford Press}.
\newblock


\bibitem[Arkowitz and Westra(2004)]%
        {arkowitz2004integrating}
\bibfield{author}{\bibinfo{person}{Hal Arkowitz} {and} \bibinfo{person}{Henny~A Westra}.} \bibinfo{year}{2004}\natexlab{}.
\newblock \showarticletitle{Integrating motivational interviewing and cognitive behavioral therapy in the treatment of depression and anxiety}.
\newblock \bibinfo{journal}{\emph{Journal of Cognitive Psychotherapy}} \bibinfo{volume}{18}, \bibinfo{number}{4} (\bibinfo{year}{2004}), \bibinfo{pages}{337--350}.
\newblock


\bibitem[Beck and Beck(2011)]%
        {beck2011cognitive}
\bibfield{author}{\bibinfo{person}{Judith~S Beck} {and} \bibinfo{person}{AT Beck}.} \bibinfo{year}{2011}\natexlab{}.
\newblock \showarticletitle{Cognitive behavior therapy: Basic and beyond}.
\newblock \bibinfo{journal}{\emph{New York: Guilford}} (\bibinfo{year}{2011}).
\newblock


\bibitem[Bible et~al\mbox{.}(2017)]%
        {bible2017assessment}
\bibfield{author}{\bibinfo{person}{Lisa~J Bible}, \bibinfo{person}{Kristin~A Casper}, \bibinfo{person}{Jennifer~L Seifert}, {and} \bibinfo{person}{Kyle~A Porter}.} \bibinfo{year}{2017}\natexlab{}.
\newblock \showarticletitle{Assessment of self-care and medication adherence in individuals with mental health conditions}.
\newblock \bibinfo{journal}{\emph{Journal of the American Pharmacists Association}} \bibinfo{volume}{57}, \bibinfo{number}{3} (\bibinfo{year}{2017}), \bibinfo{pages}{S203--S210}.
\newblock


\bibitem[Borsci et~al\mbox{.}(2022)]%
        {borsci2022chatbot}
\bibfield{author}{\bibinfo{person}{Simone Borsci}, \bibinfo{person}{Alessio Malizia}, \bibinfo{person}{Martin Schmettow}, \bibinfo{person}{Frank Van Der~Velde}, \bibinfo{person}{Gunay Tariverdiyeva}, \bibinfo{person}{Divyaa Balaji}, {and} \bibinfo{person}{Alan Chamberlain}.} \bibinfo{year}{2022}\natexlab{}.
\newblock \showarticletitle{The Chatbot usability scale: The design and pilot of a usability scale for interaction with AI-based conversational agents}.
\newblock \bibinfo{journal}{\emph{Personal and ubiquitous computing}} \bibinfo{volume}{26}, \bibinfo{number}{1} (\bibinfo{year}{2022}), \bibinfo{pages}{95--119}.
\newblock


\bibitem[Brown et~al\mbox{.}(2016)]%
        {brown2016gamification}
\bibfield{author}{\bibinfo{person}{Menna Brown}, \bibinfo{person}{Noelle O'Neill}, \bibinfo{person}{Hugo van Woerden}, \bibinfo{person}{Parisa Eslambolchilar}, \bibinfo{person}{Matt Jones}, \bibinfo{person}{Ann John}, {et~al\mbox{.}}} \bibinfo{year}{2016}\natexlab{}.
\newblock \showarticletitle{Gamification and adherence to web-based mental health interventions: a systematic review}.
\newblock \bibinfo{journal}{\emph{JMIR mental health}} \bibinfo{volume}{3}, \bibinfo{number}{3} (\bibinfo{year}{2016}), \bibinfo{pages}{e5710}.
\newblock


\bibitem[Bubeck et~al\mbox{.}(2023)]%
        {bubeck2023sparks}
\bibfield{author}{\bibinfo{person}{S{\'e}bastien Bubeck}, \bibinfo{person}{Varun Chandrasekaran}, \bibinfo{person}{Ronen Eldan}, \bibinfo{person}{Johannes Gehrke}, \bibinfo{person}{Eric Horvitz}, \bibinfo{person}{Ece Kamar}, \bibinfo{person}{Peter Lee}, \bibinfo{person}{Yin~Tat Lee}, \bibinfo{person}{Yuanzhi Li}, \bibinfo{person}{Scott Lundberg}, {et~al\mbox{.}}} \bibinfo{year}{2023}\natexlab{}.
\newblock \showarticletitle{Sparks of artificial general intelligence: Early experiments with gpt-4}.
\newblock \bibinfo{journal}{\emph{arXiv preprint arXiv:2303.12712}} (\bibinfo{year}{2023}).
\newblock


\bibitem[Burns and Beck(1999)]%
        {burns1999feeling}
\bibfield{author}{\bibinfo{person}{David~D Burns} {and} \bibinfo{person}{Aaron~T Beck}.} \bibinfo{year}{1999}\natexlab{}.
\newblock \showarticletitle{Feeling good: The new mood therapy}.
\newblock  (\bibinfo{year}{1999}).
\newblock


\bibitem[Cay(2023)]%
        {yusuf2023how}
\bibfield{author}{\bibinfo{person}{Yusuf Cay}.} \bibinfo{year}{2023}\natexlab{}.
\newblock \bibinfo{title}{How to use {ChatGPT} as your therapist or mental coach?}
\newblock
\newblock
\urldef\tempurl%
\url{https://chatgptplus.blog/use-chatgpt-as-your-therapist-or-mental-coach/}
\showURL{%
\tempurl}


\bibitem[Chandrashekar(2018)]%
        {chandrashekar2018mental}
\bibfield{author}{\bibinfo{person}{Pooja Chandrashekar}.} \bibinfo{year}{2018}\natexlab{}.
\newblock \showarticletitle{Do mental health mobile apps work: evidence and recommendations for designing high-efficacy mental health mobile apps}.
\newblock \bibinfo{journal}{\emph{Mhealth}}  \bibinfo{volume}{4} (\bibinfo{year}{2018}).
\newblock


\bibitem[Channon et~al\mbox{.}(2007)]%
        {channon2007multicenter}
\bibfield{author}{\bibinfo{person}{Sue~J Channon}, \bibinfo{person}{Michelle~V Huws-Thomas}, \bibinfo{person}{Stephen Rollnick}, \bibinfo{person}{Kerenza Hood}, \bibinfo{person}{Rebecca~L Cannings-John}, \bibinfo{person}{Carol Rogers}, {and} \bibinfo{person}{John~W Gregory}.} \bibinfo{year}{2007}\natexlab{}.
\newblock \showarticletitle{A multicenter randomized controlled trial of motivational interviewing in teenagers with diabetes}.
\newblock \bibinfo{journal}{\emph{Diabetes care}} \bibinfo{volume}{30}, \bibinfo{number}{6} (\bibinfo{year}{2007}), \bibinfo{pages}{1390--1395}.
\newblock


\bibitem[Chen et~al\mbox{.}(2023)]%
        {chen2023llm}
\bibfield{author}{\bibinfo{person}{Siyuan Chen}, \bibinfo{person}{Mengyue Wu}, \bibinfo{person}{Kenny~Q Zhu}, \bibinfo{person}{Kunyao Lan}, \bibinfo{person}{Zhiling Zhang}, {and} \bibinfo{person}{Lyuchun Cui}.} \bibinfo{year}{2023}\natexlab{}.
\newblock \showarticletitle{LLM-empowered Chatbots for Psychiatrist and Patient Simulation: Application and Evaluation}.
\newblock \bibinfo{journal}{\emph{arXiv preprint arXiv:2305.13614}} (\bibinfo{year}{2023}).
\newblock


\bibitem[Chen et~al\mbox{.}(2012)]%
        {chen2012effects}
\bibfield{author}{\bibinfo{person}{Shu~Ming Chen}, \bibinfo{person}{Debra Creedy}, \bibinfo{person}{Huey-Shyan Lin}, {and} \bibinfo{person}{Judy Wollin}.} \bibinfo{year}{2012}\natexlab{}.
\newblock \showarticletitle{Effects of motivational interviewing intervention on self-management, psychological and glycemic outcomes in type 2 diabetes: a randomized controlled trial}.
\newblock \bibinfo{journal}{\emph{International journal of nursing studies}} \bibinfo{volume}{49}, \bibinfo{number}{6} (\bibinfo{year}{2012}), \bibinfo{pages}{637--644}.
\newblock


\bibitem[Clark et~al\mbox{.}(2003)]%
        {clark2003cognitive}
\bibfield{author}{\bibinfo{person}{David~M Clark}, \bibinfo{person}{Anke Ehlers}, \bibinfo{person}{Freda McManus}, \bibinfo{person}{Ann Hackmann}, \bibinfo{person}{Melanie Fennell}, \bibinfo{person}{Helen Campbell}, \bibinfo{person}{Teresa Flower}, \bibinfo{person}{Clare Davenport}, {and} \bibinfo{person}{Beverley Louis}.} \bibinfo{year}{2003}\natexlab{}.
\newblock \showarticletitle{Cognitive therapy versus fluoxetine in generalized social phobia: a randomized placebo-controlled trial.}
\newblock \bibinfo{journal}{\emph{Journal of consulting and clinical psychology}} \bibinfo{volume}{71}, \bibinfo{number}{6} (\bibinfo{year}{2003}), \bibinfo{pages}{1058}.
\newblock


\bibitem[Clausen et~al\mbox{.}(2016)]%
        {clausen2016health}
\bibfield{author}{\bibinfo{person}{Whitney Clausen}, \bibinfo{person}{Shinobu Watanabe-Galloway}, \bibinfo{person}{M Bill~Baerentzen}, {and} \bibinfo{person}{Denise~H Britigan}.} \bibinfo{year}{2016}\natexlab{}.
\newblock \showarticletitle{Health literacy among people with serious mental illness}.
\newblock \bibinfo{journal}{\emph{Community mental health journal}} \bibinfo{volume}{52}, \bibinfo{number}{4} (\bibinfo{year}{2016}), \bibinfo{pages}{399--405}.
\newblock


\bibitem[Corey(2013)]%
        {corey2013theory}
\bibfield{author}{\bibinfo{person}{Gerald Corey}.} \bibinfo{year}{2013}\natexlab{}.
\newblock \bibinfo{booktitle}{\emph{Theory and practice of counseling and psychotherapy}}.
\newblock \bibinfo{publisher}{Cengage learning}.
\newblock


\bibitem[Dai et~al\mbox{.}(2023)]%
        {dai2023detecting}
\bibfield{author}{\bibinfo{person}{Ruixuan Dai}, \bibinfo{person}{Thomas Kannampallil}, \bibinfo{person}{Seunghwan Kim}, \bibinfo{person}{Vera Thornton}, \bibinfo{person}{Laura Bierut}, {and} \bibinfo{person}{Chenyang Lu}.} \bibinfo{year}{2023}\natexlab{}.
\newblock \showarticletitle{Detecting Mental Disorders with Wearables: A Large Cohort Study}. In \bibinfo{booktitle}{\emph{Proceedings of the 8th ACM/IEEE Conference on Internet of Things Design and Implementation}}. \bibinfo{pages}{39--51}.
\newblock


\bibitem[Deb et~al\mbox{.}(2023)]%
        {deb2023fill}
\bibfield{author}{\bibinfo{person}{Aniruddha Deb}, \bibinfo{person}{Neeva Oza}, \bibinfo{person}{Sarthak Singla}, \bibinfo{person}{Dinesh Khandelwal}, \bibinfo{person}{Dinesh Garg}, {and} \bibinfo{person}{Parag Singla}.} \bibinfo{year}{2023}\natexlab{}.
\newblock \showarticletitle{Fill in the blank: Exploring and enhancing LLM capabilities for backward reasoning in math word problems}.
\newblock \bibinfo{journal}{\emph{arXiv preprint arXiv:2310.01991}} (\bibinfo{year}{2023}).
\newblock


\bibitem[Dickstein et~al\mbox{.}(2013)]%
        {dickstein2013comparing}
\bibfield{author}{\bibinfo{person}{Benjamin~D Dickstein}, \bibinfo{person}{Kristen~H Walter}, \bibinfo{person}{Jeremiah~A Schumm}, {and} \bibinfo{person}{Kathleen~M Chard}.} \bibinfo{year}{2013}\natexlab{}.
\newblock \showarticletitle{Comparing response to cognitive processing therapy in military veterans with subthreshold and threshold posttraumatic stress disorder}.
\newblock \bibinfo{journal}{\emph{Journal of Traumatic Stress}} \bibinfo{volume}{26}, \bibinfo{number}{6} (\bibinfo{year}{2013}), \bibinfo{pages}{703--709}.
\newblock


\bibitem[DSM-IV-TR.(2000)]%
        {dsm2000diagnostic}
\bibfield{author}{\bibinfo{person}{DSM-IV-TR.}} \bibinfo{year}{2000}\natexlab{}.
\newblock \bibinfo{booktitle}{\emph{Diagnostic and statistical manual of mental disorders}}.
\newblock \bibinfo{publisher}{American Psychiatric Association}.
\newblock


\bibitem[El~ASRS(2009)]%
        {el2009adult}
\bibfield{author}{\bibinfo{person}{Conclusiones El~ASRS}.} \bibinfo{year}{2009}\natexlab{}.
\newblock \showarticletitle{Adult ADHD Self-Report Scale (ASRS-v1. 1) symptom checklist in patients with substance use disorders}.
\newblock \bibinfo{journal}{\emph{Actas Esp Psiquiatr}} \bibinfo{volume}{37}, \bibinfo{number}{6} (\bibinfo{year}{2009}), \bibinfo{pages}{299--305}.
\newblock


\bibitem[Emilsson et~al\mbox{.}(2011)]%
        {emilsson2011cognitive}
\bibfield{author}{\bibinfo{person}{Brynjar Emilsson}, \bibinfo{person}{Gisli Gudjonsson}, \bibinfo{person}{Jon~F Sigurdsson}, \bibinfo{person}{Gisli Baldursson}, \bibinfo{person}{Emil Einarsson}, \bibinfo{person}{Halldora Olafsdottir}, {and} \bibinfo{person}{Susan Young}.} \bibinfo{year}{2011}\natexlab{}.
\newblock \showarticletitle{Cognitive behaviour therapy in medication-treated adults with ADHD and persistent symptoms: a randomized controlled trial}.
\newblock \bibinfo{journal}{\emph{BMC psychiatry}} \bibinfo{volume}{11}, \bibinfo{number}{1} (\bibinfo{year}{2011}), \bibinfo{pages}{1--10}.
\newblock


\bibitem[Explodingtopics(2023)]%
        {SmartphoneMarket}
\bibfield{author}{\bibinfo{person}{Explodingtopics}.} \bibinfo{year}{2023}\natexlab{}.
\newblock \bibinfo{booktitle}{\emph{How Many People Own Smartphones? (2024-2029)}}.
\newblock
\urldef\tempurl%
\url{https://explodingtopics.com/blog/smartphone-stats}
\showURL{%
\tempurl}


\bibitem[Foa et~al\mbox{.}(2005)]%
        {foa2005randomized}
\bibfield{author}{\bibinfo{person}{Edna~B Foa}, \bibinfo{person}{Michael~R Liebowitz}, \bibinfo{person}{Michael~J Kozak}, \bibinfo{person}{Sharon Davies}, \bibinfo{person}{Rafael Campeas}, \bibinfo{person}{Martin~E Franklin}, \bibinfo{person}{Jonathan~D Huppert}, \bibinfo{person}{Kevin Kjernisted}, \bibinfo{person}{Vivienne Rowan}, \bibinfo{person}{Andrew~B Schmidt}, {et~al\mbox{.}}} \bibinfo{year}{2005}\natexlab{}.
\newblock \showarticletitle{Randomized, placebo-controlled trial of exposure and ritual prevention, clomipramine, and their combination in the treatment of obsessive-compulsive disorder}.
\newblock \bibinfo{journal}{\emph{American Journal of psychiatry}} \bibinfo{volume}{162}, \bibinfo{number}{1} (\bibinfo{year}{2005}), \bibinfo{pages}{151--161}.
\newblock


\bibitem[Fu and Fu(2022)]%
        {fu2022distributed}
\bibfield{author}{\bibinfo{person}{Baoyan Fu} {and} \bibinfo{person}{XinXin Fu}.} \bibinfo{year}{2022}\natexlab{}.
\newblock \showarticletitle{Distributed Simulation System for Athletes’ Mental Health in the Internet of Things Environment}.
\newblock \bibinfo{journal}{\emph{Computational Intelligence and Neuroscience}}  \bibinfo{volume}{2022} (\bibinfo{year}{2022}).
\newblock


\bibitem[Gould et~al\mbox{.}(2002)]%
        {gould2002seeking}
\bibfield{author}{\bibinfo{person}{Madelyn~S Gould}, \bibinfo{person}{Jimmie Lou~Harris Munfakh}, \bibinfo{person}{Keri Lubell}, \bibinfo{person}{Marjorie Kleinman}, {and} \bibinfo{person}{Sarah Parker}.} \bibinfo{year}{2002}\natexlab{}.
\newblock \showarticletitle{Seeking help from the internet during adolescence}.
\newblock \bibinfo{journal}{\emph{Journal of the American Academy of Child \& Adolescent Psychiatry}} \bibinfo{volume}{41}, \bibinfo{number}{10} (\bibinfo{year}{2002}), \bibinfo{pages}{1182--1189}.
\newblock


\bibitem[Guze(1995)]%
        {guze1995diagnostic}
\bibfield{author}{\bibinfo{person}{Samuel~B Guze}.} \bibinfo{year}{1995}\natexlab{}.
\newblock \showarticletitle{Diagnostic and statistical manual of mental disorders, (DSM-IV)}.
\newblock \bibinfo{journal}{\emph{American Journal of Psychiatry}} \bibinfo{volume}{152}, \bibinfo{number}{8} (\bibinfo{year}{1995}), \bibinfo{pages}{1228--1228}.
\newblock


\bibitem[Halmi et~al\mbox{.}(2005)]%
        {halmi2005predictors}
\bibfield{author}{\bibinfo{person}{Katherine~A Halmi}, \bibinfo{person}{W~Stewart Agras}, \bibinfo{person}{Scott Crow}, \bibinfo{person}{James Mitchell}, \bibinfo{person}{G~Terence Wilson}, \bibinfo{person}{Susan~W Bryson}, {and} \bibinfo{person}{Helena~C Kraemer}.} \bibinfo{year}{2005}\natexlab{}.
\newblock \showarticletitle{Predictors of treatment acceptance and completion in anorexia nervosa: implications for future study designs}.
\newblock \bibinfo{journal}{\emph{Archives of general psychiatry}} \bibinfo{volume}{62}, \bibinfo{number}{7} (\bibinfo{year}{2005}), \bibinfo{pages}{776--781}.
\newblock


\bibitem[Helfrich et~al\mbox{.}(2008)]%
        {helfrich2008mental}
\bibfield{author}{\bibinfo{person}{Christine~A Helfrich}, \bibinfo{person}{Glenn~T Fujiura}, {and} \bibinfo{person}{Violet Rutkowski-Kmitta}.} \bibinfo{year}{2008}\natexlab{}.
\newblock \showarticletitle{Mental health disorders and functioning of women in domestic violence shelters}.
\newblock \bibinfo{journal}{\emph{Journal of interpersonal violence}} \bibinfo{volume}{23}, \bibinfo{number}{4} (\bibinfo{year}{2008}), \bibinfo{pages}{437--453}.
\newblock


\bibitem[House(2022)]%
        {WhiteHouse2022}
\bibfield{author}{\bibinfo{person}{The~White House}.} \bibinfo{year}{2022}\natexlab{}.
\newblock \bibinfo{booktitle}{\emph{Reducing the Economic Burden of Unmet Mental Health Needs}}.
\newblock
\urldef\tempurl%
\url{https://www.whitehouse.gov/cea/written-materials/2022/05/31/reducing-the-economic-burden-of-unmet-mental-health-needs/}
\showURL{%
\tempurl}


\bibitem[Jiang et~al\mbox{.}(2023)]%
        {jiang2023multimodal}
\bibfield{author}{\bibinfo{person}{Zifan Jiang}, \bibinfo{person}{Salman Seyedi}, \bibinfo{person}{Emily~Lynn Griner}, \bibinfo{person}{Ahmed Abbasi}, \bibinfo{person}{Ali Bahrami~Rad}, \bibinfo{person}{Hyeokhyen Kwon}, \bibinfo{person}{Robert~O Cotes}, {and} \bibinfo{person}{Gari~D Clifford}.} \bibinfo{year}{2023}\natexlab{}.
\newblock \showarticletitle{Multimodal mental health assessment with remote interviews using facial, vocal, linguistic, and cardiovascular patterns}.
\newblock \bibinfo{journal}{\emph{medRxiv}} (\bibinfo{year}{2023}), \bibinfo{pages}{2023--09}.
\newblock


\bibitem[Kertes et~al\mbox{.}(2011)]%
        {kertes2011impact}
\bibfield{author}{\bibinfo{person}{Angela Kertes}, \bibinfo{person}{Henny~A Westra}, \bibinfo{person}{Lynne Angus}, {and} \bibinfo{person}{Madalyn Marcus}.} \bibinfo{year}{2011}\natexlab{}.
\newblock \showarticletitle{The impact of motivational interviewing on client experiences of cognitive behavioral therapy for generalized anxiety disorder}.
\newblock \bibinfo{journal}{\emph{Cognitive and Behavioral Practice}} \bibinfo{volume}{18}, \bibinfo{number}{1} (\bibinfo{year}{2011}), \bibinfo{pages}{55--69}.
\newblock


\bibitem[Kian et~al\mbox{.}(2024)]%
        {kian2024can}
\bibfield{author}{\bibinfo{person}{Mina~J Kian}, \bibinfo{person}{Mingyu Zong}, \bibinfo{person}{Katrin Fischer}, \bibinfo{person}{Abhyuday Singh}, \bibinfo{person}{Anna-Maria Velentza}, \bibinfo{person}{Pau Sang}, \bibinfo{person}{Shriya Upadhyay}, \bibinfo{person}{Anika Gupta}, \bibinfo{person}{Misha~A Faruki}, \bibinfo{person}{Wallace Browning}, {et~al\mbox{.}}} \bibinfo{year}{2024}\natexlab{}.
\newblock \showarticletitle{Can an LLM-Powered Socially Assistive Robot Effectively and Safely Deliver Cognitive Behavioral Therapy? A Study With University Students}.
\newblock \bibinfo{journal}{\emph{arXiv preprint arXiv:2402.17937}} (\bibinfo{year}{2024}).
\newblock


\bibitem[Kroenke et~al\mbox{.}(2001)]%
        {kroenke2001phq}
\bibfield{author}{\bibinfo{person}{Kurt Kroenke}, \bibinfo{person}{Robert~L Spitzer}, {and} \bibinfo{person}{Janet~BW Williams}.} \bibinfo{year}{2001}\natexlab{}.
\newblock \showarticletitle{The PHQ-9: validity of a brief depression severity measure}.
\newblock \bibinfo{journal}{\emph{Journal of general internal medicine}} \bibinfo{volume}{16}, \bibinfo{number}{9} (\bibinfo{year}{2001}), \bibinfo{pages}{606--613}.
\newblock


\bibitem[Kruzan et~al\mbox{.}(2022)]%
        {kruzan2022wanted}
\bibfield{author}{\bibinfo{person}{Kaylee~Payne Kruzan}, \bibinfo{person}{Jonah Meyerhoff}, \bibinfo{person}{Theresa Nguyen}, \bibinfo{person}{Madhu Reddy}, \bibinfo{person}{David~C Mohr}, {and} \bibinfo{person}{Rachel Kornfield}.} \bibinfo{year}{2022}\natexlab{}.
\newblock \showarticletitle{“I Wanted to See How Bad it Was”: Online Self-screening as a Critical Transition Point Among Young Adults with Common Mental Health Conditions}. In \bibinfo{booktitle}{\emph{Proceedings of the 2022 CHI Conference on Human Factors in Computing Systems}}. \bibinfo{pages}{1--16}.
\newblock


\bibitem[Lai et~al\mbox{.}(2023)]%
        {lai2023psy}
\bibfield{author}{\bibinfo{person}{Tin Lai}, \bibinfo{person}{Yukun Shi}, \bibinfo{person}{Zicong Du}, \bibinfo{person}{Jiajie Wu}, \bibinfo{person}{Ken Fu}, \bibinfo{person}{Yichao Dou}, {and} \bibinfo{person}{Ziqi Wang}.} \bibinfo{year}{2023}\natexlab{}.
\newblock \showarticletitle{Psy-llm: Scaling up global mental health psychological services with ai-based large language models}.
\newblock \bibinfo{journal}{\emph{arXiv preprint arXiv:2307.11991}} (\bibinfo{year}{2023}).
\newblock


\bibitem[Liu et~al\mbox{.}(2019)]%
        {liu2019roberta}
\bibfield{author}{\bibinfo{person}{Yinhan Liu}, \bibinfo{person}{Myle Ott}, \bibinfo{person}{Naman Goyal}, \bibinfo{person}{Jingfei Du}, \bibinfo{person}{Mandar Joshi}, \bibinfo{person}{Danqi Chen}, \bibinfo{person}{Omer Levy}, \bibinfo{person}{Mike Lewis}, \bibinfo{person}{Luke Zettlemoyer}, {and} \bibinfo{person}{Veselin Stoyanov}.} \bibinfo{year}{2019}\natexlab{}.
\newblock \showarticletitle{Roberta: A robustly optimized bert pretraining approach}.
\newblock \bibinfo{journal}{\emph{arXiv preprint arXiv:1907.11692}} (\bibinfo{year}{2019}).
\newblock


\bibitem[Liu et~al\mbox{.}(2022)]%
        {liu2022aimse}
\bibfield{author}{\bibinfo{person}{Yanchen Liu}, \bibinfo{person}{Stephen Xia}, \bibinfo{person}{Jingping Nie}, \bibinfo{person}{Peter Wei}, \bibinfo{person}{Zhan Shu}, \bibinfo{person}{Jeffrey~Andrew Chang}, {and} \bibinfo{person}{Xiaofan Jiang}.} \bibinfo{year}{2022}\natexlab{}.
\newblock \showarticletitle{aiMSE: Toward an AI-Based Online Mental Status Examination}.
\newblock \bibinfo{journal}{\emph{IEEE Pervasive Computing}} (\bibinfo{year}{2022}).
\newblock


\bibitem[Marker and Norton(2018)]%
        {marker2018efficacy}
\bibfield{author}{\bibinfo{person}{Isabella Marker} {and} \bibinfo{person}{Peter~J Norton}.} \bibinfo{year}{2018}\natexlab{}.
\newblock \showarticletitle{The efficacy of incorporating motivational interviewing to cognitive behavior therapy for anxiety disorders: A review and meta-analysis}.
\newblock \bibinfo{journal}{\emph{Clinical Psychology Review}}  \bibinfo{volume}{62} (\bibinfo{year}{2018}), \bibinfo{pages}{1--10}.
\newblock


\bibitem[Matthews et~al\mbox{.}(2015)]%
        {matthews2015situ}
\bibfield{author}{\bibinfo{person}{Mark Matthews}, \bibinfo{person}{Stephen Voida}, \bibinfo{person}{Saeed Abdullah}, \bibinfo{person}{Gavin Doherty}, \bibinfo{person}{Tanzeem Choudhury}, \bibinfo{person}{Sangha Im}, {and} \bibinfo{person}{Geri Gay}.} \bibinfo{year}{2015}\natexlab{}.
\newblock \showarticletitle{In situ design for mental illness: Considering the pathology of bipolar disorder in mhealth design}. In \bibinfo{booktitle}{\emph{Proceedings of the 17th International Conference on Human-Computer Interaction with Mobile Devices and Services}}. \bibinfo{pages}{86--97}.
\newblock


\bibitem[Merlo et~al\mbox{.}(2010)]%
        {merlo2010cognitive}
\bibfield{author}{\bibinfo{person}{Lisa~J Merlo}, \bibinfo{person}{Eric~A Storch}, \bibinfo{person}{Heather~D Lehmkuhl}, \bibinfo{person}{Marni~L Jacob}, \bibinfo{person}{Tanya~K Murphy}, \bibinfo{person}{Wayne~K Goodman}, {and} \bibinfo{person}{Gary~R Geffken}.} \bibinfo{year}{2010}\natexlab{}.
\newblock \showarticletitle{Cognitive-behavioral therapy plus motivational interviewing improves outcome for pediatric obsessive-compulsive disorder: A preliminary study}.
\newblock \bibinfo{journal}{\emph{Cognitive Behaviour Therapy}} \bibinfo{volume}{39}, \bibinfo{number}{1} (\bibinfo{year}{2010}), \bibinfo{pages}{24--27}.
\newblock


\bibitem[Miller et~al\mbox{.}(2011)]%
        {miller2011effectiveness}
\bibfield{author}{\bibinfo{person}{Lynn~D Miller}, \bibinfo{person}{Aviva Laye-Gindhu}, \bibinfo{person}{Joanna~L Bennett}, \bibinfo{person}{Yan Liu}, \bibinfo{person}{Stephenie Gold}, \bibinfo{person}{John~S March}, \bibinfo{person}{Brent~F Olson}, {and} \bibinfo{person}{Vanessa~E Waechtler}.} \bibinfo{year}{2011}\natexlab{}.
\newblock \showarticletitle{An effectiveness study of a culturally enriched school-based CBT anxiety prevention program}.
\newblock \bibinfo{journal}{\emph{Journal of Clinical Child \& Adolescent Psychology}} \bibinfo{volume}{40}, \bibinfo{number}{4} (\bibinfo{year}{2011}), \bibinfo{pages}{618--629}.
\newblock


\bibitem[Miller and Rollnick(2012)]%
        {miller2012motivational}
\bibfield{author}{\bibinfo{person}{William~R Miller} {and} \bibinfo{person}{Stephen Rollnick}.} \bibinfo{year}{2012}\natexlab{}.
\newblock \bibinfo{booktitle}{\emph{Motivational interviewing: Helping people change}}.
\newblock \bibinfo{publisher}{Guilford press}.
\newblock


\bibitem[Mishra(2019)]%
        {mishra2019sensing}
\bibfield{author}{\bibinfo{person}{Varun Mishra}.} \bibinfo{year}{2019}\natexlab{}.
\newblock \showarticletitle{From sensing to intervention for mental and behavioral health}. In \bibinfo{booktitle}{\emph{Adjunct Proceedings of the 2019 ACM International Joint Conference on Pervasive and Ubiquitous Computing and Proceedings of the 2019 ACM International Symposium on Wearable Computers}}. \bibinfo{pages}{388--392}.
\newblock


\bibitem[Mohadisdudis and Ali(2014)]%
        {mohadisdudis2014study}
\bibfield{author}{\bibinfo{person}{Hazwani~Mohd Mohadisdudis} {and} \bibinfo{person}{Nazlena~Mohamad Ali}.} \bibinfo{year}{2014}\natexlab{}.
\newblock \showarticletitle{A study of smartphone usage and barriers among the elderly}. In \bibinfo{booktitle}{\emph{2014 3rd international conference on user science and engineering (i-USEr)}}. IEEE, \bibinfo{pages}{109--114}.
\newblock


\bibitem[Morshed et~al\mbox{.}(2019)]%
        {morshed2019prediction}
\bibfield{author}{\bibinfo{person}{Mehrab~Bin Morshed}, \bibinfo{person}{Koustuv Saha}, \bibinfo{person}{Richard Li}, \bibinfo{person}{Sidney~K D'Mello}, \bibinfo{person}{Munmun De~Choudhury}, \bibinfo{person}{Gregory~D Abowd}, {and} \bibinfo{person}{Thomas Pl{\"o}tz}.} \bibinfo{year}{2019}\natexlab{}.
\newblock \showarticletitle{Prediction of mood instability with passive sensing}.
\newblock \bibinfo{journal}{\emph{Proceedings of the ACM on Interactive, Mobile, Wearable and Ubiquitous Technologies}} \bibinfo{volume}{3}, \bibinfo{number}{3} (\bibinfo{year}{2019}), \bibinfo{pages}{1--21}.
\newblock


\bibitem[Moss-Morris et~al\mbox{.}(2013)]%
        {moss2013randomized}
\bibfield{author}{\bibinfo{person}{Rona Moss-Morris}, \bibinfo{person}{Laura Dennison}, \bibinfo{person}{Sabine Landau}, \bibinfo{person}{Lucy Yardley}, \bibinfo{person}{Eli Silber}, {and} \bibinfo{person}{Trudie Chalder}.} \bibinfo{year}{2013}\natexlab{}.
\newblock \showarticletitle{A randomized controlled trial of cognitive behavioral therapy (CBT) for adjusting to multiple sclerosis (the saMS trial): does CBT work and for whom does it work?}
\newblock \bibinfo{journal}{\emph{Journal of consulting and clinical psychology}} \bibinfo{volume}{81}, \bibinfo{number}{2} (\bibinfo{year}{2013}), \bibinfo{pages}{251}.
\newblock


\bibitem[Naar and Safren(2017)]%
        {naar2017motivational}
\bibfield{author}{\bibinfo{person}{Sylvie Naar} {and} \bibinfo{person}{Steven~A Safren}.} \bibinfo{year}{2017}\natexlab{}.
\newblock \bibinfo{booktitle}{\emph{Motivational interviewing and CBT: Combining strategies for maximum effectiveness}}.
\newblock \bibinfo{publisher}{Guilford Publications}.
\newblock


\bibitem[Nie et~al\mbox{.}(2021)]%
        {nie2021spiders+}
\bibfield{author}{\bibinfo{person}{Jingping Nie}, \bibinfo{person}{Yanchen Liu}, \bibinfo{person}{Yigong Hu}, \bibinfo{person}{Yuanyuting Wang}, \bibinfo{person}{Stephen Xia}, \bibinfo{person}{Matthias Preindl}, {and} \bibinfo{person}{Xiaofan Jiang}.} \bibinfo{year}{2021}\natexlab{}.
\newblock \showarticletitle{SPIDERS+: A light-weight, wireless, and low-cost glasses-based wearable platform for emotion sensing and bio-signal acquisition}.
\newblock \bibinfo{journal}{\emph{Pervasive and Mobile Computing}}  \bibinfo{volume}{75} (\bibinfo{year}{2021}), \bibinfo{pages}{101424}.
\newblock


\bibitem[Nie et~al\mbox{.}(2022)]%
        {nie2022conversational}
\bibfield{author}{\bibinfo{person}{Jingping Nie}, \bibinfo{person}{Hanya Shao}, \bibinfo{person}{Minghui Zhao}, \bibinfo{person}{Stephen Xia}, \bibinfo{person}{Matthias Preindl}, {and} \bibinfo{person}{Xiaofan Jiang}.} \bibinfo{year}{2022}\natexlab{}.
\newblock \showarticletitle{Conversational AI Therapist for Daily Function Screening in Home Environments}. In \bibinfo{booktitle}{\emph{Proceedings of the 1st ACM International Workshop on Intelligent Acoustic Systems and Applications}}. \bibinfo{pages}{31--36}.
\newblock


\bibitem[Nori et~al\mbox{.}(2023)]%
        {nori2023capabilities}
\bibfield{author}{\bibinfo{person}{Harsha Nori}, \bibinfo{person}{Nicholas King}, \bibinfo{person}{Scott~Mayer McKinney}, \bibinfo{person}{Dean Carignan}, {and} \bibinfo{person}{Eric Horvitz}.} \bibinfo{year}{2023}\natexlab{}.
\newblock \showarticletitle{Capabilities of gpt-4 on medical challenge problems}.
\newblock \bibinfo{journal}{\emph{arXiv preprint arXiv:2303.13375}} (\bibinfo{year}{2023}).
\newblock


\bibitem[on~Evidence-Based~Practice et~al\mbox{.}(2006)]%
        {apa2006evidence}
\bibfield{author}{\bibinfo{person}{APA Presidential Task~Force on Evidence-Based~Practice} {et~al\mbox{.}}} \bibinfo{year}{2006}\natexlab{}.
\newblock \showarticletitle{Evidence-based practice in psychology}.
\newblock \bibinfo{journal}{\emph{The American Psychologist}} \bibinfo{volume}{61}, \bibinfo{number}{4} (\bibinfo{year}{2006}), \bibinfo{pages}{271--285}.
\newblock


\bibitem[OpenAI(2023a)]%
        {openai2023gpt35}
\bibfield{author}{\bibinfo{person}{OpenAI}.} \bibinfo{year}{2023}\natexlab{a}.
\newblock \bibinfo{title}{GPT-3.5 Turbo}.
\newblock \bibinfo{howpublished}{\url{https://platform.openai.com/docs/models/gpt-3-5-turbo}}.
\newblock
\newblock
\shownote{Accessed: 2024-02-06}.


\bibitem[OpenAI(2023b)]%
        {openai2023gpt4}
\bibfield{author}{\bibinfo{person}{OpenAI}.} \bibinfo{year}{2023}\natexlab{b}.
\newblock \bibinfo{title}{GPT-4 and GPT-4 Turbo}.
\newblock \bibinfo{howpublished}{\url{https://platform.openai.com/docs/models/gpt-4-and-gpt-4-turbo}}.
\newblock
\newblock
\shownote{Accessed: 2024-02-06}.


\bibitem[OpenAI(2023c)]%
        {gpt35finetune}
\bibfield{author}{\bibinfo{person}{OpenAI}.} \bibinfo{year}{2023}\natexlab{c}.
\newblock \bibinfo{title}{What models can be fine-tuned?}
\newblock \bibinfo{howpublished}{\url{https://platform.openai.com/docs/guides/fine-tuning}}.
\newblock
\newblock
\shownote{Accessed: 2024-02-06}.


\bibitem[Pendse et~al\mbox{.}(2021)]%
        {pendse2021can}
\bibfield{author}{\bibinfo{person}{Sachin~R Pendse}, \bibinfo{person}{Amit Sharma}, \bibinfo{person}{Aditya Vashistha}, \bibinfo{person}{Munmun De~Choudhury}, {and} \bibinfo{person}{Neha Kumar}.} \bibinfo{year}{2021}\natexlab{}.
\newblock \showarticletitle{“Can I Not Be Suicidal on a Sunday?”: Understanding Technology-Mediated Pathways to Mental Health Support}. In \bibinfo{booktitle}{\emph{Proceedings of the 2021 CHI Conference on Human Factors in Computing Systems}}. \bibinfo{pages}{1--16}.
\newblock


\bibitem[Pichai and Hassabis(2024)]%
        {google2024gemini15}
\bibfield{author}{\bibinfo{person}{Sundar Pichai} {and} \bibinfo{person}{Demis Hassabis}.} \bibinfo{year}{2024}\natexlab{}.
\newblock \bibinfo{title}{Our next-generation model: Gemini 1.5}.
\newblock \bibinfo{howpublished}{\url{https://blog.google/technology/ai/google-gemini-next-generation-model-february-2024/\#sundar-note}}.
\newblock
\newblock
\shownote{Accessed: 2024-03-06}.


\bibitem[Radhakrishnan et~al\mbox{.}(2023)]%
        {radhakrishnan2023question}
\bibfield{author}{\bibinfo{person}{Ansh Radhakrishnan}, \bibinfo{person}{Karina Nguyen}, \bibinfo{person}{Anna Chen}, \bibinfo{person}{Carol Chen}, \bibinfo{person}{Carson Denison}, \bibinfo{person}{Danny Hernandez}, \bibinfo{person}{Esin Durmus}, \bibinfo{person}{Evan Hubinger}, \bibinfo{person}{Jackson Kernion}, \bibinfo{person}{Kamil{\.e} Luko{\v{s}}i{\=u}t{\.e}}, {et~al\mbox{.}}} \bibinfo{year}{2023}\natexlab{}.
\newblock \showarticletitle{Question decomposition improves the faithfulness of model-generated reasoning}.
\newblock \bibinfo{journal}{\emph{arXiv preprint arXiv:2307.11768}} (\bibinfo{year}{2023}).
\newblock


\bibitem[Radwan et~al\mbox{.}(2024)]%
        {radwan2024predictive}
\bibfield{author}{\bibinfo{person}{Ahmad Radwan}, \bibinfo{person}{Mohannad Amarneh}, \bibinfo{person}{Hussam Alawneh}, \bibinfo{person}{Huthaifa~I Ashqar}, \bibinfo{person}{Anas AlSobeh}, {and} \bibinfo{person}{Aws Abed Al~Raheem Magableh}.} \bibinfo{year}{2024}\natexlab{}.
\newblock \showarticletitle{Predictive Analytics in Mental Health Leveraging LLM Embeddings and Machine Learning Models for Social Media Analysis}.
\newblock \bibinfo{journal}{\emph{International Journal of Web Services Research (IJWSR)}} \bibinfo{volume}{21}, \bibinfo{number}{1} (\bibinfo{year}{2024}), \bibinfo{pages}{1--22}.
\newblock


\bibitem[Robins and Rosenthal(2011)]%
        {robins2011dialectical}
\bibfield{author}{\bibinfo{person}{Clive~J Robins} {and} \bibinfo{person}{M~Zachary Rosenthal}.} \bibinfo{year}{2011}\natexlab{}.
\newblock \showarticletitle{Dialectical behavior therapy}.
\newblock \bibinfo{journal}{\emph{Acceptance and mindfulness in cognitive behavior therapy: Understanding and applying the new therapies}} (\bibinfo{year}{2011}), \bibinfo{pages}{164--192}.
\newblock


\bibitem[Roy-Byrne et~al\mbox{.}(2005)]%
        {roy2005randomized}
\bibfield{author}{\bibinfo{person}{Peter~P Roy-Byrne}, \bibinfo{person}{Michelle~G Craske}, \bibinfo{person}{Murray~B Stein}, \bibinfo{person}{Greer Sullivan}, \bibinfo{person}{Alexander Bystritsky}, \bibinfo{person}{Wayne Katon}, \bibinfo{person}{Daniela Golinelli}, {and} \bibinfo{person}{Cathy~D Sherbourne}.} \bibinfo{year}{2005}\natexlab{}.
\newblock \showarticletitle{A randomized effectiveness trial of cognitive-behavioral therapy and medication for primary care panic disorder}.
\newblock \bibinfo{journal}{\emph{Archives of General Psychiatry}} \bibinfo{volume}{62}, \bibinfo{number}{3} (\bibinfo{year}{2005}), \bibinfo{pages}{290--298}.
\newblock


\bibitem[Sabour et~al\mbox{.}(2022)]%
        {sabour2022chatbots}
\bibfield{author}{\bibinfo{person}{Sahand Sabour}, \bibinfo{person}{Wen Zhang}, \bibinfo{person}{Xiyao Xiao}, \bibinfo{person}{Yuwei Zhang}, \bibinfo{person}{Yinhe Zheng}, \bibinfo{person}{Jiaxin Wen}, \bibinfo{person}{Jialu Zhao}, {and} \bibinfo{person}{Minlie Huang}.} \bibinfo{year}{2022}\natexlab{}.
\newblock \showarticletitle{Chatbots for Mental Health Support: Exploring the Impact of Emohaa on Reducing Mental Distress in China}.
\newblock \bibinfo{journal}{\emph{arXiv preprint arXiv:2209.10183}} (\bibinfo{year}{2022}).
\newblock


\bibitem[Salekin et~al\mbox{.}(2017)]%
        {salekin2017distant}
\bibfield{author}{\bibinfo{person}{Asif Salekin}, \bibinfo{person}{Zeya Chen}, \bibinfo{person}{Mohsin~Y Ahmed}, \bibinfo{person}{John Lach}, \bibinfo{person}{Donna Metz}, \bibinfo{person}{Kayla De~La~Haye}, \bibinfo{person}{Brooke Bell}, {and} \bibinfo{person}{John~A Stankovic}.} \bibinfo{year}{2017}\natexlab{}.
\newblock \showarticletitle{Distant emotion recognition}.
\newblock \bibinfo{journal}{\emph{Proceedings of the ACM on Interactive, Mobile, Wearable and Ubiquitous Technologies}} \bibinfo{volume}{1}, \bibinfo{number}{3} (\bibinfo{year}{2017}), \bibinfo{pages}{1--25}.
\newblock


\bibitem[Schroeder et~al\mbox{.}(2018)]%
        {schroeder2018pocket}
\bibfield{author}{\bibinfo{person}{Jessica Schroeder}, \bibinfo{person}{Chelsey Wilkes}, \bibinfo{person}{Kael Rowan}, \bibinfo{person}{Arturo Toledo}, \bibinfo{person}{Ann Paradiso}, \bibinfo{person}{Mary Czerwinski}, \bibinfo{person}{Gloria Mark}, {and} \bibinfo{person}{Marsha~M Linehan}.} \bibinfo{year}{2018}\natexlab{}.
\newblock \showarticletitle{Pocket skills: A conversational mobile web app to support dialectical behavioral therapy}. In \bibinfo{booktitle}{\emph{Proceedings of the 2018 CHI Conference on Human Factors in Computing Systems}}. \bibinfo{pages}{1--15}.
\newblock


\bibitem[Scott and Presmanes(2001)]%
        {scott2001reliability}
\bibfield{author}{\bibinfo{person}{Roger~L Scott} {and} \bibinfo{person}{Willa~S Presmanes}.} \bibinfo{year}{2001}\natexlab{}.
\newblock \showarticletitle{Reliability and validity of the daily living activities scale: A functional assessment measure for severe mental disorders}.
\newblock \bibinfo{journal}{\emph{Research on Social Work Practice}} \bibinfo{volume}{11}, \bibinfo{number}{3} (\bibinfo{year}{2001}), \bibinfo{pages}{373--389}.
\newblock


\bibitem[Screening(2021)]%
        {OnlineTest}
\bibfield{author}{\bibinfo{person}{MHA Screening}.} \bibinfo{year}{2021}\natexlab{}.
\newblock \bibinfo{title}{{Take a Mental Health Test}}.
\newblock \bibinfo{howpublished}{\url{https://screening.mhanational.org/screening-tools/}}.
\newblock
\newblock
\shownote{[Online; accessed 24-July-2022]}.


\bibitem[Sharan et~al\mbox{.}(2023)]%
        {sharan2023llm}
\bibfield{author}{\bibinfo{person}{SP Sharan}, \bibinfo{person}{Francesco Pittaluga}, \bibinfo{person}{Manmohan Chandraker}, {et~al\mbox{.}}} \bibinfo{year}{2023}\natexlab{}.
\newblock \showarticletitle{Llm-assist: Enhancing closed-loop planning with language-based reasoning}.
\newblock \bibinfo{journal}{\emph{arXiv preprint arXiv:2401.00125}} (\bibinfo{year}{2023}).
\newblock


\bibitem[Shin et~al\mbox{.}(2022)]%
        {shin2022chatbots}
\bibfield{author}{\bibinfo{person}{Joongi Shin}, \bibinfo{person}{Michael~A Hedderich}, \bibinfo{person}{Andr{\'e}S Lucero}, {and} \bibinfo{person}{Antti Oulasvirta}.} \bibinfo{year}{2022}\natexlab{}.
\newblock \showarticletitle{Chatbots Facilitating Consensus-Building in Asynchronous Co-Design}. In \bibinfo{booktitle}{\emph{Proceedings of the 35th Annual ACM Symposium on User Interface Software and Technology}}. \bibinfo{pages}{1--13}.
\newblock


\bibitem[Singhal et~al\mbox{.}(2023a)]%
        {singhal2023large}
\bibfield{author}{\bibinfo{person}{Karan Singhal}, \bibinfo{person}{Shekoofeh Azizi}, \bibinfo{person}{Tao Tu}, \bibinfo{person}{S~Sara Mahdavi}, \bibinfo{person}{Jason Wei}, \bibinfo{person}{Hyung~Won Chung}, \bibinfo{person}{Nathan Scales}, \bibinfo{person}{Ajay Tanwani}, \bibinfo{person}{Heather Cole-Lewis}, \bibinfo{person}{Stephen Pfohl}, {et~al\mbox{.}}} \bibinfo{year}{2023}\natexlab{a}.
\newblock \showarticletitle{Large language models encode clinical knowledge}.
\newblock \bibinfo{journal}{\emph{Nature}} \bibinfo{volume}{620}, \bibinfo{number}{7972} (\bibinfo{year}{2023}), \bibinfo{pages}{172--180}.
\newblock


\bibitem[Singhal et~al\mbox{.}(2023b)]%
        {singhal2023towards}
\bibfield{author}{\bibinfo{person}{Karan Singhal}, \bibinfo{person}{Tao Tu}, \bibinfo{person}{Juraj Gottweis}, \bibinfo{person}{Rory Sayres}, \bibinfo{person}{Ellery Wulczyn}, \bibinfo{person}{Le Hou}, \bibinfo{person}{Kevin Clark}, \bibinfo{person}{Stephen Pfohl}, \bibinfo{person}{Heather Cole-Lewis}, \bibinfo{person}{Darlene Neal}, {et~al\mbox{.}}} \bibinfo{year}{2023}\natexlab{b}.
\newblock \showarticletitle{Towards expert-level medical question answering with large language models}.
\newblock \bibinfo{journal}{\emph{arXiv preprint arXiv:2305.09617}} (\bibinfo{year}{2023}).
\newblock


\bibitem[Sokol and Fox(2019)]%
        {sokol2019comprehensive}
\bibfield{author}{\bibinfo{person}{Leslie Sokol} {and} \bibinfo{person}{Marci Fox}.} \bibinfo{year}{2019}\natexlab{}.
\newblock \bibinfo{booktitle}{\emph{The comprehensive clinician's guide to cognitive behavioral therapy}}.
\newblock \bibinfo{publisher}{Pesi}.
\newblock


\bibitem[Spring(2007)]%
        {spring2007evidence}
\bibfield{author}{\bibinfo{person}{Bonnie Spring}.} \bibinfo{year}{2007}\natexlab{}.
\newblock \showarticletitle{Evidence-based practice in clinical psychology: What it is, why it matters; what you need to know}.
\newblock \bibinfo{journal}{\emph{Journal of clinical psychology}} \bibinfo{volume}{63}, \bibinfo{number}{7} (\bibinfo{year}{2007}), \bibinfo{pages}{611--631}.
\newblock


\bibitem[Statista(2023)]%
        {SpeakerMarket}
\bibfield{author}{\bibinfo{person}{Statista}.} \bibinfo{year}{2023}\natexlab{}.
\newblock \bibinfo{booktitle}{\emph{Sales volume of the smart speakers industry Worldwide 2018-2028}}.
\newblock
\urldef\tempurl%
\url{https://www.statista.com/forecasts/1367982/smart-speaker-market-volume-worldwide}
\showURL{%
\tempurl}


\bibitem[Talevi et~al\mbox{.}(2020)]%
        {talevi2020mental}
\bibfield{author}{\bibinfo{person}{Dalila Talevi}, \bibinfo{person}{Valentina Socci}, \bibinfo{person}{Margherita Carai}, \bibinfo{person}{Giulia Carnaghi}, \bibinfo{person}{Serena Faleri}, \bibinfo{person}{Edoardo Trebbi}, \bibinfo{person}{Arianna di Bernardo}, \bibinfo{person}{Francesco Capelli}, {and} \bibinfo{person}{Francesca Pacitti}.} \bibinfo{year}{2020}\natexlab{}.
\newblock \showarticletitle{Mental health outcomes of the CoViD-19 pandemic}.
\newblock \bibinfo{journal}{\emph{Rivista di psichiatria}} \bibinfo{volume}{55}, \bibinfo{number}{3} (\bibinfo{year}{2020}), \bibinfo{pages}{137--144}.
\newblock


\bibitem[Tan et~al\mbox{.}(2014)]%
        {tan2014preventing}
\bibfield{author}{\bibinfo{person}{Leona Tan}, \bibinfo{person}{Min-Jung Wang}, \bibinfo{person}{Matthew Modini}, \bibinfo{person}{Sadhbh Joyce}, \bibinfo{person}{Arnstein Mykletun}, \bibinfo{person}{Helen Christensen}, {and} \bibinfo{person}{Samuel~B Harvey}.} \bibinfo{year}{2014}\natexlab{}.
\newblock \showarticletitle{Preventing the development of depression at work: a systematic review and meta-analysis of universal interventions in the workplace}.
\newblock \bibinfo{journal}{\emph{BMC medicine}} \bibinfo{volume}{12}, \bibinfo{number}{1} (\bibinfo{year}{2014}), \bibinfo{pages}{1--11}.
\newblock


\bibitem[Tlachac et~al\mbox{.}(2022)]%
        {tlachac2022studentsadd}
\bibfield{author}{\bibinfo{person}{ML Tlachac}, \bibinfo{person}{Ricardo Flores}, \bibinfo{person}{Miranda Reisch}, \bibinfo{person}{Rimsha Kayastha}, \bibinfo{person}{Nina Taurich}, \bibinfo{person}{Veronica Melican}, \bibinfo{person}{Connor Bruneau}, \bibinfo{person}{Hunter Caouette}, \bibinfo{person}{Joshua Lovering}, \bibinfo{person}{Ermal Toto}, {et~al\mbox{.}}} \bibinfo{year}{2022}\natexlab{}.
\newblock \showarticletitle{StudentSADD: Rapid mobile depression and suicidal ideation screening of college students during the coronavirus pandemic}.
\newblock \bibinfo{journal}{\emph{Proceedings of the ACM on Interactive, Mobile, Wearable and Ubiquitous Technologies}} \bibinfo{volume}{6}, \bibinfo{number}{2} (\bibinfo{year}{2022}), \bibinfo{pages}{1--32}.
\newblock


\bibitem[Torous et~al\mbox{.}(2018)]%
        {torous2018mental}
\bibfield{author}{\bibinfo{person}{John Torous}, \bibinfo{person}{Hannah Wisniewski}, \bibinfo{person}{Gang Liu}, \bibinfo{person}{Matcheri Keshavan}, {et~al\mbox{.}}} \bibinfo{year}{2018}\natexlab{}.
\newblock \showarticletitle{Mental health mobile phone app usage, concerns, and benefits among psychiatric outpatients: comparative survey study}.
\newblock \bibinfo{journal}{\emph{JMIR mental health}} \bibinfo{volume}{5}, \bibinfo{number}{4} (\bibinfo{year}{2018}), \bibinfo{pages}{e11715}.
\newblock


\bibitem[Touvron et~al\mbox{.}(2023)]%
        {touvron2023llama}
\bibfield{author}{\bibinfo{person}{Hugo Touvron}, \bibinfo{person}{Louis Martin}, \bibinfo{person}{Kevin Stone}, \bibinfo{person}{Peter Albert}, \bibinfo{person}{Amjad Almahairi}, \bibinfo{person}{Yasmine Babaei}, \bibinfo{person}{Nikolay Bashlykov}, \bibinfo{person}{Soumya Batra}, \bibinfo{person}{Prajjwal Bhargava}, \bibinfo{person}{Shruti Bhosale}, {et~al\mbox{.}}} \bibinfo{year}{2023}\natexlab{}.
\newblock \showarticletitle{Llama 2: Open foundation and fine-tuned chat models}.
\newblock \bibinfo{journal}{\emph{arXiv preprint arXiv:2307.09288}} (\bibinfo{year}{2023}).
\newblock


\bibitem[Trzepacz and Baker(1993)]%
        {trzepacz1993psychiatric}
\bibfield{author}{\bibinfo{person}{Paula~T Trzepacz} {and} \bibinfo{person}{Robert~W Baker}.} \bibinfo{year}{1993}\natexlab{}.
\newblock \bibinfo{booktitle}{\emph{The psychiatric mental status examination}}.
\newblock \bibinfo{publisher}{Oxford University Press}.
\newblock


\bibitem[van Heerden et~al\mbox{.}(2023)]%
        {van2023global}
\bibfield{author}{\bibinfo{person}{Alastair~C van Heerden}, \bibinfo{person}{Julia~R Pozuelo}, {and} \bibinfo{person}{Brandon~A Kohrt}.} \bibinfo{year}{2023}\natexlab{}.
\newblock \showarticletitle{Global mental health services and the impact of artificial intelligence--powered large language models}.
\newblock \bibinfo{journal}{\emph{JAMA psychiatry}} \bibinfo{volume}{80}, \bibinfo{number}{7} (\bibinfo{year}{2023}), \bibinfo{pages}{662--664}.
\newblock


\bibitem[Vega et~al\mbox{.}(2018)]%
        {vega2018back}
\bibfield{author}{\bibinfo{person}{Julio Vega}, \bibinfo{person}{Samuel Couth}, \bibinfo{person}{Ellen Poliakoff}, \bibinfo{person}{Sonja Kotz}, \bibinfo{person}{Matthew Sullivan}, \bibinfo{person}{Caroline Jay}, \bibinfo{person}{Markel Vigo}, {and} \bibinfo{person}{Simon Harper}.} \bibinfo{year}{2018}\natexlab{}.
\newblock \showarticletitle{Back to analogue: Self-reporting for Parkinson's Disease}. In \bibinfo{booktitle}{\emph{Proceedings of the 2018 CHI conference on human factors in computing systems}}. \bibinfo{pages}{1--13}.
\newblock


\bibitem[Waisberg et~al\mbox{.}(2023)]%
        {waisberg2023gpt}
\bibfield{author}{\bibinfo{person}{Ethan Waisberg}, \bibinfo{person}{Joshua Ong}, \bibinfo{person}{Mouayad Masalkhi}, \bibinfo{person}{Sharif~Amit Kamran}, \bibinfo{person}{Nasif Zaman}, \bibinfo{person}{Prithul Sarker}, \bibinfo{person}{Andrew~G Lee}, {and} \bibinfo{person}{Alireza Tavakkoli}.} \bibinfo{year}{2023}\natexlab{}.
\newblock \showarticletitle{GPT-4: a new era of artificial intelligence in medicine}.
\newblock \bibinfo{journal}{\emph{Irish Journal of Medical Science (1971-)}} (\bibinfo{year}{2023}), \bibinfo{pages}{1--4}.
\newblock


\bibitem[Walsh et~al\mbox{.}(2004)]%
        {walsh2004treatment}
\bibfield{author}{\bibinfo{person}{B~Timothy Walsh}, \bibinfo{person}{Christopher~G Fairburn}, \bibinfo{person}{Diane Mickley}, \bibinfo{person}{Robyn Sysko}, {and} \bibinfo{person}{Michael~K Parides}.} \bibinfo{year}{2004}\natexlab{}.
\newblock \showarticletitle{Treatment of bulimia nervosa in a primary care setting}.
\newblock \bibinfo{journal}{\emph{American Journal of Psychiatry}} \bibinfo{volume}{161}, \bibinfo{number}{3} (\bibinfo{year}{2004}), \bibinfo{pages}{556--561}.
\newblock


\bibitem[Xu et~al\mbox{.}(2023a)]%
        {xu2023penetrative}
\bibfield{author}{\bibinfo{person}{Huatao Xu}, \bibinfo{person}{Liying Han}, \bibinfo{person}{Mo Li}, {and} \bibinfo{person}{Mani Srivastava}.} \bibinfo{year}{2023}\natexlab{a}.
\newblock \showarticletitle{Penetrative ai: Making llms comprehend the physical world}.
\newblock \bibinfo{journal}{\emph{arXiv preprint arXiv:2310.09605}} (\bibinfo{year}{2023}).
\newblock


\bibitem[Xu et~al\mbox{.}(2023b)]%
        {xu2023leveraging}
\bibfield{author}{\bibinfo{person}{Xuhai Xu}, \bibinfo{person}{Bingshen Yao}, \bibinfo{person}{Yuanzhe Dong}, \bibinfo{person}{Hong Yu}, \bibinfo{person}{James Hendler}, \bibinfo{person}{Anind~K Dey}, {and} \bibinfo{person}{Dakuo Wang}.} \bibinfo{year}{2023}\natexlab{b}.
\newblock \showarticletitle{Leveraging large language models for mental health prediction via online text data}.
\newblock \bibinfo{journal}{\emph{arXiv preprint arXiv:2307.14385}} (\bibinfo{year}{2023}).
\newblock


\bibitem[Yin et~al\mbox{.}(2023)]%
        {yin2023survey}
\bibfield{author}{\bibinfo{person}{Shukang Yin}, \bibinfo{person}{Chaoyou Fu}, \bibinfo{person}{Sirui Zhao}, \bibinfo{person}{Ke Li}, \bibinfo{person}{Xing Sun}, \bibinfo{person}{Tong Xu}, {and} \bibinfo{person}{Enhong Chen}.} \bibinfo{year}{2023}\natexlab{}.
\newblock \showarticletitle{A Survey on Multimodal Large Language Models}.
\newblock \bibinfo{journal}{\emph{arXiv preprint arXiv:2306.13549}} (\bibinfo{year}{2023}).
\newblock


\bibitem[You et~al\mbox{.}(2023)]%
        {you2023idealgpt}
\bibfield{author}{\bibinfo{person}{Haoxuan You}, \bibinfo{person}{Rui Sun}, \bibinfo{person}{Zhecan Wang}, \bibinfo{person}{Long Chen}, \bibinfo{person}{Gengyu Wang}, \bibinfo{person}{Hammad~A Ayyubi}, \bibinfo{person}{Kai-Wei Chang}, {and} \bibinfo{person}{Shih-Fu Chang}.} \bibinfo{year}{2023}\natexlab{}.
\newblock \showarticletitle{IdealGPT: Iteratively Decomposing Vision and Language Reasoning via Large Language Models}.
\newblock \bibinfo{journal}{\emph{arXiv preprint arXiv:2305.14985}} (\bibinfo{year}{2023}).
\newblock


\bibitem[Yunxiang et~al\mbox{.}(2023)]%
        {yunxiang2023chatdoctor}
\bibfield{author}{\bibinfo{person}{Li Yunxiang}, \bibinfo{person}{Li Zihan}, \bibinfo{person}{Zhang Kai}, \bibinfo{person}{Dan Ruilong}, {and} \bibinfo{person}{Zhang You}.} \bibinfo{year}{2023}\natexlab{}.
\newblock \showarticletitle{Chatdoctor: A medical chat model fine-tuned on llama model using medical domain knowledge}.
\newblock \bibinfo{journal}{\emph{arXiv preprint arXiv:2303.14070}} (\bibinfo{year}{2023}).
\newblock


\bibitem[Zhou et~al\mbox{.}(2018)]%
        {zhou2018emotional}
\bibfield{author}{\bibinfo{person}{Hao Zhou}, \bibinfo{person}{Minlie Huang}, \bibinfo{person}{Tianyang Zhang}, \bibinfo{person}{Xiaoyan Zhu}, {and} \bibinfo{person}{Bing Liu}.} \bibinfo{year}{2018}\natexlab{}.
\newblock \showarticletitle{Emotional chatting machine: Emotional conversation generation with internal and external memory}. In \bibinfo{booktitle}{\emph{Proceedings of the AAAI Conference on Artificial Intelligence}}, Vol.~\bibinfo{volume}{32}.
\newblock


\bibitem[Zhu et~al\mbox{.}(2021)]%
        {zhu2021eit}
\bibfield{author}{\bibinfo{person}{Junyi Zhu}, \bibinfo{person}{Jackson~C Snowden}, \bibinfo{person}{Joshua Verdejo}, \bibinfo{person}{Emily Chen}, \bibinfo{person}{Paul Zhang}, \bibinfo{person}{Hamid Ghaednia}, \bibinfo{person}{Joseph~H Schwab}, {and} \bibinfo{person}{Stefanie Mueller}.} \bibinfo{year}{2021}\natexlab{}.
\newblock \showarticletitle{EIT-kit: An electrical impedance tomography toolkit for health and motion sensing}. In \bibinfo{booktitle}{\emph{The 34th Annual ACM Symposium on User Interface Software and Technology}}. \bibinfo{pages}{400--413}.
\newblock


\end{thebibliography}

\appendix

% % \begin{appendices}

\section{Dimensions for Day-to-day Functioning Screening}
\label{appendix:37Dimension}

This section presents the 37 dimensions proposed by~\cite{nie2022conversational} to screen day-to-day functioning. These dimensions are based on the Diagnostic and Statistical Manual of Mental Disorders (DSM)-IV and the Daily Living Activities–20 (DLA-20), two gold standards for mental health care providers. As mentioned in Section~\ref{sec:implementation_and_method}, for each dimension, therapists provide a set of sample questions. With the GPT-4-based \texttt{Rephraser}, \sys{} rephrases and formulates the questions when it converses with the user. Note that as some of the dimensions might be sensitive or uncomfortable for users, \sys{} offers the option for users to select the dimensions to work on manually through the smartphone interface. The 37 dimensions and the example questions asked by \sys{} are listed below:
\begin{enumerate}
    \item \emph{Maintaining stable weight}: ``Have your weight changed significantly recently?'';
    \item \emph{Managing mood}: ``How's your mood recently?'';
    \item \emph{Taking medication as prescribed}: ``Have you been taking medication according to your prescriptions?'';
    \item \emph{Participating primary and mental health care}: ``Have you been seeing your doctor, therapist or case manager consistently?'';
    \item \emph{Organizing personal possessions \& doing housework}: ``Have you been doing house chores?'';
    \item \emph{Talking to other people}: ``Have you been talking to other people?'';
    \item \emph{Expressing feelings to other people}: ``Have you expressed feelings towards others?'';
    \item \emph{Managing personal safety}: ``Have you been taking safety into consideration when making decisions?'';
    \item \emph{Managing risk}: ``Have you taken any risk recently?'';
    \item \emph{Following regular schedule for bedtime \& sleeping enough}: ``How has your sleep been? Do you have a regular schedule for bedtime?'';
    \item \emph{Maintaining regular schedule for eating}: ``How's your eating? Are you eating regularly?'';
    \item \emph{Managing work/school}: ``Are you showing up for work or school?'';
    \item \emph{Having work-life balance}: ``How's your work-life balance? Have you taken days off recently?'';
    \item \emph{Showing up for appointments and obligations}: ``Are you showing up for appointments and other obligations?'';
    \item \emph{Managing finance and items of value}: ``How's your finances? Any concern with your spending habits?'';
    \item \emph{Getting adequate nutrition}: ``How's your nutrition? Are you eating healthy?'';
    \item \emph{Problem solving and decision making capability}: ``Are you able to make decisions yourself?'';
    \item \emph{Family support}: ``Do you feel supported by your family?'';
    \item \emph{Family relationship}: ``How's your relationship with your family members?'';
    \item \emph{Alcohol abuse}: ``Do you often drink alone?'';
    \item \emph{Tobacco abuse}: ``Do you smoke cigarettes or vape? And how frequent?''; 
    \item \emph{Other substances abuse}: ``Do you use any substance and what's the frequency of using?'';
    \item \emph{Enjoying personal choices for leisure activities}: ``What do you like to do when you have free time?'';
    \item \emph{Creativity}: ``Have you done any creative work recently?'';
    \item \emph{Participation in community}: ``What do you do in your neighborhood or community?'';
    \item \emph{Support from social network}: ``Other than family members, who do you consider as your close support?'';
    \item \emph{Relationship with friends and colleagues}: ``Do you hang out with friends or coworkers'';
    \item \emph{Managing boundaries in close relationship}: ``Do you feel comfortable with your partner or partners if you have?'';
    \item \emph{Managing sexual safety}: ``If you are sexually active, do you try to avoid risky sexual behaviors?'';
    \item \emph{Productivity at work or school}: ``Are you productive at work or school?'';
    \item \emph{Motivation at work or school}: ``How's your motivation for work or school?'';
    \item \emph{Coping skills to de-stress}: ``What kind of coping do you use to calm yourself?'';
    \item \emph{Exhibiting control over self-harming behavior}: ``Do you have self-harming behaviours?'';
    \item \emph{Law-abiding}: ``Have you been arrested recently?'';
    \item \emph{Managing legal issue}: ``Do you have any legal issue recently?'';
    \item \emph{Maintaining personal hygiene}: ``How's your personal hygiene? Are you taking care of your personal hygiene?'';
    \item \emph{Doing exercises and sports}: ``Have you recently exercised?'';
    
\end{enumerate}
As mentioned in Section~\ref{sec:architecture_chat}, users can freely chat with \sys{} using open-ended responses, including both deterministic YES/NO answers and detailed responses. For example, when \sys{} asks: \emph{``Do you often drink alone?''}, the user can respond deterministically: \emph{``Yes, I drink alone.''} or: \emph{``I drink alone almost every other night recently.''}. \sys{} classifies both responses to \textbf{Dimension} \emph{Alcohol usage} with \textbf{Score} of 2. With \textbf{Score}\thinspace$=$\thinspace2, \sys{} follows up and provides \emph{Psychotherapeutic Conversational Interventions} starting by asking for more information. In addition, if the user says: \emph{``I don't drink alone. But I like to drink with my friends when we hang out together.''}, \sys{} segments the responses into two parts and classifies the first part to \textbf{Dimension} \emph{Alcohol usage} with \textbf{Score} of 0, and the second part to \textbf{Dimension} \emph{Relationship with friends and colleagues} with \textbf{Score} of 0.

\end{document}